\documentclass[11pt]{article}

\PassOptionsToPackage{table,dvipsnames}{xcolor} 
\usepackage[final]{acl}

\usepackage{times}
\usepackage{latexsym}
\usepackage[T1]{fontenc}
\usepackage[utf8]{inputenc}
\usepackage{microtype}
\usepackage{inconsolata}
\usepackage{graphicx}
\graphicspath{{figures/}}

\usepackage{amsmath,amssymb,mathtools}
\usepackage{bm}
\usepackage{booktabs}
\usepackage{multirow}
\usepackage{colortbl} 
\usepackage{makecell}
\usepackage{array}
\usepackage{tabularx}
\usepackage{subcaption}
\usepackage{enumitem}
\usepackage{xspace}
\usepackage{pifont}
\usepackage[most]{tcolorbox}
\usepackage[capitalize,nameinlink,noabbrev]{cleveref} 

\definecolor{cOurs}{HTML}{0072B2}     
\definecolor{cCmpA}{HTML}{E69F00}     
\definecolor{cCmpB}{HTML}{009E73}     
\definecolor{cInterv}{HTML}{D55E00}   
\definecolor{cJudge}{HTML}{CC79A7}    
\definecolor{cCmpC}{HTML}{56B4E9}     
\definecolor{cBase}{HTML}{8C8C8C}     
\definecolor{cInk}{HTML}{1A1A1A}      
\definecolor{cUp}{HTML}{0072B2}       
\definecolor{cDown}{HTML}{D55E00}     
\definecolor{cNeutral}{HTML}{8C8C8C}  
\definecolor{cRowHi}{HTML}{EAF2F8}    
\definecolor{cBoxBg}{HTML}{F2F2F2}    
\definecolor{cBoxFrame}{HTML}{A3A3A3} 
\definecolor{cMarkBg}{HTML}{FFF6E5}   

\newif\ifpivotA
\pivotAtrue 

\newcommand{\thesistext}{\ifpivotA In all but one of the runs we observe, outcome-based RL does not coin the reflection marker \form{wait}. Where the base model lacks it, zero-RL leaves it absent while accuracy rises; the marker enters with mid-training or with training on reasoning traces; and where it is present, RL changes its frequency and form in a way that depends on the recipe and on the marker's starting state.\else Outcome-based RL is a grammaticalization process for the language of thought.\fi}

\newcommand{\form}[1]{\textit{#1}}
\newcommand{\gloss}[1]{`#1'}
\newcommand{\tok}[1]{\texttt{\small #1}}

\newcounter{hypothesis}
\newenvironment{hypothesis}[1][]{%
  \refstepcounter{hypothesis}\par\smallskip\noindent
  \textbf{H\thehypothesis}\if\relax\detokenize{#1}\relax\else\ (\textit{#1})\fi.\ \ignorespaces}%
  {\par\smallskip}
\crefformat{hypothesis}{#2H#1#3}
\Crefformat{hypothesis}{#2H#1#3}
\crefmultiformat{hypothesis}{#2H#1#3}{ and~#2H#1#3}{, #2H#1#3}{, and~#2H#1#3}
\Crefmultiformat{hypothesis}{#2H#1#3}{ and~#2H#1#3}{, #2H#1#3}{, and~#2H#1#3}
\crefrangeformat{hypothesis}{#3H#1#4--#5H#2#6}
\Crefrangeformat{hypothesis}{#3H#1#4--#5H#2#6}
\newcommand{\hyp}[1]{\textbf{H#1}\xspace}

\newtcolorbox{claimbox}{enhanced,colback=cBoxBg,colframe=cBoxFrame,boxrule=0.5pt,
  arc=1.5pt,left=5pt,right=5pt,top=3pt,bottom=3pt,fontupper=\itshape}
\newcommand{\takeaway}[1]{\begin{tcolorbox}[enhanced,colback=cBoxBg,colframe=cBoxFrame,
  boxrule=0.5pt,arc=1.5pt,left=5pt,right=5pt,top=2pt,bottom=2pt]
  \textbf{Takeaway.} #1\end{tcolorbox}}

\newcommand{\circled}[1]{\ding{\numexpr181+#1\relax}}

\newcommand{\best}[1]{\textbf{#1}}
\newcommand{\second}[1]{\underline{#1}}
\newcommand{\std}[1]{{\scriptsize\,$\pm$\,#1}}
\newcommand{\pms}[2]{#1\std{#2}}

\newcommand{\hup}{\,$\uparrow$}
\newcommand{\hdown}{\,$\downarrow$}
\newcommand{\na}{--}
\newcommand{\ci}[2]{{\scriptsize\,\textcolor{gray}{[#1, #2]}}}   
\newcommand{\nsig}[1]{\textcolor{gray}{#1}}       
\newsavebox{\fitbox} 
\newcommand{\fitwidth}[1]{\sbox{\fitbox}{#1}\ifdim\wd\fitbox>\linewidth\resizebox{\linewidth}{!}{\usebox{\fitbox}}\else\usebox{\fitbox}\fi}

\newcommand{\tabstyle}{\small\setlength{\tabcolsep}{4pt}\renewcommand{\arraystretch}{1.05}}

\newcommand{\resdef}[2]{\expandafter\def\csname res@#1\endcsname{#2}}
\newcommand{\res}[1]{\ifcsname res@#1\endcsname\csname res@#1\endcsname\else\textbf{\textcolor{red}{??\detokenize{#1}??}}\fi} 
\newcommand{\rt}[1]{\ifcsname res@#1\endcsname\csname res@#1\endcsname\else\ifdraft\textcolor{gray}{--.-}\else\errmessage{Missing result key in table: \detokenize{#1}}\fi\fi}
\newcommand{\cell}[1]{\pms{\rt{#1}}{\rt{#1.hw}}}

\newif\ifdraft
\drafttrue

\newcommand{\Prob}{\mathbb{P}}

\newcommand{\vtheta}{\bm{\theta}}

\newcommand{\ckpt}[1]{\vtheta_{#1}}          

\renewcommand{\topfraction}{0.85}
\renewcommand{\dbltopfraction}{0.85}

\renewcommand{\floatpagefraction}{0.75}
\renewcommand{\dblfloatpagefraction}{0.75}

\usepackage{titletoc}
\usepackage{placeins} 
\resdef{h1.s1.base.chain}{117.9}  
\resdef{h1.s1.base.commadrop}{2.8}  
\resdef{h1.s1.base.freq}{5.2}  
\resdef{h1.s1.base.freq.hw}{0.4}  
\resdef{h1.s1.base.freq100}{9.4}  
\resdef{h1.s1.base.hpara}{2.54}  
\resdef{h1.s1.base.hpara.hw}{0.09}  
\resdef{h1.s1.base.neff}{5.82}  
\resdef{h1.s1.base.obl}{55.3}  
\resdef{h1.s1.base.obl.hw}{3.2}  
\resdef{h1.s1.base.orth}{87.9}  
\resdef{h1.s1.base.orth.hw}{1.2}  
\resdef{h1.s1.base.rel}{10.3}  
\resdef{h1.s1.base.rel.hw}{1.5}  
\resdef{h1.s1.base.revrate}{2.24}  
\resdef{h1.s1.base.revrate.hw}{0.32}  
\resdef{h1.s1.base.sentlen}{11.5}  
\resdef{h1.s1.base.sharewait}{33.1}  
\resdef{h1.s1.base.si}{88.7}  
\resdef{h1.s1.base.si.hw}{1.2}  
\resdef{h1.s1.late.freq}{18.1}  
\resdef{h1.s1.late.freq.hw}{3.1}  
\resdef{h1.s1.late.hpara}{2.33}  
\resdef{h1.s1.late.hpara.hw}{0.07}  
\resdef{h1.s1.late.neff}{5.05}  
\resdef{h1.s1.late.obl}{52.8}  
\resdef{h1.s1.late.obl.hw}{4.1}  
\resdef{h1.s1.late.orth}{41.5}  
\resdef{h1.s1.late.orth.hw}{6.2}  
\resdef{h1.s1.late.rel}{3.0}  
\resdef{h1.s1.late.rel.hw}{0.5}  
\resdef{h1.s1.late.revrate}{0.9}  
\resdef{h1.s1.late.revrate.hw}{0.1}  
\resdef{h1.s1.late.sharewait}{47.4}  
\resdef{h1.s1.late.si}{89.7}  
\resdef{h1.s1.late.si.hw}{1.7}  
\resdef{h1.s1.mid.freq}{7.5}  
\resdef{h1.s1.mid.freq.hw}{0.5}  
\resdef{h1.s1.mid.hpara}{2.53}  
\resdef{h1.s1.mid.hpara.hw}{0.06}  
\resdef{h1.s1.mid.obl}{61.5}  
\resdef{h1.s1.mid.obl.hw}{2.9}  
\resdef{h1.s1.mid.orth}{88.4}  
\resdef{h1.s1.mid.orth.hw}{1.2}  
\resdef{h1.s1.mid.rel}{7.7}  
\resdef{h1.s1.mid.rel.hw}{0.9}  
\resdef{h1.s1.mid.revrate}{1.77}  
\resdef{h1.s1.mid.revrate.hw}{0.19}  
\resdef{h1.s1.mid.si}{89.0}  
\resdef{h1.s1.mid.si.hw}{1.1}  
\resdef{h1.s1.trend.rho_freq}{1.0}  
\resdef{h1.s1.trend.rho_hpara}{-0.9}  
\resdef{h1.s1.trend.rho_obl}{-0.3}  
\resdef{h1.s1.trend.rho_orth}{-0.1}  
\resdef{h1.s1.trend.rho_rel}{-1.0}  
\resdef{h1.s1.trend.rho_si}{0.8}  
\resdef{h4.s1.base.falserev_wait_late}{0.0}  
\resdef{h4.s1.base.or_op_late}{2.43}  
\resdef{h4.s1.base.prev_hmm_late}{7.0}  
\resdef{h4.s1.base.prev_holdon_late}{14.5}  
\resdef{h4.s1.base.prev_next_late}{5.0}  
\resdef{h4.s1.base.prev_none_late}{4.5}  
\resdef{h4.s1.base.prev_okay_late}{4.8}  
\resdef{h4.s1.base.prev_wait_late}{11.5}  
\resdef{h4.s1.base.rd_op_late}{6.5}  
\resdef{h4.s1.base.rd_op_late.hw}{4.0}  
\resdef{h4.s1.base.rd_syn_late}{0.8}  
\resdef{h4.s1.base.rr_op_late}{2.27}  
\resdef{h4.s1.late.falserev_wait_late}{0.5}  
\resdef{h4.s1.late.or_op_late}{2.84}  
\resdef{h4.s1.late.prev_hmm_late}{12.2}  
\resdef{h4.s1.late.prev_holdon_late}{16.2}  
\resdef{h4.s1.late.prev_next_late}{7.0}  
\resdef{h4.s1.late.prev_none_late}{5.0}  
\resdef{h4.s1.late.prev_okay_late}{7.2}  
\resdef{h4.s1.late.prev_wait_late}{17.8}  
\resdef{h4.s1.late.rd_op_late}{10.8}  
\resdef{h4.s1.late.rd_op_late.hw}{4.6}  
\resdef{h4.s1.late.rd_syn_late}{3.5}  
\resdef{h4.s1.late.rr_op_late}{2.51}  
\resdef{h4.s1.mid.falserev_wait_late}{0.2}  
\resdef{h4.s1.mid.or_op_late}{3.42}  
\resdef{h4.s1.mid.prev_hmm_late}{8.2}  
\resdef{h4.s1.mid.prev_holdon_late}{12.0}  
\resdef{h4.s1.mid.prev_next_late}{3.8}  
\resdef{h4.s1.mid.prev_none_late}{4.0}  
\resdef{h4.s1.mid.prev_okay_late}{3.5}  
\resdef{h4.s1.mid.prev_wait_late}{12.0}  
\resdef{h4.s1.mid.rd_op_late}{8.2}  
\resdef{h4.s1.mid.rd_op_late.hw}{4.0}  
\resdef{h4.s1.mid.rd_syn_late}{1.9}  
\resdef{h4.s1.mid.rr_op_late}{3.13}  
\resdef{h4.s1.trend.r_amp_late}{1.65}  
\resdef{gold.all.kappabin_human_judge}{0.19}  
\resdef{gold.all.kappabin_human_oss}{0.63}  
\resdef{gold.all.n_human_judge}{59}  
\resdef{gold.s1.agree5_llm_llm}{71.7}  
\resdef{gold.s1.kappa5_llm_llm}{0.54}  
\resdef{gold.v2.kappabin_human_judge}{0.3}  
\resdef{gold.v2.kappabin_human_oss}{0.43}  
\resdef{gold.v2.sens_judge}{26.8}  
\resdef{gold.v2.spec_judge}{93.9}  
\resdef{gold.v4.B.n_stage1}{27}  
\resdef{gold.v4.B.n_stage2}{27}  
\resdef{gold.v4.B.stage1_yes}{2}  
\resdef{gold.v4.B.stage2_yes}{15}  
\resdef{gold.v4.kappa_human_oss}{0.71}  
\resdef{gold.v4.kappa_human_qwen}{0.64}  
\resdef{gold.v4.kappa_human_rule}{0.3}  
\resdef{gold.v4.n_A}{60}  
\resdef{gold.v4.revrate_1900}{15.9}  
\resdef{gold.v4.revrate_1900.hi}{36.0}  
\resdef{gold.v4.revrate_1900.lo}{4.4}  
\resdef{gold.v4.revrate_base}{1.3}  
\resdef{gold.v4.revrate_judge_1900}{12.9}  
\resdef{gold.v4.revrate_judge_base}{26.0}  
\resdef{gold.v4.revrate_judge_oss_1900}{12.9}  
\resdef{gold.v4.revrate_judge_oss_base}{1.5}  
\resdef{gold.gpt6.kappa_oss}{0.66}  
\resdef{gold.gpt6.kappa_qwen}{0.54}  
\resdef{gold.gpt6.kappa_rule}{0.33}  
\resdef{gold.gpt6.n_A}{48}  
\resdef{h1.s1.base.chain}{117.6}  
\resdef{h1.s1.base.commadrop}{2.7}  
\resdef{h1.s1.base.freq}{4.9}  
\resdef{h1.s1.base.freq.hw}{0.4}  
\resdef{h1.s1.base.freq100}{9.2}  
\resdef{h1.s1.base.hpara}{2.6}  
\resdef{h1.s1.base.hpara.hw}{0.08}  
\resdef{h1.s1.base.neff}{6.05}  
\resdef{h1.s1.base.obl}{49.6}  
\resdef{h1.s1.base.obl.hw}{5.6}  
\resdef{h1.s1.base.orth}{88.9}  
\resdef{h1.s1.base.orth.hw}{1.0}  
\resdef{h1.s1.base.rel}{51.0}  
\resdef{h1.s1.base.rel.hw}{4.4}  
\resdef{h1.s1.base.revrate}{17.99}  
\resdef{h1.s1.base.revrate.hw}{2.16}  
\resdef{h1.s1.base.sentlen}{11.8}  
\resdef{h1.s1.base.sharewait}{31.2}  
\resdef{h1.s1.base.si}{89.6}  
\resdef{h1.s1.base.si.hw}{0.9}  
\resdef{h1.s1.final.chain}{213.0}  
\resdef{h1.s1.final.commadrop}{57.8}  
\resdef{h1.s1.final.freq}{12.4}  
\resdef{h1.s1.final.obl}{46.1}  
\resdef{h1.s1.final.orth}{37.5}  
\resdef{h1.s1.final.sentlen}{6.1}  
\resdef{h1.s1.final.si}{89.4}  
\resdef{h1.s1.late.freq}{17.0}  
\resdef{h1.s1.late.freq.hw}{2.7}  
\resdef{h1.s1.late.hpara}{2.37}  
\resdef{h1.s1.late.hpara.hw}{0.05}  
\resdef{h1.s1.late.neff}{5.19}  
\resdef{h1.s1.late.obl}{51.8}  
\resdef{h1.s1.late.obl.hw}{7.0}  
\resdef{h1.s1.late.orth}{42.3}  
\resdef{h1.s1.late.orth.hw}{5.4}  
\resdef{h1.s1.late.rel}{42.0}  
\resdef{h1.s1.late.rel.hw}{6.8}  
\resdef{h1.s1.late.revrate}{13.8}  
\resdef{h1.s1.late.revrate.hw}{1.83}  
\resdef{h1.s1.late.sharewait}{45.6}  
\resdef{h1.s1.late.si}{89.7}  
\resdef{h1.s1.late.si.hw}{1.5}  
\resdef{h1.s1.mid.freq}{8.0}  
\resdef{h1.s1.mid.freq.hw}{0.3}  
\resdef{h1.s1.mid.hpara}{2.49}  
\resdef{h1.s1.mid.hpara.hw}{0.05}  
\resdef{h1.s1.mid.obl}{61.9}  
\resdef{h1.s1.mid.obl.hw}{6.4}  
\resdef{h1.s1.mid.orth}{89.4}  
\resdef{h1.s1.mid.orth.hw}{0.7}  
\resdef{h1.s1.mid.rel}{50.8}  
\resdef{h1.s1.mid.rel.hw}{4.6}  
\resdef{h1.s1.mid.revrate}{14.4}  
\resdef{h1.s1.mid.revrate.hw}{1.77}  
\resdef{h1.s1.mid.si}{90.1}  
\resdef{h1.s1.mid.si.hw}{0.7}  
\resdef{h1.s1.step0400.freq100}{13.2}  
\resdef{h1.s1.trend.rho_freq}{1.0}  
\resdef{h1.s1.trend.rho_hpara}{\ensuremath{-}0.9}  
\resdef{h1.s1.trend.rho_obl}{\ensuremath{-}0.17}  
\resdef{h1.s1.trend.rho_orth}{\ensuremath{-}0.33}  
\resdef{h1.s1.trend.rho_rel}{\ensuremath{-}0.83}  
\resdef{h1.s1.trend.rho_si}{0.55}  
\resdef{h4.s1.base.falserev_wait_late}{0.2}  
\resdef{h4.s1.base.or_op_late}{2.5}  
\resdef{h4.s1.base.prev_hmm_late}{8.8}  
\resdef{h4.s1.base.prev_holdon_late}{13.3}  
\resdef{h4.s1.base.prev_next_late}{5.0}  
\resdef{h4.s1.base.prev_none_late}{3.8}  
\resdef{h4.s1.base.prev_okay_late}{4.4}  
\resdef{h4.s1.base.prev_wait_late}{11.7}  
\resdef{h4.s1.base.rd_op_late}{6.7}  
\resdef{h4.s1.base.rd_op_late.hw}{2.2}  
\resdef{h4.s1.base.rd_syn_late}{0.6}  
\resdef{h4.s1.base.rr_op_late}{2.32}  
\resdef{h4.s1.final.rd_op_late}{15.8}  
\resdef{h4.s1.late.falserev_wait_late}{0.3}  
\resdef{h4.s1.late.or_op_late}{4.5}  
\resdef{h4.s1.late.prev_hmm_late}{13.2}  
\resdef{h4.s1.late.prev_holdon_late}{21.3}  
\resdef{h4.s1.late.prev_next_late}{5.5}  
\resdef{h4.s1.late.prev_none_late}{6.7}  
\resdef{h4.s1.late.prev_okay_late}{5.4}  
\resdef{h4.s1.late.prev_wait_late}{20.8}  
\resdef{h4.s1.late.rd_op_late}{15.3}  
\resdef{h4.s1.late.rd_op_late.hw}{3.3}  
\resdef{h4.s1.late.rd_syn_late}{3.6}  
\resdef{h4.s1.late.rr_op_late}{3.77}  
\resdef{h4.s1.mid.falserev_wait_late}{0.4}  
\resdef{h4.s1.mid.or_op_late}{2.86}  
\resdef{h4.s1.mid.prev_hmm_late}{9.0}  
\resdef{h4.s1.mid.prev_holdon_late}{15.1}  
\resdef{h4.s1.mid.prev_next_late}{4.8}  
\resdef{h4.s1.mid.prev_none_late}{4.8}  
\resdef{h4.s1.mid.prev_okay_late}{5.0}  
\resdef{h4.s1.mid.prev_wait_late}{12.8}  
\resdef{h4.s1.mid.rd_op_late}{7.9}  
\resdef{h4.s1.mid.rd_op_late.hw}{2.4}  
\resdef{h4.s1.mid.rd_syn_late}{0.7}  
\resdef{h4.s1.mid.rr_op_late}{2.62}  
\resdef{h4.s1.trend.r_amp_late}{2.36}  
\resdef{h4.s1.final.rd_op_late_oss}{5.4}  
\resdef{h4.s1.late.rd_op_late_oss}{6.2}  
\resdef{h2.s1.base.crit}{0.11}  
\resdef{h2.s1.base.dprime}{0.57}  
\resdef{h2.s1.base.fa}{34.7}  
\resdef{h2.s1.base.fa_wait}{19.9}  
\resdef{h2.s1.base.hr_wait}{47.1}  
\resdef{h2.s1.final.crit}{\ensuremath{-}0.24}  
\resdef{h2.s1.final.dprime}{1.09}  
\resdef{h2.s1.final.dprime.hi}{1.24}  
\resdef{h2.s1.final.dprime.lo}{0.97}  
\resdef{h2.s1.final.fa}{37.8}  
\resdef{h2.s1.final.fa_wait}{25.4}  
\resdef{h2.s1.final.hr_wait}{73.7}  
\resdef{h4.s1.final.rdval_op_late}{\ensuremath{-}2.3}  
\resdef{h4.s1.final.rdval_op_late.hi}{0.8}  
\resdef{h4.s1.final.rdval_op_late.lo}{\ensuremath{-}5.7}  
\resdef{h4.s1.late.nexcl_val_late}{172}  
\resdef{h4.s1.late.rdval_op_base}{0.0}  
\resdef{h4.s1.late.rdval_op_base.hi}{2.5}  
\resdef{h4.s1.late.rdval_op_base.lo}{\ensuremath{-}2.5}  
\resdef{h4.s1.late.rdval_op_late}{\ensuremath{-}1.2}  
\resdef{h4.s1.late.rdval_op_late.hi}{2.0}  
\resdef{h4.s1.late.rdval_op_late.lo}{\ensuremath{-}4.5}  
\resdef{h4.s1.trend.rho_rdval_late}{\ensuremath{-}0.38}  
\resdef{h1.s1pre.stage2-step47684.orth}{87.8}  
\resdef{h1.s1pre.stage2-step47684.si}{88.6}  
\resdef{h1.s1pre.stage3-step6000.orth}{89.7}  
\resdef{h1.s1pre.trend.rho_chain}{0.89}  
\resdef{h1.s1pre.trend.rho_rel}{\ensuremath{-}1.0}  
\resdef{h1.s2.base.freq}{3.9}  
\resdef{h1.s2.base.freq.hw}{0.3}  
\resdef{h1.s2.base.hpara}{2.87}  
\resdef{h1.s2.base.hpara.hw}{0.07}  
\resdef{h1.s2.base.obl}{44.0}  
\resdef{h1.s2.base.obl.hw}{6.8}  
\resdef{h1.s2.base.orth}{55.0}  
\resdef{h1.s2.base.orth.hw}{3.3}  
\resdef{h1.s2.base.rel}{46.6}  
\resdef{h1.s2.base.rel.hw}{3.8}  
\resdef{h1.s2.base.revrate}{13.92}  
\resdef{h1.s2.base.revrate.hw}{1.98}  
\resdef{h1.s2.base.si}{86.6}  
\resdef{h1.s2.base.si.hw}{1.5}  
\resdef{h1.s2.late.freq}{4.5}  
\resdef{h1.s2.late.freq.hw}{0.3}  
\resdef{h1.s2.late.hpara}{2.8}  
\resdef{h1.s2.late.hpara.hw}{0.04}  
\resdef{h1.s2.late.obl}{43.6}  
\resdef{h1.s2.late.obl.hw}{6.2}  
\resdef{h1.s2.late.orth}{78.2}  
\resdef{h1.s2.late.orth.hw}{2.9}  
\resdef{h1.s2.late.rel}{46.2}  
\resdef{h1.s2.late.rel.hw}{4.9}  
\resdef{h1.s2.late.revrate}{14.11}  
\resdef{h1.s2.late.revrate.hw}{2.02}  
\resdef{h1.s2.late.si}{93.6}  
\resdef{h1.s2.late.si.hw}{0.7}  
\resdef{h1.s2.mid.freq}{4.4}  
\resdef{h1.s2.mid.freq.hw}{0.3}  
\resdef{h1.s2.mid.hpara}{2.78}  
\resdef{h1.s2.mid.hpara.hw}{0.05}  
\resdef{h1.s2.mid.obl}{53.7}  
\resdef{h1.s2.mid.obl.hw}{6.5}  
\resdef{h1.s2.mid.orth}{58.3}  
\resdef{h1.s2.mid.orth.hw}{3.7}  
\resdef{h1.s2.mid.rel}{44.8}  
\resdef{h1.s2.mid.rel.hw}{4.3}  
\resdef{h1.s2.mid.revrate}{11.96}  
\resdef{h1.s2.mid.revrate.hw}{1.78}  
\resdef{h1.s2.mid.si}{89.1}  
\resdef{h1.s2.mid.si.hw}{1.4}  
\resdef{h1.s2.trend.rho_freq}{0.4}  
\resdef{h1.s2.trend.rho_hpara}{\ensuremath{-}0.48}  
\resdef{h1.s2.trend.rho_obl}{\ensuremath{-}0.14}  
\resdef{h1.s2.trend.rho_orth}{0.88}  
\resdef{h1.s2.trend.rho_rel}{\ensuremath{-}0.48}  
\resdef{h1.s2.trend.rho_si}{0.67}  
\resdef{h4.s2.base.falserev_wait_late}{1.3}  
\resdef{h4.s2.base.or_op_late}{3.47}  
\resdef{h4.s2.base.prev_hmm_late}{7.7}  
\resdef{h4.s2.base.prev_holdon_late}{10.2}  
\resdef{h4.s2.base.prev_next_late}{3.8}  
\resdef{h4.s2.base.prev_none_late}{4.7}  
\resdef{h4.s2.base.prev_okay_late}{5.3}  
\resdef{h4.s2.base.prev_wait_late}{12.2}  
\resdef{h4.s2.base.rd_op_late}{8.4}  
\resdef{h4.s2.base.rd_op_late.hw}{3.0}  
\resdef{h4.s2.base.rd_syn_late}{3.2}  
\resdef{h4.s2.base.rr_op_late}{3.17}  
\resdef{h4.s2.late.falserev_wait_late}{1.2}  
\resdef{h4.s2.late.or_op_late}{2.78}  
\resdef{h4.s2.late.prev_hmm_late}{8.9}  
\resdef{h4.s2.late.prev_holdon_late}{11.8}  
\resdef{h4.s2.late.prev_next_late}{6.0}  
\resdef{h4.s2.late.prev_none_late}{5.1}  
\resdef{h4.s2.late.prev_okay_late}{4.7}  
\resdef{h4.s2.late.prev_wait_late}{15.1}  
\resdef{h4.s2.late.rd_op_late}{9.1}  
\resdef{h4.s2.late.rd_op_late.hw}{3.2}  
\resdef{h4.s2.late.rd_syn_late}{4.7}  
\resdef{h4.s2.late.rr_op_late}{2.51}  
\resdef{h4.s2.mid.falserev_wait_late}{0.8}  
\resdef{h4.s2.mid.or_op_late}{3.68}  
\resdef{h4.s2.mid.prev_hmm_late}{8.2}  
\resdef{h4.s2.mid.prev_holdon_late}{10.8}  
\resdef{h4.s2.mid.prev_next_late}{4.1}  
\resdef{h4.s2.mid.prev_none_late}{5.1}  
\resdef{h4.s2.mid.prev_okay_late}{4.3}  
\resdef{h4.s2.mid.prev_wait_late}{13.8}  
\resdef{h4.s2.mid.rd_op_late}{9.7}  
\resdef{h4.s2.mid.rd_op_late.hw}{3.1}  
\resdef{h4.s2.mid.rd_syn_late}{4.3}  
\resdef{h4.s2.mid.rr_op_late}{3.31}  
\resdef{h4.s2.trend.r_amp_late}{1.09}  
\resdef{h4.s2.trend.r_amp_late.hi}{1.61}  
\resdef{h4.s2.trend.r_amp_late.lo}{0.72}  
\resdef{h2.s2.base.crit}{\ensuremath{-}0.14}  
\resdef{h2.s2.base.dprime}{0.82}  
\resdef{h2.s2.late.crit}{\ensuremath{-}0.19}  
\resdef{h2.s2.late.crit.hi}{\ensuremath{-}0.06}  
\resdef{h2.s2.late.crit.lo}{\ensuremath{-}0.31}  
\resdef{h2.s2.late.dprime}{1.02}  
\resdef{h4.s2.late.rdval_op_late}{\ensuremath{-}1.5}  
\resdef{h4.s2.late.rdval_op_late.hi}{1.2}  
\resdef{h4.s2.late.rdval_op_late.lo}{\ensuremath{-}4.9}  
\resdef{h1.s3a.base.freq}{0.0}  
\resdef{h1.s3a.base.freq.hw}{0.0}  
\resdef{h1.s3a.base.obl}{3.8}  
\resdef{h1.s3a.base.obl.hw}{1.0}  
\resdef{h1.s3a.base.orth}{0.0}  
\resdef{h1.s3a.base.orth.hw}{0.0}  
\resdef{h1.s3a.base.rel}{54.5}  
\resdef{h1.s3a.base.rel.hw}{5.3}  
\resdef{h1.s3a.base.si}{0.0}  
\resdef{h1.s3a.base.si.hw}{0.0}  
\resdef{h1.s3a.late.freq}{0.0}  
\resdef{h1.s3a.late.freq.hw}{0.0}  
\resdef{h1.s3a.late.obl}{3.6}  
\resdef{h1.s3a.late.obl.hw}{1.0}  
\resdef{h1.s3a.late.orth}{0.0}  
\resdef{h1.s3a.late.orth.hw}{0.0}  
\resdef{h1.s3a.late.rel}{54.5}  
\resdef{h1.s3a.late.rel.hw}{6.0}  
\resdef{h1.s3a.late.si}{0.0}  
\resdef{h1.s3a.late.si.hw}{0.0}  
\resdef{h1.s3a.mid.freq}{0.0}  
\resdef{h1.s3a.mid.freq.hw}{0.0}  
\resdef{h1.s3a.mid.obl}{4.5}  
\resdef{h1.s3a.mid.obl.hw}{1.3}  
\resdef{h1.s3a.mid.orth}{0.0}  
\resdef{h1.s3a.mid.orth.hw}{0.0}  
\resdef{h1.s3a.mid.rel}{55.7}  
\resdef{h1.s3a.mid.rel.hw}{5.0}  
\resdef{h1.s3a.mid.si}{0.0}  
\resdef{h1.s3a.mid.si.hw}{0.0}  
\resdef{h1.s3a.trend.rho_freq}{\ensuremath{-}0.2}  
\resdef{h1.s3a.trend.rho_obl}{0.29}  
\resdef{h1.s3a.trend.rho_orth}{\ensuremath{-}0.35}  
\resdef{h1.s3a.trend.rho_rel}{0.21}  
\resdef{h1.s3a.trend.rho_si}{\ensuremath{-}0.35}  
\resdef{h2.s1.base.jsd_floor}{0.002}  
\resdef{h2.s1.final.jsd_base}{0.05}  
\resdef{ctl.s8dapo.final.acc}{76.4}  
\resdef{h1.s10high.final.freq}{0.8}  
\resdef{h1.s10high.final.freq.hw}{0.1}  
\resdef{h1.s10high.final.obl}{20.5}  
\resdef{h1.s10high.final.obl.hw}{3.8}  
\resdef{h1.s10high.final.orth}{9.9}  
\resdef{h1.s10high.final.orth.hw}{2.5}  
\resdef{h1.s10high.final.rel}{2.3}  
\resdef{h1.s10high.final.rel.hw}{0.6}  
\resdef{h1.s10high.final.si}{90.0}  
\resdef{h1.s10high.final.si.hw}{2.4}  
\resdef{h1.s10low.final.freq}{0.4}  
\resdef{h1.s10low.final.freq.hw}{0.1}  
\resdef{h1.s10low.final.obl}{13.0}  
\resdef{h1.s10low.final.obl.hw}{7.1}  
\resdef{h1.s10low.final.orth}{0.0}  
\resdef{h1.s10low.final.orth.hw}{0.0}  
\resdef{h1.s10low.final.rel}{2.7}  
\resdef{h1.s10low.final.rel.hw}{1.5}  
\resdef{h1.s10low.final.si}{95.7}  
\resdef{h1.s10low.final.si.hw}{2.6}  
\resdef{h1.s10med.final.freq}{0.8}  
\resdef{h1.s10med.final.freq.hw}{0.1}  
\resdef{h1.s10med.final.obl}{16.3}  
\resdef{h1.s10med.final.obl.hw}{7.0}  
\resdef{h1.s10med.final.orth}{0.4}  
\resdef{h1.s10med.final.orth.hw}{0.5}  
\resdef{h1.s10med.final.rel}{2.6}  
\resdef{h1.s10med.final.rel.hw}{1.0}  
\resdef{h1.s10med.final.si}{97.4}  
\resdef{h1.s10med.final.si.hw}{1.5}  
\resdef{h1.s8dapo.final.freq}{0.0}  
\resdef{h1.s8dapo.final.freq.hw}{0.0}  
\resdef{h1.s8dapo.final.obl}{6.1}  
\resdef{h1.s8dapo.final.obl.hw}{3.6}  
\resdef{h1.s8dapo.final.orth}{0.0}  
\resdef{h1.s8dapo.final.orth.hw}{0.0}  
\resdef{h1.s8dapo.final.rel}{9.7}  
\resdef{h1.s8dapo.final.rel.hw}{3.5}  
\resdef{h1.s8dapo.final.si}{0.0}  
\resdef{h1.s8dapo.final.si.hw}{0.0}  
\resdef{h1.s8orz.final.freq}{0.0}  
\resdef{h1.s8orz.final.freq.hw}{0.0}  
\resdef{h1.s8orz.final.obl}{7.5}  
\resdef{h1.s8orz.final.obl.hw}{2.0}  
\resdef{h1.s8orz.final.orth}{27.8}  
\resdef{h1.s8orz.final.orth.hw}{22.2}  
\resdef{h1.s8orz.final.rel}{4.2}  
\resdef{h1.s8orz.final.rel.hw}{1.2}  
\resdef{h1.s8orz.final.si}{27.8}  
\resdef{h1.s8orz.final.si.hw}{22.2}  
\resdef{h1.s8qwq.final.freq}{4.4}  
\resdef{h1.s8qwq.final.freq.hw}{0.4}  
\resdef{h1.s8qwq.final.obl}{62.6}  
\resdef{h1.s8qwq.final.obl.hw}{3.8}  
\resdef{h1.s8qwq.final.orth}{89.3}  
\resdef{h1.s8qwq.final.orth.hw}{1.3}  
\resdef{h1.s8qwq.final.rel}{9.3}  
\resdef{h1.s8qwq.final.rel.hw}{1.2}  
\resdef{h1.s8qwq.final.si}{94.8}  
\resdef{h1.s8qwq.final.si.hw}{0.7}  
\resdef{h1.s8r1d.final.freq}{3.3}  
\resdef{h1.s8r1d.final.freq.hw}{0.5}  
\resdef{h1.s8r1d.final.obl}{56.0}  
\resdef{h1.s8r1d.final.obl.hw}{5.8}  
\resdef{h1.s8r1d.final.orth}{87.3}  
\resdef{h1.s8r1d.final.orth.hw}{2.0}  
\resdef{h1.s8r1d.final.rel}{8.9}  
\resdef{h1.s8r1d.final.rel.hw}{1.6}  
\resdef{h1.s8r1d.final.si}{88.8}  
\resdef{h1.s8r1d.final.si.hw}{1.8}  
\resdef{h4.s8base.final.rd_op}{36.2}  
\resdef{h4.s8base.final.prevval_next}{3.8}  
\resdef{h4.s8base.final.prevval_wait}{3.8}  
\resdef{h1.s9.base.freq}{4.5}  
\resdef{h1.s9.base.freq.hw}{0.3}  
\resdef{h1.s9.base.hpara}{2.81}  
\resdef{h1.s9.base.hpara.hw}{0.05}  
\resdef{h1.s9.base.obl}{44.3}  
\resdef{h1.s9.base.obl.hw}{6.9}  
\resdef{h1.s9.base.orth}{80.7}  
\resdef{h1.s9.base.orth.hw}{1.4}  
\resdef{h1.s9.base.rel}{48.6}  
\resdef{h1.s9.base.rel.hw}{4.3}  
\resdef{h1.s9.base.revrate}{14.58}  
\resdef{h1.s9.base.revrate.hw}{2.09}  
\resdef{h1.s9.base.si}{93.2}  
\resdef{h1.s9.base.si.hw}{0.7}  
\resdef{h1.s9.late.freq}{4.2}  
\resdef{h1.s9.late.freq.hw}{0.3}  
\resdef{h1.s9.late.hpara}{2.76}  
\resdef{h1.s9.late.hpara.hw}{0.08}  
\resdef{h1.s9.late.obl}{49.2}  
\resdef{h1.s9.late.obl.hw}{7.3}  
\resdef{h1.s9.late.orth}{68.5}  
\resdef{h1.s9.late.orth.hw}{2.1}  
\resdef{h1.s9.late.rel}{49.6}  
\resdef{h1.s9.late.rel.hw}{4.8}  
\resdef{h1.s9.late.revrate}{12.45}  
\resdef{h1.s9.late.revrate.hw}{1.67}  
\resdef{h1.s9.late.si}{85.6}  
\resdef{h1.s9.late.si.hw}{1.3}  
\resdef{h1.s9.mid.freq}{3.8}  
\resdef{h1.s9.mid.freq.hw}{0.2}  
\resdef{h1.s9.mid.hpara}{2.75}  
\resdef{h1.s9.mid.hpara.hw}{0.06}  
\resdef{h1.s9.mid.obl}{43.8}  
\resdef{h1.s9.mid.obl.hw}{6.9}  
\resdef{h1.s9.mid.orth}{82.0}  
\resdef{h1.s9.mid.orth.hw}{1.5}  
\resdef{h1.s9.mid.rel}{44.6}  
\resdef{h1.s9.mid.rel.hw}{4.1}  
\resdef{h1.s9.mid.revrate}{13.38}  
\resdef{h1.s9.mid.revrate.hw}{1.96}  
\resdef{h1.s9.mid.si}{94.1}  
\resdef{h1.s9.mid.si.hw}{0.7}  
\resdef{h1.s9.trend.rho_freq}{\ensuremath{-}0.61}  
\resdef{h1.s9.trend.rho_hpara}{\ensuremath{-}0.75}  
\resdef{h1.s9.trend.rho_obl}{0.0}  
\resdef{h1.s9.trend.rho_orth}{\ensuremath{-}0.75}  
\resdef{h1.s9.trend.rho_rel}{0.14}  
\resdef{h1.s9.trend.rho_si}{\ensuremath{-}0.71}  
\resdef{bm.s1.base.r1kw}{13.71}  
\resdef{bm.s1.base.r1kw.hw}{0.7}  
\resdef{bm.s1.base.simplerl}{15.12}  
\resdef{bm.s1.base.simplerl.hw}{1.74}  
\resdef{bm.s1.late.r1kw}{25.77}  
\resdef{bm.s1.late.r1kw.hw}{2.7}  
\resdef{bm.s1.late.simplerl}{84.01}  
\resdef{bm.s1.late.simplerl.hw}{12.13}  
\resdef{bm.s1.mid.r1kw}{17.02}  
\resdef{bm.s1.mid.r1kw.hw}{0.52}  
\resdef{bm.s1.mid.simplerl}{35.43}  
\resdef{bm.s1.mid.simplerl.hw}{2.93}  
\resdef{bm.s1.trend.rho_r1kw}{0.98}  
\resdef{bm.s1.trend.rho_simplerl}{1.0}  
\resdef{bm.s2.base.r1kw}{11.43}  
\resdef{bm.s2.base.r1kw.hw}{0.64}  
\resdef{bm.s2.base.simplerl}{24.66}  
\resdef{bm.s2.base.simplerl.hw}{3.38}  
\resdef{bm.s2.late.r1kw}{11.63}  
\resdef{bm.s2.late.r1kw.hw}{0.52}  
\resdef{bm.s2.late.simplerl}{32.35}  
\resdef{bm.s2.late.simplerl.hw}{3.47}  
\resdef{bm.s2.mid.r1kw}{12.02}  
\resdef{bm.s2.mid.r1kw.hw}{0.59}  
\resdef{bm.s2.mid.simplerl}{27.08}  
\resdef{bm.s2.mid.simplerl.hw}{3.28}  
\resdef{bm.s2.trend.rho_r1kw}{\ensuremath{-}0.1}  
\resdef{bm.s2.trend.rho_simplerl}{0.64}  
\resdef{bm.s2ctl.trend.rho_r1kw}{\ensuremath{-}0.8}  
\resdef{bm.s2ctl.trend.rho_r1wait}{\ensuremath{-}0.8}  
\resdef{bm.s2ctl.trend.rho_simplerl}{\ensuremath{-}0.9}  
\resdef{h1.s1greedy.base.freq}{8.2}  
\resdef{h1.s1greedy.base.freq100}{13.4}  
\resdef{h1.s1greedy.step0400.freq}{12.7}  
\resdef{h1.s1greedy.step0400.freq100}{20.5}  
\resdef{bm.s1.late_half.rho_r1kw}{1.0}  
\resdef{bm.s1.late_half.rho_simplerl}{1.0}  
\resdef{h1.s1.late_half.rho_hpara}{\ensuremath{-}0.8}  
\resdef{h1.s1.late_half.rho_orth}{\ensuremath{-}0.8}  
\resdef{fig1.case.base.nchain}{0}  
\resdef{fig1.case.base.ntok}{2{,}162}  
\resdef{fig1.case.base.nwait}{10}  
\resdef{fig1.case.last.nchain}{9}  
\resdef{fig1.case.last.ntok}{2{,}764}  
\resdef{fig1.case.last.nwait}{36}  
\resdef{fig1.case.mid.nchain}{4}  
\resdef{fig1.case.mid.ntok}{2{,}022}  
\resdef{fig1.case.mid.nwait}{16}  
\resdef{fig1.case.pid}{2}  
\resdef{h4.s1.late.n_both_late}{28}  
\resdef{h4.s1.late.n_rev_late}{349}  
\resdef{h4.s1.late.p_rev_given_vc_late}{17.1}  
\resdef{h4.s1.late.p_vc_given_rev_late}{8.0}  
\resdef{h4.s8base.final.p_rev_given_vc}{23.3}  
\resdef{h4.s8base.final.p_vc_given_rev}{2.5}  
\resdef{h1.fork_b0m.step0200.a_boot}{0.82}  
\resdef{h1.fork_b0m.step0400.a_boot}{0.81}  
\resdef{h1.fork_b0m.step0400.a_boot.hi}{0.87}  
\resdef{h1.fork_b0m.step0400.a_boot.lo}{0.76}  
\resdef{h1.fork_b0m.step0400.a_boot_fin}{0.93}  
\resdef{h1.fork_b0m.step0400.a_boot_fin.hi}{1.01}  
\resdef{h1.fork_b0m.step0400.a_boot_fin.lo}{0.85}  
\resdef{h1.fork_ref1m.step0200.a_boot}{0.87}  
\resdef{h1.fork_ref1m.step0400.a_boot}{0.97}  
\resdef{h1.fork_ref1m.step0400.a_boot.hi}{1.04}  
\resdef{h1.fork_ref1m.step0400.a_boot.lo}{0.91}  
\resdef{h1.fork_ref1m.step0400.a_boot_fin}{1.13}  
\resdef{h1.fork_ref1m.step0400.a_boot_fin.hi}{1.22}  
\resdef{h1.fork_ref1m.step0400.a_boot_fin.lo}{1.05}  
\resdef{h1.s1.step0200.a_boot}{1.24}  
\resdef{h1.s1.step0200.a_boot_fin}{1.34}  
\resdef{h1.s1.step0400.a_boot}{1.58}  
\resdef{h1.s1.step0400.a_boot_fin}{1.7}  
\resdef{design.nckpt}{96}  
\resdef{n.traces.s1}{9{,}000}  
\resdef{n.traces.s10}{2{,}400}  
\resdef{n.traces.s1pre}{5{,}000}  
\resdef{n.traces.s2}{4{,}800}  
\resdef{n.traces.s2ctl}{3{,}000}  
\resdef{n.traces.s3a}{14{,}000}  
\resdef{n.traces.s7}{80{,}000}  
\resdef{n.traces.s8}{4{,}000}  
\resdef{n.traces.s9}{7{,}000}  
\resdef{h1.ds_gsm8k.szq7.final.nwait}{3}  
\resdef{h1.fork_b0.trend.a_wait}{0.87}  
\resdef{h1.fork_ref1.trend.a_wait}{0.96}  
\resdef{h1.fork_b0m.trend.a_wait}{0.81}  
\resdef{h1.fork_ref1m.trend.a_wait}{0.97}  
\resdef{ctl.s7r1.final.acc}{74.4}  
\resdef{ctl.s7r2.final.acc}{74.2}  
\resdef{ctl.s7r7b.base.acc}{16.9}  
\resdef{ctl.s7r7b.final.acc}{76.0}  
\resdef{h1.s1pre.stage1-step1413814.freq2}{0.04}  
\resdef{h1.s1pre.stage2-step47684.freq2}{4.03}  
\resdef{h1.s1pre.stage3-step6000.freq2}{5.95}  
\resdef{h1.s7r1.final.nwait}{0}  
\resdef{h1.s7r2.final.nwait}{0}  
\resdef{h1.s7r7b.final.ntok}{967{,}101}  
\resdef{h1.s7r7b.final.nwait}{0}  
\resdef{h1.s8qwq.final.freq2}{4.42}  
\resdef{h1.s8r1d.final.freq2}{3.25}  
\resdef{gold.all.ac1bin_human_judge}{0.79}  
\resdef{gold.all.ac1bin_human_judge_oss}{0.92}  
\resdef{gold.s1.v2_ac1bin_human_judge}{0.34}  
\resdef{gold.s1.v2_ac1bin_human_judge_oss}{0.69}  
\resdef{h1.s2.trend.rho_orth.p_holm}{0.029}  
\resdef{h1.pop_phi4r.final.freq2}{1.56}  
\resdef{h1.pop_phi4rp.final.freq2}{2.58}  
\resdef{h1.pop_qwen3_8b.final.freq2}{3.93}  
\resdef{h1.pop_r1d_llama8.final.freq2}{4.8}  
\resdef{h1.prov.n_rows}{120}  
\resdef{h1.prov.rl_instruct.n}{8}  
\resdef{h1.prov.sft_free.n}{14}  
\resdef{h1.prov.sft_free.n_basemissing}{4}  
\resdef{h1.prov.sft_free.n_ge05}{13}  
\resdef{h1.prov.sft_marked.n}{9}  
\resdef{h1.s7r1.final.reflopen_share_hmm}{0.0}  
\resdef{h1.s7r1.final.reflopen_share_hold_on}{0.0}  
\resdef{h1.s7r1.final.reflopen_share_however}{99.6}  
\resdef{h1.s7r1.final.reflopen_share_let_me_check}{0.0}  
\resdef{h1.s7r2.final.reflopen_share_however}{100.0}  
\resdef{h1.s7r7b.final.reflopen_share_however}{100.0}  
\resdef{ctl.s1t06.base.trunc}{17.8}  
\resdef{ctl.s1t06.final.trunc}{29.1}  
\resdef{h1.s1seeds.final.freq_mean}{12.68}  
\resdef{h1.s1seeds.final.freq_n}{6}  
\resdef{h1.s1seeds.final.freq_std}{1.55}  
\resdef{h1.s1t06.base.freq}{5.1}  
\resdef{h1.s1t06.final.freq}{76.5}  
\resdef{h4.s1seed1.base.rd_op_base}{12.0}  
\resdef{h4.s1seed1.base.rd_op_base_oss}{5.0}  
\resdef{h4.s1seed1.final.rd_op_base}{26.3}  
\resdef{h4.s1seed1.final.rd_op_base.hi}{32.5}  
\resdef{h4.s1seed1.final.rd_op_base.lo}{20.7}  
\resdef{h4.s1seed1.final.rd_op_base_oss}{12.8}  
\resdef{h4.s1seed1.late.rd_op_base}{24.2}  
\resdef{h4.s1seed1.late.rd_op_base_oss}{14.0}  
\resdef{h2.s9.late.crit}{\ensuremath{-}0.05}  
\resdef{h2.s9.late.dprime}{0.86}  
\resdef{h2.s9.late.dprime.hi}{1.0}  
\resdef{h2.s9.late.dprime.lo}{0.73}  
\resdef{h2.s9.step0100.crit}{\ensuremath{-}0.2}  
\resdef{h2.s9.step0100.dprime}{0.97}  
\resdef{ctl.seed.W512.sft.acc}{41.0}  
\resdef{ctl.seed.W512.sft.toks_per_resp}{2{,}466}  
\resdef{ctl.seed.W512.sft.trunc}{48.4}  
\resdef{ctl.seed.W512.step0400.acc}{77.6}  
\resdef{ctl.seed.W512.step0400.toks_per_resp}{767}  
\resdef{ctl.seed.W64.sft.trunc}{18.5}  
\resdef{h1.seed.D512.trend.a_nextopen}{7.5015}  
\resdef{h1.seed.D512.trend.a_wait}{0.0249}  
\resdef{h1.seed.D64.trend.a_nextopen}{4.7824}  
\resdef{h1.seed.D64.trend.a_wait}{0.0}  
\resdef{h1.seed.W512.sft.freq4}{7.5221}  
\resdef{h1.seed.W512.sft.freq_finished2}{6.91}  
\resdef{h1.seed.W512.step0400.freq4}{0.0}  
\resdef{h1.seed.W512.step0400.freq_finished2}{0.0}  
\resdef{h1.seed.W512.step0400.ntok}{3{,}066{,}514}  
\resdef{h1.seed.W512.step0400.nwait}{0}  
\resdef{h1.seed.W512.trend.a_nextopen}{4.8272}  
\resdef{h1.seed.W512.trend.a_wait}{0.0}  
\resdef{h1.seed.W64.step0400.ntok}{2{,}523{,}676}  
\resdef{h1.seed.W64.step0400.nwait}{2}  
\resdef{h1.seed.W64.trend.a_nextopen}{3.3708}  
\resdef{h1.seed.W64.trend.a_wait}{0.0005}  
\resdef{h1.seed.X512.trend.a_nextopen}{0.1173}  
\resdef{h1.seed.X512.trend.a_wait}{0.0986}  
\resdef{h1.seed.W512LY.sft.acc}{43.9}  
\resdef{h1.seed.W512LY.sft.trunc}{23.3}  
\resdef{h1.seed.W512LY.step0400.acc}{68.2}  
\resdef{h1.seed.W512LY.step0400.trunc}{18.6}  
\resdef{h1.seed.W512LY.trend.a_wait}{0.0003}  
\resdef{h1.seed.W512LY.trend.a_wait.hi}{0.0005}  
\resdef{h1.seed.W512LY.trend.a_wait.lo}{0.0002}  
\resdef{h1.seed.W512LY.trend.a_wait_fin}{0.0003}  
\resdef{h1.seed.W512LY.trend.a_wait_fin.hi}{0.0006}  
\resdef{h1.seed.W512LY.trend.a_wait_fin.lo}{0.0001}  
\resdef{h1.seed.X512LY.sft.acc}{40.2}  
\resdef{h1.seed.X512LY.step0400.acc}{66.6}  
\resdef{h1.seed.X512LY.step0400.ntok}{8{,}239{,}334}  
\resdef{h1.seed.X512LY.step0400.nwait}{0}  
\resdef{h1.seed.X512LY.trend.a_wait}{0.0}  
\resdef{ctl.seed.W512.sftcond.c3072.acc}{45.4}  
\resdef{ctl.seed.W512.sftcond.c3072.loop}{15.4}  
\resdef{ctl.seed.W512.sftcond.c3072.toks}{2{,}416}  
\resdef{ctl.seed.W512.sftcond.c3072.trunc}{46.8}  
\resdef{ctl.seed.W512.sftcond.c8192.acc}{45.2}  
\resdef{ctl.seed.W512.sftcond.c8192.loop}{15.2}  
\resdef{ctl.seed.W512.sftcond.c8192.toks}{3{,}954}  
\resdef{ctl.seed.W512.sftcond.c8192.trunc}{23.1}  
\resdef{ctl.seed.W512.sftcond.yarn.acc}{47.4}  
\resdef{ctl.seed.W512.sftcond.yarn.loop}{20.4}  
\resdef{ctl.seed.W512.sftcond.yarn.toks}{4{,}670}  
\resdef{ctl.seed.W512.sftcond.yarn.trunc}{22.3}  
\resdef{h1.seed.W512.sftcond.c3072.freq2}{7.55}  
\resdef{h1.seed.W512.sftcond.c3072.freq_finished2}{7.07}  
\resdef{h1.seed.W512.sftcond.c8192.freq2}{9.13}  
\resdef{h1.seed.W512.sftcond.c8192.freq_finished2}{6.32}  
\resdef{h1.seed.W512.sftcond.yarn.freq2}{6.82}  
\resdef{h1.seed.W512.sftcond.yarn.freq_finished2}{7.52}  
\resdef{gold.s1.rev_qwen3_yes_n}{19}  
\resdef{gold.s1.vc_n_checked}{50}  
\resdef{gold.s1.vc_n_yes}{5}  
\resdef{gold.s1.vc_precision_qwen3}{10.0}  
\resdef{gold.s1.vc_precision_qwen3.hi}{21.4}  
\resdef{gold.s1.vc_precision_qwen3.lo}{4.3}  
\resdef{gold.s1.vc_recall_proxy_qwen3}{0.0}  
\resdef{h1.bb_mimo_base.final.freq2}{4.95}  
\resdef{h1.bb_mimo_rlzero.final.freq2}{6.62}  
\resdef{h1.bb_olmo2_base.final.freq2}{0.16}  
\resdef{h1.bb_orn7.final.open_but}{98.3}  
\resdef{h1.bb_orn7.final.open_perhaps}{28.1}  
\resdef{h1.bb_orn7.final.open_wait}{0.2}  
\resdef{h1.bb_smollm3_base.final.freq2}{4.18}  
\resdef{ctl.bb_pangu_v21pro.final.n}{745}  
\resdef{ctl.bb_pangu_v21pro.final.n_problems}{190}  
\resdef{h1.bb_pangu_v21pro.complete.freq2}{0.75}  
\resdef{h1.bb_pangu_v21pro.complete.n_problems}{182}  
\resdef{h1.bb_pangu_v21pro.final.freq2}{0.77}  
\resdef{h1.bb_gr3_4_base.final.ntok}{434{,}190}  
\resdef{h1.bb_gr3_4_base.final.nwait}{1}  
\resdef{h1.bb_sz_m24.final.ntok}{1{,}500{,}001}  
\resdef{h1.bb_sz_m24.final.nwait}{7}  
\resdef{h1.s2ctl.step43000.commadrop}{4.3}  
\resdef{h1.s2ctl.step43000.freq}{4.8}  
\resdef{h1.s2ctl.step43000.orth}{91.7}  
\resdef{h1.s2ctl.step5000.commadrop}{4.0}  
\resdef{h1.s2ctl.step5000.freq}{5.1}  
\resdef{h1.s2ctl.step5000.orth}{92.1}  
\resdef{h1.s2ctl.trend.rho_chain}{\ensuremath{-}1.0}  
\resdef{h1.s2ctl.trend.rho_commadrop}{0.8}  
\resdef{h1.s2ctl.trend.rho_freq}{\ensuremath{-}0.6}  
\resdef{h1.s2ctl.trend.rho_hpara}{0.0}  
\resdef{h1.s2ctl.trend.rho_obl}{\ensuremath{-}0.7}  
\resdef{h1.s2ctl.trend.rho_orth}{\ensuremath{-}0.6}  
\resdef{h1.s2ctl.trend.rho_rel}{\ensuremath{-}0.1}  
\resdef{h1.s2ctl.trend.rho_si}{\ensuremath{-}0.1}  
\resdef{h1.route.nobase.median}{4.24}  
\resdef{h1.route.nobase.n}{4}  
\resdef{h1.route.nobase.n_ge05}{4}  
\resdef{h1.route.postmarked.median}{3.77}  
\resdef{h1.route.postmarked.n}{9}  
\resdef{h1.route.postmarked.n_ge05}{9}  
\resdef{h1.route.rlmarked.median}{4.23}  
\resdef{h1.route.rlmarked.n}{9}  
\resdef{h1.route.rlmarked.n_ge05}{9}  
\resdef{h1.route.sftfree.median}{3.54}  
\resdef{h1.route.sftfree.n}{14}  
\resdef{h1.route.sftfree.n_ge05}{13}  
\resdef{h2.s1.base.crit_wait}{0.46}  
\resdef{h2.s1.base.dprime_wait}{0.77}  
\resdef{h2.s1.final.crit_wait}{0.01}  
\resdef{h2.s1.final.dprime_wait}{1.29}  
\resdef{h2.s1.final.dprime_wait.hi}{1.44}  
\resdef{h2.s1.final.dprime_wait.lo}{1.16}  
\resdef{ctl.zr.dacc_median}{42.9}  
\resdef{ctl.zr.orz32.base.acc}{10.8}  
\resdef{ctl.zr.orz32.final.acc}{91.4}  
\resdef{h1.bb_deepscaler.final.fin.r2}{4.07}  
\resdef{h1.bb_gr3_14.final.r2}{0.04}  
\resdef{h1.bb_gr3_4_s1.final.r2}{0.18}  
\resdef{h1.bb_gr3_4_s2.final.r2}{0.15}  
\resdef{h1.bb_gr3_4_s3.final.r2}{0.14}  
\resdef{h1.bb_gr3_4m.final.r2}{0.37}  
\resdef{h1.bb_light14.final.fin.r2}{3.21}  
\resdef{h1.bb_pangu_e7m.final.fin.r2}{1.32}  
\resdef{h1.bb_pangu_e7v11.final.fin.r2}{0.41}  
\resdef{h1.bb_q3r1.step0100.r2}{0.08}  
\resdef{h1.bb_q3r1.step0400.r2}{0.03}  
\resdef{h1.bb_q3r1_base.base.r2}{0.02}  
\resdef{h1.bb_qwen35_0p8b.final.fin.r2}{0.90}  
\resdef{h1.bb_qwen35_0p8b_base.final.fin.r2}{2.21}  
\resdef{h1.bb_qwen35_0p8b_base.final.r2}{1.96}  
\resdef{h1.bb_qwen35_2b.final.fin.r2}{0.72}  
\resdef{h1.bb_qwen35_2b_base.final.fin.r2}{1.39}  
\resdef{h1.bb_qwen35_4b.final.fin.r2}{1.68}  
\resdef{h1.bb_qwen35_4b_base.final.fin.r2}{1.18}  
\resdef{h1.bb_qwen35_9b.final.fin.r2}{1.43}  
\resdef{h1.bb_qwen35_9b_base.final.fin.r2}{0.97}  
\resdef{h1.bb_qwen35_9b_base.final.r2}{0.93}  
\resdef{h1.bb_qwen38_27b.final.r2}{0.91}  
\resdef{h1.bb_r1d14.final.fin.r2}{2.79}  
\resdef{h1.bb_r1d1p5b.final.fin.r2}{3.65}  
\resdef{h1.ds_gsm8k.s1.base.fin.r2}{4.62}  
\resdef{h1.ds_gsm8k.s1.final.fin.r2}{17.10}  
\resdef{h1.ds_mmlupro.s1.base.fin.r2}{0.03}  
\resdef{h1.ds_mmlupro.s1.final.fin.r2}{0.22}  
\resdef{h1.seed.D512.sft.r2}{0.03}  
\resdef{h1.seed.D512.step0400.r2}{$<$.01}  
\resdef{h1.seed.D64.sft.r2}{0.03}  
\resdef{h1.seed.D64.step0400.r2}{0}  
\resdef{h1.seed.W512.sft.r2}{7.52}  
\resdef{h1.seed.W512.trend.a_nextopen.r2}{4.83}  
\resdef{h1.seed.W512LY.sft.fin.r2}{7.45}  
\resdef{h1.seed.W512LY.step0400.fin.r2}{$<$.01}  
\resdef{h1.seed.W64.sft.r2}{1.60}  
\resdef{h1.seed.X512.trend.a_nextopen.r2}{0.12}  
\resdef{h1.seed.X512LY.sft.r2}{0.02}  
\resdef{h1.zr.ce.gr3_4.hi}{0.34}  
\resdef{h1.zr.ce.gr3_4.lo}{0.24}  
\resdef{h1.zr.ce.gr3_4.rate}{0.29}  
\resdef{h1.zr.families}{Qwen2.5, Mistral and Qwen3}  
\resdef{h1.zr.final_median}{0}  
\resdef{h1.zr.n_bases}{9}  
\resdef{h1.zr.n_confirm}{17}  
\resdef{h1.zr.n_counter}{1}  
\resdef{h1.zr.n_runs}{18}  
\resdef{h1.zr.orz32.base.ntok}{424{,}454}  
\resdef{h1.zr.orz32.base.nwait}{0}  
\resdef{h1.zr.orz32.final.ntok}{2{,}379{,}766}  
\resdef{h1.zr.orz32.final.nwait}{4}  
\resdef{h1.zr.size_max}{32B}  
\resdef{h1.zr.size_min}{0.5B}  
\resdef{h1.zr.upper95_ceil}{0.06}  
\resdef{sens.s1.greedy.ratio}{1.55}  
\resdef{sens.s1.greedy_sent.ratio}{1.53}  
\resdef{sens.s1.seedbase.max}{5.4}  
\resdef{sens.s1.seedbase.min}{4.9}  
\resdef{sens.s1.seedbase.n}{6}  
\resdef{sens.s1.sent.ratio}{1.43}  
\resdef{sens.zr.nogate.n_confirm}{22}  
\resdef{sens.zr.nogate.n_runs}{23}  
\resdef{sens.zr.point.n_confirm}{17}  
\resdef{sens.zr.point.n_runs}{18}  
\resdef{sens.zr.reg.n_confirm}{17}  
\resdef{sens.zr.reg.n_runs}{18}  
\resdef{sens.zr.released.n_confirm}{13}  
\resdef{sens.zr.released.n_runs}{14}  
\resdef{sens.zr.t10.n_confirm}{5}  
\resdef{sens.zr.t10.n_runs}{5}  
\resdef{sens.zr.thr02.n_confirm}{15}  
\resdef{sens.zr.thr02.n_runs}{17}  
\resdef{sens.zr.thr10.n_confirm}{17}  
\resdef{sens.zr.thr10.n_runs}{18}  
\resdef{sens.zr.withsec.n_confirm}{23}  
\resdef{sens.zr.withsec.n_runs}{24}  

\draftfalse

\title{Inherited, Not Invented: Tracing the Provenance of \\ \emph{Wait} in Reasoning Models} 

\author{
\bfseries Zhizhuo Kou\textsuperscript{1,3} \quad Sizai Hou \textsuperscript{1} \quad
Long Chen \textsuperscript{1} \quad Jian Yang \textsuperscript{1} \quad Junyu Luo \textsuperscript{2} \\
\bfseries Jingshu Peng \textsuperscript{1}
\quad Wenkai Zhang \textsuperscript{3} \quad Sirui Han\textsuperscript{1}\NoHyper\footnotemark[1]\endNoHyper \quad Yike Guo\textsuperscript{1} \\[2pt]
\normalfont\textsuperscript{1}The Hong Kong University of Science and Technology \\
\normalfont\textsuperscript{2}Peking University \quad
\textsuperscript{3}Huawei Noah's Ark Lab \\[2pt]
\normalfont\small\texttt{zkouaa@connect.ust.hk, siruihan@ust.hk}
}
\begin{document}
\maketitle

\begin{abstract}
Reinforcement learning (RL) is credited with teaching reasoning models to write \emph{wait} before reconsidering a step, yet large RL accuracy gains do not require the word. This view can confuse what RL discovers with what a model inherits from its training data. We trace the word through open models' training histories, reading it with label-free indices of frequency, position, form and usage and an injected-error test, all computed from sampled outputs. We find that zero-RL almost never coins the marker: on bases that lack the word, RL raises MATH-500 accuracy substantially while \emph{wait} stays near absent. It arrives with training data: mid-training in OLMo-3, and fine-tuning on reasoning traces installs it in most models whose base lacks it. Where the base already uses it, both released RL-Zero recipes (OLMo-3, MiMo) multiply it, whereas KL-free GRPO from the same base does not; OLMo-3's recipe wears down its form and makes it a more sensitive but more liberal alarm; RL after supervised fine-tuning barely changes it, and our GRPO recipe erases a marker seeded by fine-tuning. Finally, we show that marker frequency alone does not indicate RL-induced reasoning and caution against LLM-as-a-judge revision metrics, as automated judges can reverse human-labelled trends.


\end{abstract}

\section{Introduction}
\label{sec:intro}

\begin{figure}[t]
  \centering
  \includegraphics[width=\linewidth]{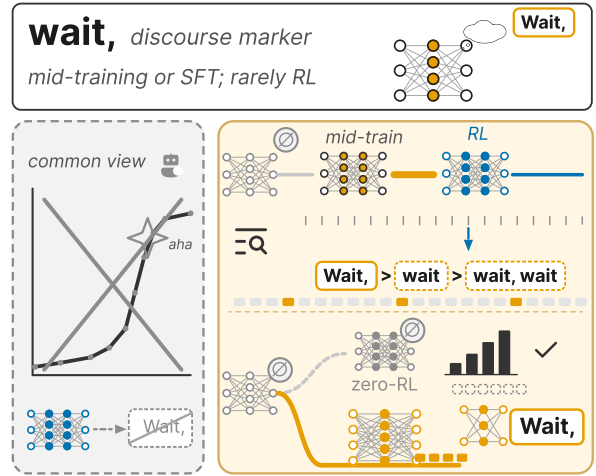}
  \caption{\textbf{RL rarely coins \form{wait}: the marker enters with mid-training or fine-tuning on traces.}}
  \label{fig:teaser}
  \vspace{-18pt}
\end{figure}

Revision markers have emerged as the most visible sign of reasoning in large language models (LLMs). For example, when asked a competition problem, OLMo-3 writes \emph{``Wait, is that correct?''} before it re-checks a step; after RL, the same model writes \emph{``Wait, but hold on.''} and \emph{``Wait, 2 is 2/1.''} among \res{fig1.case.last.nwait} occurrences of the word in one solution. In the R1-Zero run, \form{wait} and its kin become several times more frequent, prompting reports of an ``aha moment'' in which the model learns to re-evaluate \citep{deepseekai2025r1}. Consequently, uncertainty heuristics count these markers as doubt \citep{pandey2026selfdoubt,vanhoyweghen2025lexical}, and chain-of-thought monitors process traces full of them \citep{korbak2025chain,baker2025monitoring}.

While recent studies track reflective behaviours during RL runs, they leave the true origin of revision markers untested. Existing studies count reflective keywords or judged behaviours \citep{hu2025openreasonerzero,zeng2025simplerlzoo}, and critiques of the ``aha moment'' show that reflective \emph{behaviour} already exists in base models \citep{liu2025understanding,essentialai2025rethinking,gandhi2025cognitive}. However, they rest on \textbf{an untested assumption that RL explicitly coins the revision marker and fixes its semantic meaning throughout training}. Neither assumption has been tested along open training trajectories that start from a base model before RL. Where the \emph{word} comes from, and what RL does to it, remains open. We therefore ask: \emph{where does \form{wait} come from, and what does RL do to it?} Leveraging open releases, we track the pretraining and mid-training checkpoints of open source model (eg. OLMo-3 \citep{olmo2025olmo}), their RL trajectories, zero-RL runs from bases that do not use the marker, and descendants of one base trained on reasoning traces. At the analytical level, we construct indices of frequency, position, form, and usage, alongside an injected-error test of what the marker signals. We evaluate this history exactly as a historical linguist reads a word, from its occurrences alone.

Our analysis shows in all traced lineages but one, the revision marker originates in pre-RL training data, drawing upon grammaticalization \citep{hopper2003grammaticalization,lehmann2015thoughts,bybee2003mechanisms} (\Cref{fig:teaser}). In OLMo-3 marker is absent ($\varnothing$, below our 0.05 per 1k token threshold) after pretraining and appears during mid-training, RL-Zero wears down its form (\tok{Wait,} $>$ \tok{wait} $>$ \tok{wait, wait}). Zero-RL across \res{h1.zr.families} bases (\res{h1.zr.size_min}--\res{h1.zr.size_max}) almost never produces it, confirming that RL alone rarely generates the marker.


Our contributions are fourfold:
\ding{182} \textbf{RL rarely coins the marker; training data supplies it.} Zero-RL from bases without \form{wait} raises accuracy by a median of \res{ctl.zr.dacc_median} points and keeps the marker below 0.05 per 1k tokens in \res{h1.zr.n_confirm}/\res{h1.zr.n_runs} runs across \res{h1.zr.n_bases} \res{h1.zr.families} bases. The marker enters OLMo-3 mid-training, and \res{h1.prov.sft_free.n_ge05}/\res{h1.prov.sft_free.n} models fine-tuned on traces from a non-marking base adopt it.
\ding{183} \textbf{RL reshapes an inherited marker differently across recipes.} Both RL-Zero recipes (OLMo-3, MiMo) multiply the marker, whereas KL-free GRPO from the same base does not. OLMo-3's recipe turns its form into a more sensitive yet liberal alarm. SFT followed by RL changes the marker rate little, and our GRPO recipe erases a seeded marker.
\ding{184} \textbf{A label-free instrument tracks the marker.} Frequency, position, form, and usage indices with an injected-error alarm test read a marker without labels, judges, or internal access, making it portable across open lineages. An LLM judge, by contrast, reverses human revision trends.
\ding{185} \textbf{Provenance maps and blind audits provide reusable resources.} We provide a provenance map of \res{h1.prov.n_rows} model endpoints with lineages and marker indices, plus blind revision audits with expert consensus.

\section{Background and Related Work}
\label{sec:background}
\label{sec:related:main}

\paragraph{Pre-RL origin of reflective markers.}
Outcome-based RL with verifiable rewards trains a model on its own sampled solutions and rewards only the final answer \citep{shao2024deepseekmath,deepseekai2025r1}. Technical reports have described reflection emerging during these training runs, measured by counting words like \form{wait}, \form{recheck} and \form{verify} \citep{deepseekai2025r1,hu2025openreasonerzero,zeng2025simplerlzoo}. Two lines of work qualify this timeline: base models exhibit reflective behaviour before RL \citep{liu2025understanding,essentialai2025rethinking,yeo2025demystifying,gandhi2025cognitive}, and verifiable rewards mostly sharpen this existing capability \citep{yue2025does,zhao2025echo,shao2025spurious}; mid-training dictates RL responsiveness \citep{wang2025octothinker,zhang2025interplay}, and distilled models use hesitation words far more often than zero-RL models \citep{hu2025why}. While these works have achieved commendable progress in locating the \emph{behaviour} before RL, \textbf{the developmental origin of the marker form remains open}. We date the arrival of the \emph{form} within a single open lineage and track it through RL from bases with and without it.

\paragraph{Functional interpretation and temporal dynamics.}
The functional interpretation of linguistic markers in final models remains actively disputed. Post-answer reflections are mostly confirmatory \citep{kang2025first}, many reasoning steps are decorative \citep{zhao2025aha}, and suppressing marker tokens costs RL-trained models little \citep{wang2025wait}, yet inflections occur where probes register belief shifts \citep{boppana2026reasoning}. Interventions have tested these tokens by deleting, appending or inserting markers \citep{qian2025demystifying,muennighoff2025s1}. Chain-of-thought monitors and uncertainty heuristics treat these words as evidence of reasoning \citep{korbak2025chain,baker2025monitoring,pandey2026selfdoubt}, while behaviour labels often rely on LLM judges of uneven validity \citep{zheng2023judging,bavaresco2025llms}. These studies examine final models only, without the axis of training time. We introduce this temporal dimension, identifying where a word enters a model's history and measuring how each training stage alters its meaning for future model deployment.

\paragraph{Grammaticalization as a tracking framework.}
Grammaticalization provides a measurable framework from historical linguistics to explain how a word becomes a grammatical marker \citep{hopper2003grammaticalization}. The process requires a \emph{source} construction already in use; it then raises the marker's frequency \citep{bybee2003mechanisms}, fixes its position, wears down its form (attrition) and binds it to neighbouring material \citep{lehmann2015thoughts}. Researchers can measure these parameters directly from occurrences, just as lexical-semantic-change methods quantify change from usage \citep{hamilton2016diachronic,giulianelli2020analysing}, making them suitable for tracking model checkpoints. Language also changes under optimisation in learning agents: communication optimised for reward drifts from natural language \citep{lewis2017deal,lazaridou2020multi}, and o3 has developed hard-to-read terms that grew more frequent over RL \citep{schoen2025stress}. We using these grammaticalization parameters as a rigorous vocabulary and construct a set of predictions fixed before the main run. 

\section{Method}
\label{sec:method}

\begin{figure*}[t]
  \centering
  \includegraphics[width=\textwidth]{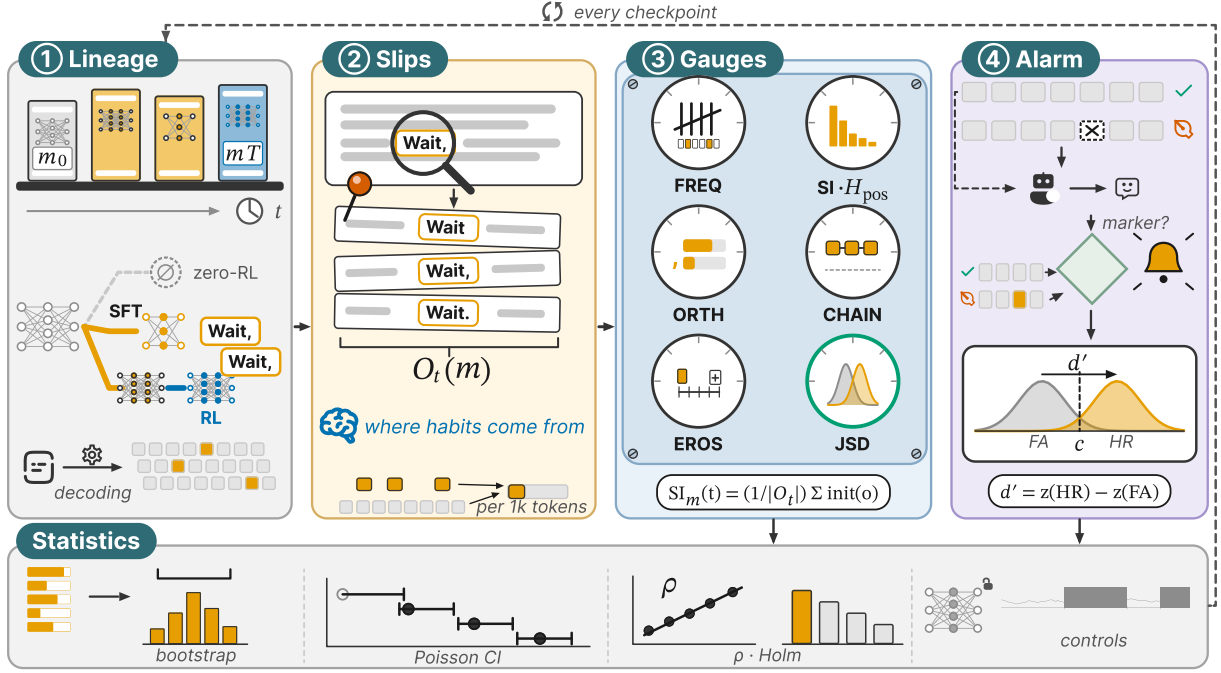}
  \caption{\textbf{Our methodology extracts a marker's evolutionary history from its occurrences, one checkpoint at a time.} The pipeline processes \circled{1}~lineages and sampled traces into \circled{2}~matched occurrences, from which we construct \circled{3}~label-free indices and validate the \circled{4}~injected-error alarm test, with statistics at every checkpoint (bottom).}
  \label{fig:method}
\end{figure*}

To analyse language evolution during training, we treat the samples of one model across its checkpoints as a diachronic corpus. We construct a marker's history from its occurrences through the four stages introduced in \Cref{fig:method}.

\paragraph{Reasoning lineages}
To isolate the origins of reasoning markers, we analyse three distinct classes of model history.  (i) Fully open lineages with all stages released: OLMo-3 7B from pretraining through mid-training to base \citep{olmo2025olmo}, followed by RL-Zero or SFT/preference/RL tuning (OLMo-3 Think; 32B replication: OLMo-3.1 Think). (ii) RL-Zero runs from marker-free bases: our GRPO on Qwen2.5-Math (1.5B with two seeds, 7B), plus DAPO \citep{yu2025dapo} and Open-Reasoner-Zero \citep{hu2025openreasonerzero} on Qwen2.5-32B. (iii) Fine-tuned descendants on reasoning traces from Qwen2.5-32B (QwQ-32B, R1-Distill), Phi-4, Qwen3, Llama, GLM, MiMo, and SmolLM3. \Cref{app:models,app:methoddetails} detail all \res{h1.prov.n_rows} endpoints and recipes. Each checkpoint evaluates MATH-500 problems \citep{hendrycks2021measuring,lightman2024lets} using fixed decoding per series, sampling 2--8 responses per problem.

\paragraph{Occurrence matching.}
Let $O_t(m)$ denote the occurrences of marker $m$ in the traces of checkpoint $\ckpt{t}$. To ensure robust quantification, we match these occurrences on detokenised text across every spelling variant (\tok{Wait,}, \tok{wait}, \tok{Wait.}, \dots). We define the marker frequency $\mathrm{FREQ}_m(t)$ as $|O_t(m)|$ per 1k generated tokens. Per-response and per-word rates consistently produce the same ordering (\Cref{sec:exp:source}).

\paragraph{Label-free linguistic indices.}
At the linguistic level, we extract the parameters of \Cref{sec:background} solely from $O_t(m)$. \emph{Fixation} represents the sentence-initial share:
\begin{equation}
  \mathrm{SI}_m(t) = \frac{1}{|O_t(m)|}\sum_{o \in O_t(m)} \mathrm{init}(o),
  \label{eq:si}
\end{equation}
where $\mathrm{init}(o) = 1$ if $o$ starts its sentence, complemented by positional entropy $H_{\mathrm{pos}}$. \emph{Form} (ORTH) measures the canonical \tok{Wait,} share. \emph{Bonding} (CHAIN) calculates the proportion continuing a reduplicated chain (\tok{Wait, wait}). \emph{Erosion} (EROS) tracks token distance to the next mathematical action and sentence length. Finally, \emph{usage drift} computes Jensen--Shannon divergence between marker usage-cluster distributions at checkpoint $t$ versus the base using a fixed encoder \citep{giulianelli2020analysing,kutuzov2020uiouva}, benchmarked against the base split-half noise floor.

\paragraph{Injected-error alarm test.}
To determine what \form{wait} signals, we construct a prefix of correct steps and a parallel copy with one injected arithmetic error for each problem. We continue both prefixes with $\ckpt{t}$ and record whether a revision marker appears within two sentences. The registered outcome counts any form of the revision paradigm (\form{wait}, \form{hmm}, \form{actually}, \dots; \Cref{app:methoddetails}); we report \form{wait} alone beside it. Given the hit rate $\mathrm{HR} = \Prob(\text{marker} \mid \text{error})$ and the false-alarm rate $\mathrm{FA} = \Prob(\text{marker} \mid \text{no error})$, we compute signal detection metrics:
\begin{equation}
\begin{aligned}
  d'(t) &= z(\mathrm{HR}) - z(\mathrm{FA}),\\
  c(t) &= -\tfrac12\big[z(\mathrm{HR}) + z(\mathrm{FA})\big],
\end{aligned}
  \label{eq:sdt}
\end{equation}
where $d'$ measures how effectively the marker separates errors from correct steps and $c$ measures how readily it fires \citep{green1966signal,macmillan2004detection}. The marker indices and the alarm test operate strictly on deterministic rules, requiring no human or LLM judgment of behaviour.

\paragraph{Portability \& Statistical design.}
The indices and the alarm test need only sampled outputs and the list of marker forms. We apply them unchanged to \res{h1.prov.n_rows} model endpoints, from pretraining checkpoints to released reasoning models, and the code, the provenance map and the blind audits are linked in the abstract (\Cref{app:licenses}).

\paragraph{\textbf{Statistics.}} Exact Poisson intervals treat occurrences as independent; ratios such as $A$ and the alarm's $d'$ use a cluster bootstrap over problems, which absorbs clustering within responses. The injection protocol and the usage encoder are specified in \Cref{app:operationalization,app:details}.
All reported intervals represent 95\% cluster-bootstrap intervals over problems. Rates of rare events carry Poisson-model intervals. We distinguish the registered near-absence threshold (0.05 per 1k tokens on the point estimate) from these intervals. Trends use Spearman correlations with the RL step, Holm-corrected over the primary indices. To maintain exactness, we never pool series. The open series originate from single training runs, so their intervals strictly cover evaluation sampling only. \Cref{app:methoddetails,app:stats} provide further details, including the judge-based indices and the human annotation audits.

\section{Main Experiments Results}
\label{sec:exp}

\paragraph{Zero-RL rarely coins the marker.}
\begin{table}[t]
  \centering\tabstyle\footnotesize
  \fitwidth{\begin{tabular}{@{}lcc@{}}
    \toprule
    \textbf{Training route} & \textbf{Write it}  & \textbf{Median per 1k} \\
    \midrule
    Zero-RL from a base without it & \res{h1.zr.n_counter} of \res{h1.zr.n_runs}  & \res{h1.zr.final_median} \\
    Fine-tuned on traces, base without it & \res{h1.route.sftfree.n_ge05} of \res{h1.route.sftfree.n}  & \res{h1.route.sftfree.median} \\
    Post-trained on a base that writes it & \res{h1.route.postmarked.n_ge05} of \res{h1.route.postmarked.n}  & \res{h1.route.postmarked.median} \\
    RL on a model that writes it & \res{h1.route.rlmarked.n_ge05} of \res{h1.route.rlmarked.n}  & \res{h1.route.rlmarked.median} \\
    Base not released & \res{h1.route.nobase.n_ge05} of \res{h1.route.nobase.n}  & \res{h1.route.nobase.median} \\
    \bottomrule
  \end{tabular}}
  \caption{\textbf{Zero-RL rarely yields \form{wait}; every other route to it passes through training data.} Models that write it after training ($\ge$0.05 per 1k tokens)}
  \label{tab:routes}
  \vspace{-12pt}
\end{table}

\label{sec:exp:source}

\begin{table*}[t]
  \centering
  \tabstyle
  \setlength{\tabcolsep}{2.5pt}
  \fitwidth{\begin{tabular}{@{}ll cc cc@{}}
    \toprule
    & & \multicolumn{2}{c}{\textbf{MATH-500 (\%)}} & \multicolumn{2}{c}{\textbf{\form{wait} per 1k tokens}} \\
    \cmidrule(lr){3-4} \cmidrule(lr){5-6}
    \textbf{Model} & \textbf{Training} & before & after & before & after \\
    \midrule
    \multicolumn{6}{@{}l}{\textsc{zero-rl from a base without the marker}} \\
    Open-Reasoner-Zero-32B & PPO & 10.8 & 91.4 & 0\ci{0}{$<$.01} & \nsig{$<$.01\ci{$<$.01}{$<$.01}} \\
    Qwen2.5-Math-7B & GRPO  & 16.9 & 76.0 & $<$.01\ci{$<$.01}{$<$.01} & \nsig{0\ci{0}{$<$.01}} \\
    REF-Qwen3-4B  & GRPO  & 41.2 & 84.1 & 0.02\ci{0.01}{0.03} & \cellcolor{cCmpA!2}\nsig{0.03\ci{0.02}{0.04}} \\
    Mistral-Small-24B-SimpleRL-Zoo & GRPO & 21.4 & 75.8 & 0\ci{0}{0.01} & \nsig{$<$.01\ci{$<$.01}{$<$.01}} \\
    General-Reasoner-Qwen3-4B (exception) & GRPO & 71.2 & 79.0 & $<$.01\ci{$<$.01}{0.01} & \cellcolor{cCmpA!10}0.29\ci{0.24}{0.34}\,\textcolor{cInterv}{$\uparrow$} \\
    \addlinespace[0.5mm]\midrule
    \multicolumn{6}{@{}l}{\textsc{fine-tuned on reasoning traces}} \\
    DeepSeek-R1-Distill-Qwen-32B & SFT & 35.8 & 89.5 & $<$.01\ci{$<$.01}{0.02} & \cellcolor{cCmpA!25}3.25\ci{3.18}{3.32}\,\textcolor{cInterv}{$\uparrow$} \\
    Phi-4-reasoning & SFT & 82.0 & 93.0 & 0\ci{0}{$<$.01} & \cellcolor{cCmpA!20}1.56\ci{1.50}{1.62}\,\textcolor{cInterv}{$\uparrow$} \\
    MiMo-7B-SFT & SFT & 65.8 & 85.6 & 4.95\ci{4.86}{5.04} & \cellcolor{cCmpA!26}3.77\ci{3.69}{3.84}\,\textcolor{cOurs}{$\downarrow$} \\
    \addlinespace[0.5mm]\midrule
    \multicolumn{6}{@{}l}{\textsc{qwen3: sft and rl on its own base (before = its base)}} \\
    Qwen3-8B & SFT+RL & 66.5 & 84.9 & 0\ci{0}{$<$.01} & \cellcolor{cCmpA!27}3.93\ci{3.87}{4.00}\,\textcolor{cInterv}{$\uparrow$} \\
    Qwen3-14B & SFT+RL & 66.5 & 87.9 & 0.03\ci{0.01}{0.05} & \cellcolor{cCmpA!25}3.03\ci{2.98}{3.10}\,\textcolor{cInterv}{$\uparrow$} \\
    Qwen3-30B-A3B & SFT+RL & 52.2 & 86.8 & 0.02\ci{0.01}{0.03} & \cellcolor{cCmpA!24}2.83\ci{2.77}{2.88}\,\textcolor{cInterv}{$\uparrow$} \\
    \addlinespace[0.5mm]\midrule
    \multicolumn{6}{@{}l}{\textsc{training stages of olmo-3-1025-7b (before = previous stage)}} \\
    Pretraining & stage 1 & 0.0 & 0.1 & 0.04\ci{0.03}{0.05} & \cellcolor{cCmpA!3}\nsig{0.04\ci{0.03}{0.06}} \\
    Mid-training & stage 2 & 0.1 & 35.7 & 0.04\ci{0.03}{0.06} & \cellcolor{cCmpA!27}4.03\ci{3.92}{4.14}\,\textcolor{cInterv}{$\uparrow$} \\
    Base & stage 3 & 35.7 & 51.2 & 4.03\ci{3.92}{4.14} & \cellcolor{cCmpA!28}4.94\ci{4.85}{5.03}\,\textcolor{cInterv}{$\uparrow$} \\
    \addlinespace[0.5mm]\midrule
    \multicolumn{6}{@{}l}{\textsc{rl from a model that uses the marker}} \\
    Olmo-3-7B-RL-Zero-Math & RL-Zero & 52.3\std{1.2} & 84.9\std{0.9} & 5.07\std{0.17} & \cellcolor{cCmpA!35}\best{12.68}\std{1.55}\,\textcolor{cInterv}{$\uparrow$} \\
    Olmo-3-7B-Think & SFT, DPO, RL & 90.0 & 91.0 & 3.86\ci{3.79}{3.94} & \cellcolor{cCmpA!27}4.52\ci{4.44}{4.59}\,\textcolor{cInterv}{$\uparrow$} \\
    Olmo-3.1-32B-Think & SFT, DPO, RL & 91.4 & 93.0 & 4.52\ci{4.46}{4.58} & \cellcolor{cCmpA!27}4.23\ci{4.17}{4.29}\,\textcolor{cOurs}{$\downarrow$} \\
    MiMo-7B-RL-Zero & RL-Zero & 65.8 & 76.4 & 4.95\ci{4.86}{5.04} & \cellcolor{cCmpA!30}6.62\ci{6.53}{6.71}\,\textcolor{cInterv}{$\uparrow$} \\
    MiMo-7B-RL & SFT, then RL & 85.6 & 87.6 & 3.77\ci{3.69}{3.84} & \cellcolor{cCmpA!30}\second{6.65}\ci{6.56}{6.74}\,\textcolor{cInterv}{$\uparrow$} \\
    \addlinespace[0.5mm]\midrule
    \multicolumn{6}{@{}l}{\textsc{newest generation: qwen3.5 and qwen3.8 (before = its base)}} \\
    Qwen3.5-0.8B & post-training & 43.8 & 48.1 & 1.96\ci{1.92}{1.99} & \cellcolor{cCmpA!21}1.83\ci{1.80}{1.86}$^\ddagger$\,\textcolor{cOurs}{$\downarrow$} \\
    Qwen3.5-2B & post-training & 74.5 & 76.8 & 1.31\ci{1.27}{1.34} & \cellcolor{cCmpA!16}0.86\ci{0.84}{0.89}\,\textcolor{cOurs}{$\downarrow$} \\
    Qwen3.5-4B & post-training & 89.6 & 88.2 & 1.15\ci{1.11}{1.20} & \cellcolor{cCmpA!26}3.82\ci{3.78}{3.85}$^\ddagger$\,\textcolor{cInterv}{$\uparrow$} \\
    Qwen3.5-9B & post-training & 88.1 & 92.1 & 0.93\ci{0.89}{0.98} & \cellcolor{cCmpA!24}2.94\ci{2.91}{2.97}\,\textcolor{cInterv}{$\uparrow$} \\
    \bottomrule
  \end{tabular}}
  \caption{\textbf{Zero-RL rarely installs \form{wait}; training data does, and released RL-Zero recipes amplify it.} Largest (bold), second increase (underline), intervals overlap (gray); $\pm$: s.d.\ over six decoding seeds, else [95\% CI]. Shading: \form{wait} per 1k after training (log scale, darker = more); \textcolor{cInterv}{$\uparrow$}/\textcolor{cOurs}{$\downarrow$}: rise/fall with non-overlapping intervals.}
  \label{tab:main}
  \vspace{-12pt}
\end{table*}

Large RL accuracy gains occur without frequent use of \form{wait}. Across \res{h1.zr.n_runs} zero-RL runs from bases that lack it, accuracy rises by a median of \res{ctl.zr.dacc_median} points, and in \res{h1.zr.n_confirm} of them the marker's 95\% upper bound stays below \res{h1.zr.upper95_ceil} per 1k tokens. \Cref{tab:routes} groups models by documented training recipe, while \Cref{tab:main,tab:nobase} report the main values. All \res{h1.prov.n_rows} endpoints, including instruct models, further bases and models with undisclosed training (\Cref{fig:major,fig:hero:full,tab:provenance}).

\textbf{Zero-RL from marker-free bases maintains marker absence.} Qwen2.5-32B evaluated with its own prompt template, Open-Reasoner-Zero raises MATH-500 accuracy from \res{ctl.zr.orz32.base.acc}\% to \res{ctl.zr.orz32.final.acc}\% (the base scores low under this template); \form{wait} occurs \res{h1.zr.orz32.final.nwait} times in \res{h1.zr.orz32.final.ntok} sampled tokens (base: \res{h1.zr.orz32.base.nwait} in \res{h1.zr.orz32.base.ntok}). Our GRPO runs on Qwen2.5-Math raise accuracy from \res{ctl.s7r7b.base.acc}\% to \res{ctl.s7r7b.final.acc}\% at 7B and to \res{ctl.s7r1.final.acc}\% and \res{ctl.s7r2.final.acc}\% at 1.5B, and their last checkpoints contain \res{h1.s7r7b.final.nwait}, \res{h1.s7r1.final.nwait} and \res{h1.s7r2.final.nwait} occurrences. Released zero-RL models from other recipes (SimpleRL-Zoo, Open-Reasoner-Zero, Oat-Zero) agree wherever our evaluation reproduces their published accuracy (\Cref{tab:provenance}). In \res{h1.zr.n_confirm} of these \res{h1.zr.n_runs} runs, from \res{h1.zr.n_bases} \res{h1.zr.families} bases, the exact 95\% upper bound stays below \res{h1.zr.upper95_ceil}/1k tokens (\Cref{tab:routes}). 

None of the other reflection openers in our inventory (\form{hmm}, \form{hold on}, \form{let me check}, \dots) takes its place (\Cref{app:results}). General-Reasoner on Qwen3-4B-Base \citep{ma2025generalreasoner} is the exception: it reproduces its published accuracy, its base has \res{h1.bb_gr3_4_base.final.nwait} occurrence of \form{wait} in \res{h1.bb_gr3_4_base.final.ntok} tokens, while the RL model writes it \res{h1.zr.ce.gr3_4.rate} times per 1k (95\% CI [\res{h1.zr.ce.gr3_4.lo}, \res{h1.zr.ce.gr3_4.hi}]), mostly as sentence-initial self-corrections such as \emph{``Wait, there's a discrepancy here. Let's re-examine.''} The marker persists under other decodes: \res{h1.bb_gr3_4m.final.r2} per 1k at temperature 0.6 with four samples per problem, and \res{h1.bb_gr3_4_s1.final.r2}, \res{h1.bb_gr3_4_s2.final.r2} and \res{h1.bb_gr3_4_s3.final.r2} per 1k under three sampling seeds. The same recipe on Qwen3-14B-Base stays below the threshold (\res{h1.bb_gr3_14.final.r2} per 1k). We hypothesize that the marker was latent in this base: the Qwen3 report describes a pretraining stage with a raised share of STEM, coding, reasoning and synthetic data \citep[\S3.2]{yang2025qwen3}, leaving it unclear whether reasoning traces were included. RL from marker-free instruct models does not produce the marker either (below 0.01 per 1k at all \res{h1.prov.rl_instruct.n} checkpoints). The 7B run was registered as a replication before it ran. The pattern holds beyond MATH-500 (\Cref{tab:datasets}): on AIME24, AIME25, AMC23, GSM8K and MMLU-Pro, zero-RL from Qwen2.5-7B never writes \form{wait} more than \res{h1.ds_gsm8k.szq7.final.nwait} times. OLMo-3 RL-Zero raises it on every benchmark, from \res{h1.ds_gsm8k.s1.base.fin.r2} to \res{h1.ds_gsm8k.s1.final.fin.r2} per 1k on GSM8K; only under the official few-shot prompt of MMLU-Pro does it barely write the marker (\res{h1.ds_mmlupro.s1.base.fin.r2} to \res{h1.ds_mmlupro.s1.final.fin.r2}/ 1k).

\begin{table*}[t]
  \centering
  \tabstyle
  \fitwidth{\begin{tabular}{@{}l ccccc@{}}
    \toprule
    \textbf{Model (before $\rightarrow$ after)} & \textbf{AIME24} & \textbf{AIME25} & \textbf{AMC23} & \textbf{GSM8K} & \textbf{MMLU-Pro} \\
    \midrule
    OLMo-3 7B RL-Zero & 5.03 $\rightarrow$ 7.37 & 5.23 $\rightarrow$ 7.63 & 5.49 $\rightarrow$ 9.78 & 4.62 $\rightarrow$ 17.10 & 0.03 $\rightarrow$ 0.22 \\
    OLMo-3 7B Think & 4.30 $\rightarrow$ 4.90 & 4.67 $\rightarrow$ 5.22 & 3.68 $\rightarrow$ 4.33 & 2.95 $\rightarrow$ 3.32 & 1.61 $\rightarrow$ 2.34 \\
    Qwen2.5-7B $\rightarrow$ SimpleRL-Zoo & \nsig{$<$.01 $\rightarrow$ 0} & \nsig{$<$.01 $\rightarrow$ 0} & \nsig{$<$.01 $\rightarrow$ 0} & \nsig{0.01 $\rightarrow$ $<$.01} & \nsig{0 $\rightarrow$ 0} \\
    R1-Distill-Qwen-7B & 4.75 & 6.04 & 3.93 & 3.37 & 4.10 \\
    \bottomrule
  \end{tabular}}
  \caption{\textbf{Zero-RL from Qwen2.5-7B leaves \form{wait} absent on all five benchmarks; models that inherited it keep it on the mathematical ones.} Rate per 1k tokens of finished responses, before $\rightarrow$ after training; grey: 95\% intervals overlap; MMLU-Pro: official few-shot prompt.}
  \label{tab:datasets}
\end{table*}

\textbf{Training data supplies the marker.} \res{h1.prov.sft_free.n_ge05} of \res{h1.prov.sft_free.n} models fine-tuned on reasoning traces from marker-free bases write the marker $\ge 0.5$ times/1k tokens (\Cref{tab:routes}). Another \res{h1.prov.sft_marked.n} start from marked bases, while \res{h1.prov.sft_free.n_basemissing} lack measured bases (e.g., Qwen2.5-32B R1-Distill \res{h1.s8r1d.final.freq2}, QwQ-32B \res{h1.s8qwq.final.freq2}, Qwen3-8B \res{h1.pop_qwen3_8b.final.freq2}, Phi-4-reasoning \res{h1.pop_phi4r.final.freq2}, and R1-Distill-Llama-8B \res{h1.pop_r1d_llama8.final.freq2}). 
An exception, OpenReasoning-Nemotron-7B \citep{nvidia2025openreasoning}, opens sentences with \form{but} (\res{h1.bb_orn7.final.open_but}/1k) and \form{perhaps} (\res{h1.bb_orn7.final.open_perhaps}/1k) instead of \form{wait} (\res{h1.bb_orn7.final.open_wait}/1k), showing marker forms track trace transfers. In OLMo-3 (\Cref{fig:origin}a), \form{wait} occurs $\le \res{h1.s1pre.stage1-step1413814.freq2}$ times per 1k in pretraining, but rises post-mid-training to \res{h1.s1pre.stage2-step47684.freq2} (\res{h1.s1pre.stage2-step47684.si}\% sentence-initial), coinciding with thinking traces included for RL-Zero \citep{olmo2025olmo}. 
Pretrained bases whose mixtures include reasoning data already write it, including MiMo-7B-Base (\res{h1.bb_mimo_base.final.freq2}/1k) and SmolLM3-3B-Base (\res{h1.bb_smollm3_base.final.freq2} per 1k) \citep{xiaomi2025mimo,bakouch2025smollm3}.

All four Qwen3.5 bases write \form{wait} (\res{h1.bb_qwen35_9b_base.final.r2}/1k at 9B to \res{h1.bb_qwen35_0p8b_base.final.r2}/1k at 0.8B); post-training decreases frequency at 0.8B (\res{h1.bb_qwen35_0p8b_base.final.fin.r2} to \res{h1.bb_qwen35_0p8b.final.fin.r2}) and 2B (\res{h1.bb_qwen35_2b_base.final.fin.r2} to \res{h1.bb_qwen35_2b.final.fin.r2}), but increases it at 4B (\res{h1.bb_qwen35_4b_base.final.fin.r2} to \res{h1.bb_qwen35_4b.final.fin.r2}) and 9B (\res{h1.bb_qwen35_9b_base.final.fin.r2} to \res{h1.bb_qwen35_9b.final.fin.r2}). 
Qwen3.8-27B writes it \res{h1.bb_qwen38_27b.final.r2} times/1k (\Cref{tab:nobase}), and openPangu-Embedded-7B-V1.1 and V1 write \res{h1.bb_pangu_e7v11.final.fin.r2} and \res{h1.bb_pangu_e7m.final.fin.r2} (\Cref{tab:provenance}). 
By contrast, OLMo-2 7B's base rarely writes the marker (\res{h1.bb_olmo2_base.final.freq2}/1k) and yields downstream models that never do (\Cref{tab:provenance}).

\begin{table}[t]
  \centering
  \tabstyle
  \fitwidth{\begin{tabular}{@{}llcc@{}}
    \toprule
    \textbf{Model} & \textbf{Training} & \textbf{MATH-500 (\%)} & \textbf{\form{wait} per 1k} \\
    \midrule
    Qwen3.8-27B & post-training & 96.4 & \cellcolor{cCmpA!17}0.91\ci{0.86}{0.96} \\
    openPangu-Embedded-7B-V1.1 & post-training & 95.8 & \cellcolor{cCmpA!12}0.44\ci{0.42}{0.47} \\
    openPangu-Embedded-7B & post-training & 91.8 & \cellcolor{cCmpA!19}1.41\ci{1.37}{1.45} \\
    \bottomrule
  \end{tabular}}
  \caption{\textbf{Models released without a base also write \form{wait}.} }
  \label{tab:nobase}
\end{table}

\textbf{Training data supplies the marker.} These observational comparisons date the marker's onset relative to training stages, not to individual training examples.

\takeaway{Zero-RL rarely installs \form{wait}. In all but one of \res{h1.zr.n_runs} runs from a base without it, RL raises accuracy while the marker stays below 0.05 per 1k tokens. The marker becomes frequent with mid-training, reasoning-rich pretraining or fine-tuning on traces.}

\paragraph{Where the Marker Exists, Released RL-Zero Recipes Multiply It.}
\label{sec:exp:reshape}
\label{sec:exp:h1}

\begin{figure*}[t]
  \centering
  \includegraphics[width=\textwidth]{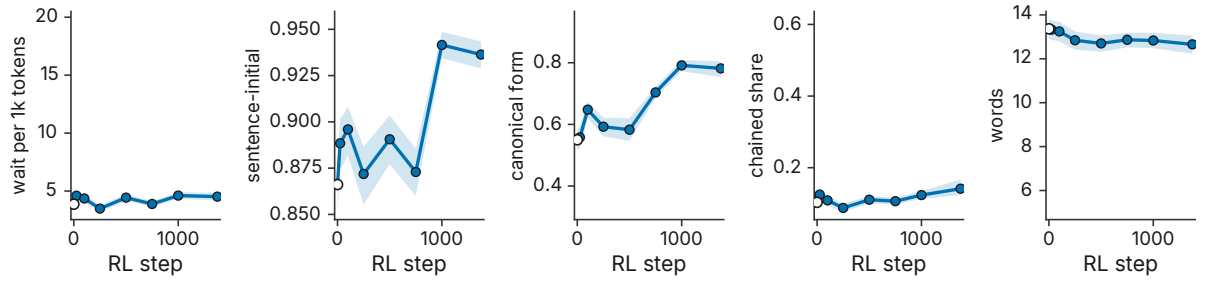}
  \caption{\textbf{OLMo-3 RL-Zero multiplies \form{wait} and erodes its form; RL after fine-tuning changes its rate little and restores its comma.}
  (top) OLMo-3 RL-Zero; (bottom) OLMo-3 Think; (a)--(e) rate, sentence-initial share, \tok{Wait,} share, chaining, sentence length; hollow: base; bands: 95\% cluster-bootstrap CIs.}
  \label{fig:clines}
\end{figure*}

\textbf{Both released RL-Zero recipes multiply the marker.} \Cref{fig:clines} follows the two OLMo-3 trajectories. The Qwen2.5 GRPO runs emit fewer than 100 occurrences at every checkpoint and are omitted. Along OLMo-3 RL-Zero, the rate of \form{wait} rises from \res{h1.s1.base.freq} to a peak of \res{h1.s1.late.freq}/ 1k tokens at step 1900, ending at \res{h1.s1.final.freq} at the released final checkpoint (\Cref{fig:clines}a; Holm-adjusted $p < 0.001$). Over \res{h1.s1seeds.final.freq_n} decoding seeds, the final rate is \res{h1.s1seeds.final.freq_mean} $\pm$ \res{h1.s1seeds.final.freq_std} (\Cref{tab:main}). The size of the rise depends on decoding: at temperature 0.6 instead of 1.0, the final checkpoint writes \res{h1.s1t06.final.freq} occurrences per 1k tokens compared to \res{h1.s1t06.base.freq} for the base, and \res{ctl.s1t06.final.trunc}\% of its responses reach the length limit (base \res{ctl.s1t06.base.trunc}\%). We therefore compare rates only within a series at fixed settings. The rise also depends on the recipe: run for 400 steps from the same base and evaluated identically, our KL-free GRPO recipe does not multiply the marker, ending at \res{h1.fork_b0m.step0400.a_boot} times the base's rate on all responses and \res{h1.fork_b0m.step0400.a_boot_fin} on finished ones, against \res{h1.s1.step0400.a_boot} for the released run, with disjoint intervals at steps 200 and 400. With a small KL penalty, the registered outcome is undetermined (\Cref{app:prereg}). RL-Zero on another base that already writes the marker also multiplies it (MiMo-7B: \res{h1.bb_mimo_base.final.freq2} to \res{h1.bb_mimo_rlzero.final.freq2}). RL after fine-tuning on traces in Phi-4 (\res{h1.pop_phi4r.final.freq2} to \res{h1.pop_phi4rp.final.freq2}), DeepScaleR (\res{h1.bb_r1d1p5b.final.fin.r2} to \res{h1.bb_deepscaler.final.fin.r2} among finished responses) and Light-R1 (\res{h1.bb_r1d14.final.fin.r2} to \res{h1.bb_light14.final.fin.r2}) shows similar amplification.

\textbf{Our GRPO recipe erases a marker seeded by fine-tuning.} In a registered transplant, we fine-tune Qwen2.5-Math-1.5B on 512 correct traces whose sentences open with \tok{Wait,} and then train it with our GRPO recipe. After fine-tuning, the model writes \form{wait} \res{h1.seed.W512.sft.r2} times per 1k tokens. After 400 RL steps, it writes none (\res{h1.seed.W512.step0400.nwait} occurrences in \res{h1.seed.W512.step0400.ntok} tokens). Accuracy rises from \res{ctl.seed.W512.sft.acc}\% to \res{ctl.seed.W512.step0400.acc}\%, and responses shrink from \res{ctl.seed.W512.sft.toks_per_resp} to \res{ctl.seed.W512.step0400.toks_per_resp} tokens. As \form{wait} disappears, the neutral \tok{Next,} grows \res{h1.seed.W512.trend.a_nextopen.r2}-fold in the same runs. When the seed traces carry \tok{Next,} instead, RL keeps \res{h1.seed.X512.trend.a_nextopen.r2} of it. This run has a confound: \res{ctl.seed.W512.sft.trunc}\% of the fine-tuned responses reach the 3,072-token limit and earn no reward, so RL also shortens responses. The arm seeded with 64 traces, whose fine-tuned responses reach the limit less often (\res{ctl.seed.W64.sft.trunc}\%), loses the marker as well (\res{h1.seed.W64.sft.r2} per 1k after fine-tuning; \res{h1.seed.W64.step0400.nwait} occurrences in \res{h1.seed.W64.step0400.ntok} tokens after RL). The marker falls just as fast among finished responses (\res{h1.seed.W512.sft.freq_finished2} to \res{h1.seed.W512.step0400.freq_finished2} per 1k). A registered control rules out length as the cause: with an 8,192-token limit, the same transplant goes from \res{h1.seed.W512LY.sft.fin.r2} to \res{h1.seed.W512LY.step0400.fin.r2} per 1k among finished responses ($A = \res{h1.seed.W512LY.trend.a_wait_fin}$, 95\% CI [\res{h1.seed.W512LY.trend.a_wait_fin.lo}, \res{h1.seed.W512LY.trend.a_wait_fin.hi}]), far below the registered threshold of 0.2 for a penalty on the marker itself, and the same control seeded with \tok{Next,} reaches similar accuracy while writing no \form{wait} at step 400 (\Cref{app:prereg}).

\textbf{The marker's form wears down.} Its position was already fixed before RL and remains so: \res{h1.s1.base.si}\% of occurrences open their sentence at the base, and \res{h1.s1.final.si}\% do so at the end (\Cref{fig:clines}b). The form itself erodes instead. The share written \tok{Wait,} falls from \res{h1.s1.base.orth}\% to \res{h1.s1.final.orth}\% as the comma drops (\tok{Wait no}, \tok{Wait let me}). This marks a change of level between steps 1000 and 1900 rather than a gradual trend (\Cref{fig:clines}c). Reduplicated chains (\tok{Wait, wait}) rise from \res{h1.s1.base.chain} to \res{h1.s1.final.chain} per 1k occurrences, and marker-initial sentences shrink from \res{h1.s1.base.sentlen} to \res{h1.s1.final.sentlen} words. In grammaticalization terms, this is attrition, the loss of substance typical of a marker's late stages \citep{lehmann2015thoughts}. Its contexts of use also shift: the usage distribution of the last checkpoint differs from the base's by a Jensen--Shannon divergence of \res{h2.s1.final.jsd_base}, compared to a noise floor of \res{h2.s1.base.jsd_floor}. Running example of this attrition is in \Cref{app:case}.

\textbf{RL follows a different path after supervised fine-tuning.} Along OLMo-3 Think, where RL starts from a fine-tuned and preference-tuned checkpoint, the rate changes little (\res{h1.s2.base.freq} to \res{h1.s2.late.freq}; RL-Zero: \res{h1.s1.base.freq} to \res{h1.s1.final.freq}). The comma form is \emph{restored} (\res{h1.s2.base.orth}\% to \res{h1.s2.late.orth}\%; Holm-adjusted $p = \res{h1.s2.trend.rho_orth.p_holm}$), and usage stays at the noise floor. At 32B, OLMo-3.1 Think keeps the rate flat (\res{h1.s9.base.freq} to \res{h1.s9.late.freq}). Along the supervised fine-tuning trajectory of the same base, the marker barely changes (rate \res{h1.s2ctl.step5000.freq} to \res{h1.s2ctl.step43000.freq}; comma share \res{h1.s2ctl.step5000.orth}\% to \res{h1.s2ctl.step43000.orth}\%). Robustness to template and decoding appears in \Cref{app:results}.

\takeaway{Where pretraining or mid-training supplies \form{wait}, both released RL-Zero recipes (OLMo-3, MiMo) multiply it, and OLMo-3's drops its comma and chains it. In OLMo-3, RL following supervised fine-tuning changes its rate little. Meanwhile, our GRPO recipe erases a marker seeded by fine-tuning. The separate roles of source and recipe remain open.}

\subsection{Analysis}
\label{sec:exp:analysis}
\paragraph{The Marker Becomes a More Sensitive but More Liberal Alarm.}
\label{sec:exp:alarm}

\begin{figure}[t]
  \centering
  \includegraphics[width=\linewidth]{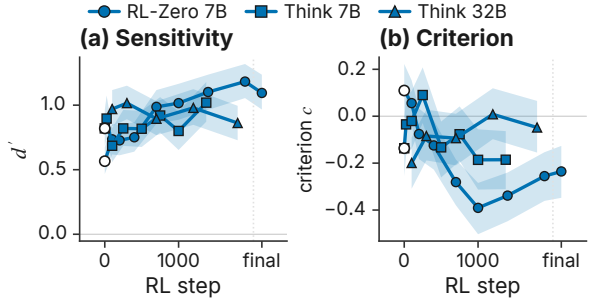}
  \caption{\textbf{Under RL-Zero, revision markers become a more sensitive but more liberal alarm.}
  Sensitivity $d'$ and criterion $c$ of any revision marker after clean versus error-injected prefixes, per checkpoint, with 95\% cluster-bootstrap bands.}
  \label{fig:alarm}
\end{figure}

On prefixes with and without an injected error, the registered outcome---any revision marker---becomes more sensitive, with $d'$ rising from \res{h2.s1.base.dprime} to \res{h2.s1.final.dprime} (95\% CI [\res{h2.s1.final.dprime.lo}, \res{h2.s1.final.dprime.hi}]). It also becomes more liberal: the criterion falls from \res{h2.s1.base.crit} to \res{h2.s1.final.crit} (\Cref{fig:alarm}). \form{Wait} alone shows a similar trend: $d'$ rises from \res{h2.s1.base.dprime_wait} to \res{h2.s1.final.dprime_wait} (CI [\res{h2.s1.final.dprime_wait.lo}, \res{h2.s1.final.dprime_wait.hi}]) and $c$ falls from \res{h2.s1.base.crit_wait} to \res{h2.s1.final.crit_wait}. \form{Wait} alone follows \res{h2.s1.final.hr_wait}\% of prefixes with an error (base \res{h2.s1.base.hr_wait}\%) but also \res{h2.s1.final.fa_wait}\% of error-free ones (base \res{h2.s1.base.fa_wait}\%). The marker therefore maintains its link to errors rather than bleaching, contradicting our preregistered prediction (\Cref{app:prereg}). After supervised fine-tuning, RL leaves the alarm almost unchanged: the criterion moves from \res{h2.s2.base.crit} to \res{h2.s2.late.crit} at 7B and from \res{h2.s9.step0100.crit} to \res{h2.s9.late.crit} at 32B, while $d'$ moves from \res{h2.s2.base.dprime} to \res{h2.s2.late.dprime} and from \res{h2.s9.step0100.dprime} to \res{h2.s9.late.dprime}, against \res{h2.s1.base.dprime} to \res{h2.s1.final.dprime} along RL-Zero. Under RL-Zero, the marker thus remains a signal of errors but fires more readily, even after error-free steps, so its evidential weight is a property of the checkpoint.

\paragraph{Judged Revisions Can Reverse the Human-Labelled Trend.}
\label{sec:exp:judge}

On a \res{gold.v4.n_A}-sentence blind audit with expert consensus labels, one of two LLM judges reverses the human-labelled direction of the revision rate along RL-Zero. LLM judges are a common shortcut for labelling reflective behaviour along training \citep{zeng2025simplerlzoo,gandhi2025cognitive}. We test this against a blind human audit of \res{gold.v4.n_A} sentences from the base and the last RL-Zero checkpoint, stratified by marker. The primary judge agrees with the consensus labels at $\kappa = \res{gold.v4.kappa_human_qwen}$. However, \textbf{the judge over-calls revisions at the base}: on the same weighted items, its revision rate \emph{falls} from \res{gold.v4.revrate_judge_base}\% to \res{gold.v4.revrate_judge_1900}\%, while the consensus rate \emph{rises} from \res{gold.v4.revrate_base}\% to \res{gold.v4.revrate_1900}\%. A second judge tracks the consensus labels (\res{gold.v4.revrate_judge_oss_base}\% to \res{gold.v4.revrate_judge_oss_1900}\%; $\kappa = \res{gold.v4.kappa_human_oss}$). Our findings therefore rest on direct marker measurements; the audit is descriptive. Judge-based indices and all audits appear in \Cref{app:stats,app:prereg}.


\section{Discussion}
\label{sec:discussion}

\paragraph{RL reshapes inherited markers; it rarely coins them.}
A rise in \form{wait} during RL does not show that RL introduced the marker; the starting checkpoint may already use it. In every lineage traced but one, the marker possesses a prior source: mid-training data, a reasoning-rich pretraining mixture, or fine-tuning on traces. Reports of an ``aha moment'' should state whether the base already writes the marker, and how often. Marker counts alone therefore cannot show that reflection was newly acquired.

\paragraph{Marker frequency and meaning shift across training.}
Marker frequency and meaning shift across training, which undermines uncertainty heuristics that read \form{wait} as a fixed signal of doubt. Along OLMo-3 RL-Zero, the marker fires more often after errors and correct steps alike. The ratio rises, so when it fires it remains a strong signal, but its overall frequency changes. RL after supervised fine-tuning leaves the alarm almost unchanged. A marker's evidential weight requires direct measurement on the specific checkpoint, for example with the label-free injected-error test of \Cref{sec:method}.

\paragraph{Automated behavioural judgments require human validation.}
Judge-labelled behaviour rates scale, but one of our judges reversed the human-labelled trend despite moderate item-level agreement. Direct marker measurements must therefore support lexical claims, and behavioural judgments need human validation.

\paragraph{Language change under optimisation} is recipe-specific rather than a single cline; \Cref{sec:theory} relates our results to grammaticalization theory.

\section{Conclusion}
\label{sec:conclusion}


We find that zero-RL almost never coins the reflection marker \form{wait} across the training history of open models. From bases lacking the marker, zero-RL raises accuracy and keeps the marker below 0.05 per 1k tokens in most runs. The marker arrives instead with mid-training, reasoning-rich pretraining, or fine-tuning on traces. When pretraining or mid-training supplies the marker, both released RL-Zero recipes multiply it, while KL-free GRPO from the same base does not. OLMo-3's recipe erodes its form and makes it a more liberal alarm. In OLMo-3, RL after supervised fine-tuning changes its rate little, and RL erases a marker seeded by light fine-tuning. Understanding reasoning traces starts with knowing where their words originate. By mapping these developmental trajectories, we establish that training data seeds reflection markers while specific RL recipes dictate their ultimate survival guiding the future reasoning model training.

\section*{Limitations}
\label{sec:limitations}

Our provenance evidence is observational. The compared lineages differ in data and recipe alongside the marker. The zero-RL result rests on \res{h1.zr.n_runs} runs from \res{h1.zr.families} bases. Each open checkpoint series constitutes a single training run, so its intervals capture evaluation sampling variance rather than training variance. All RL runs involve 0.5B--32B models trained with outcome rewards on English mathematics. Frontier-scale RL, other domains, other languages, and process rewards remain untested.

Our measurements capture the lexical form rather than underlying reflective behaviour. A model can reflect without writing \form{wait}, and markers outside our inventory escape detection. Judge-based quantities agree weakly to moderately with the human consensus labels (\Cref{app:stats}). The audits are consensus labels on a sample, with no independent double annotation, so inter-annotator agreement is not reported; because they remain descriptive, our headline claims rest on label-free indices and pre-registered tests. The alarm test covers only the arithmetic errors we inject. Several predictions we registered did not hold (\Cref{app:prereg}).
\section*{Ethical Considerations}
\label{sec:ethics}

\paragraph{Artefacts.} We use publicly released checkpoints, datasets and code under their licences (\Cref{app:licenses}); no model was trained on personal data by us, and our runs use a public mathematics dataset. Code and data are at \url{https://anonymous.4open.science/r/wait-provenance/}: the provenance map, the result keys, the label-free index suite and the three expert-consensus audits, under CC-BY-4.0 (\Cref{app:licenses}).

\paragraph{Human participants.} No crowd workers were involved. In three blind validity audits of model-generated mathematics, the third with further marker occurrences (\Cref{app:stats}), labels are consensus labels from a panel of ten annotators with PhD-level training in related fields, agreed through discussion; an author recorded them with a software tool that suggested no answers. All other reference labels were produced by LLM raters and are described as such.

\paragraph{Dual use.} Showing that a reasoning marker changes its evidential weight with training could help someone evade a monitor that reads such markers. The effect we report is a drift that arises without any adversary, and the remedy is simple and public: calibrate any marker-based reading on the checkpoint being read. We therefore see the benefit to monitor builders as outweighing the risk.

\paragraph{AI assistance.} AI assistants were used for code writing and polish but the whole research is design and checked by researchers.


\bibliography{custom}

\appendix
\setcounter{dbltopnumber}{4}\setcounter{topnumber}{4}\renewcommand{\dbltopfraction}{0.95}\renewcommand{\topfraction}{0.95}\renewcommand{\dblfloatpagefraction}{0.8}\renewcommand{\floatpagefraction}{0.8}
\section*{Appendix Contents}
\startcontents[app]
\printcontents[app]{l}{1}{\setcounter{tocdepth}{2}}

\section{Data and Resources}
The models, checkpoints and licensed artefacts the paper uses.

\subsection{Models and Checkpoints}
\label{app:models}
This section lists every checkpoint series we analyse, with its base, recipe and evaluated steps (\Cref{tab:app:models}).
\begin{table*}[t]
  \centering
  \tabstyle
  \renewcommand{\UrlBreaks}{\do\/\do-\do\_\do.}
  \begin{tabularx}{\textwidth}{@{}l>{\raggedright\arraybackslash}X>{\raggedright\arraybackslash}X>{\raggedright\arraybackslash}p{1.9cm}>{\raggedright\arraybackslash}p{3.0cm}>{\raggedright\arraybackslash}p{1.8cm}@{}}
    \toprule
    \textbf{ID} & \textbf{Series (Hugging Face)} & \textbf{Base ($t_0$)} & \textbf{Training} & \textbf{Checkpoints used} & \textbf{Role} \\
    \midrule
    S1 & \nolinkurl{allenai/Olmo-3-7B-RL-Zero-Math} & \nolinkurl{Olmo-3-1025-7B} & RL from the base & base, 100, 200, 400, 700, 1000, 1400, 1900, final & primary \\
    S1-pre & \nolinkurl{allenai/Olmo-3-1025-7B} (branches) & -- & pretraining, mid-training & stage 1: 141k, 707k, 1.41M; stage 2: 47.7k; stage 3: 6k, 11.9k & source \\
    S2 & \nolinkurl{allenai/Olmo-3-7B-Think} & \nolinkurl{Olmo-3-7B-Think-DPO} & RL after SFT and DPO & DPO, 25, 100, 250, 500, 750, 1000, 1375 & primary \\
    S2-ctl & \nolinkurl{allenai/Olmo-3-7B-Think-SFT} & \nolinkurl{Olmo-3-1025-7B} & SFT only & 5k, 15k, 25k, 35k, 43k & lineage control \\
    S9 & \nolinkurl{allenai/Olmo-3.1-32B-Think} & \nolinkurl{Olmo-3-32B-Think-DPO} & RL after SFT and DPO & DPO, 100, 300, 700, 1200, 1800, 2300 & 32B replication \\
    S8 & Qwen2.5-32B $\rightarrow$ ORZ / DAPO / QwQ / R1-Distill & \nolinkurl{Qwen2.5-32B} & RL / distillation & endpoints & same-base comparison \\
    S10 & gpt-oss-120b (low / medium / high effort) & none released & undisclosed & endpoint & frontier endpoint \\
    S3a & \nolinkurl{od2961/Qwen2.5-1.5B-Open-R1-GRPO-math-v1} & \nolinkurl{Qwen2.5-1.5B-Instruct} & GRPO & base, last & secondary (no licence) \\
    \midrule
    REF & our GRPO (verl) & \nolinkurl{Qwen2.5-Math-1.5B} (2 seeds), \nolinkurl{Qwen2.5-Math-7B} & zero-RL & base, 400 & zero-RL control \\
    Fork & our GRPO, $\beta = 0.001$ and $\beta = 0$ & \nolinkurl{Olmo-3-1025-7B} & zero-RL & every 10 steps to 100, then every 50 to 400 & controlled fork (D48) \\
    Seeded & SFT on 64 or 512 seed traces, then our GRPO & \nolinkurl{Qwen2.5-Math-1.5B} & SFT $\rightarrow$ RL & SFT-only, 400 & transplant (D43) \\
    Q3 & our GRPO, 1 seed & \nolinkurl{Qwen3-4B-Base} & zero-RL & base, 100--400 & zero-RL count (D49) \\
    \bottomrule
  \end{tabularx}
  \caption{\textbf{Checkpoint series.} Released trajectories (top) and our own runs (bottom). Every released endpoint is listed in \Cref{tab:provenance}; licences are in \Cref{app:licenses}. openPangu-v2.1-Pro (API) in \Cref{tab:nobase}: vendor API queried 2026-10-13, weights not public, thinking cannot be disabled via the API; the API is rate-limited, so $n=\rt{ctl.bb_pangu_v21pro.final.n}$ samples over \rt{ctl.bb_pangu_v21pro.final.n_problems} of 200 problems; rate per 1k output tokens (thinking and answer), \rt{h1.bb_pangu_v21pro.complete.freq2} on the \rt{h1.bb_pangu_v21pro.complete.n_problems} problems with all four samples against \rt{h1.bb_pangu_v21pro.final.freq2} on all.}
  \label{tab:app:models}
\end{table*}

\paragraph{Checkpoint provenance.}
The final OLMo-3 RL-Zero checkpoint is the released model, whose optimizer step is not documented (it is later than step 1900); we treat it as the last point without placing it on the step axis. Some secondary series have no licence or step metadata (index-only ordering for one series, repository names as the only provenance for another); they are used for robustness only and never for a headline number. Only our own GRPO condition has two training seeds. openPangu-R-7B-2512 is not runnable on our CUDA stack (part of its weights requires the vendor's NPU kernels) and is omitted. Concurrent, unpublished work on reasoning registers uses the same OLMo-3 checkpoints \citep{reasoningcues2026rollouts}.

\subsection{Artefacts and Licences}
\label{app:licenses}
Every model, dataset and tool we use is listed in \Cref{tab:licenses} with the licence stated on its official card or licence file, checked on 5 October 2026; all repositories are official and ungated.
\begin{table*}[t]
  \centering
  \tabstyle\footnotesize
  \begin{tabularx}{\textwidth}{@{}p{3.6cm}>{\raggedright\arraybackslash}X@{}}
    \toprule
    \textbf{Licence} & \textbf{Models, datasets and tools} \\
    \midrule
    Apache-2.0 & OLMo-2 and OLMo-3 (all series, branches and Think models); Qwen2.5, Qwen2.5-Math, Qwen3, Qwen3.5, Qwen3.8 and QwQ models; SmolLM3; Granite 3.3; Mistral-7B-v0.1 and Mistral-Small-24B; Yi-1.5; OpenThinker-7B; Bespoke-Stratos-7B; Light-R1-14B-DS; DAPO-Qwen-32B; SimpleRL-Zoo (all but the 14B release); Oat-Zero; General-Reasoner; One-Shot-RLVR; gpt-oss; AIME 2024 and 2025; vLLM; verl; Math-Verify \\
    MIT & Phi-4 family; DeepSeek-R1-Distill (Qwen and Llama); DeepScaleR-1.5B; Open-Reasoner-Zero; MiMo-7B; GLM-4 and GLM-Z1; GSM8K; MMLU-Pro; MATH (training split); MATH-500 (its card states no licence; both of its sources are MIT) \\
    Llama 3.1 Community License & Llama-3.1-Tülu-3 8B models \\
    NVIDIA Open Model License & AceReason-Nemotron-7B \\
    CC-BY-4.0 & OpenReasoning-Nemotron-7B; reasoning-cues rollouts \\
    Qwen License Agreement & Qwen2.5-Math-PRM-7B \\
    OpenPangu Model License v1.0 & openPangu models; research use is permitted, and its \S3 excludes use within the European Union. We run them outside the EU and distribute nothing built on them. \\
    No licence text & SimpleRL-Zoo 14B and the Qwen2.5-1.5B Open-R1 GRPO run (used as secondary rows only, never for a headline count); AMC 2023 (no licence or source stated on the Hugging Face dataset; competition problems \textcopyright{} MAA; evaluation prompts only) \\
    \bottomrule
  \end{tabularx}
  \caption{\textbf{Licences of all artefacts.} One row per licence.per-repository identifiers and checks are at \url{https://anonymous.4open.science/r/wait-provenance/}.}
  \label{tab:licenses}
\end{table*}

\paragraph{Intended use.}
We use every artefact for measurement only: we run inference on public benchmark prompts, count reasoning markers in the outputs and grade the answers, and our own GRPO runs fine-tune Apache-2.0 Qwen bases on the MIT-licensed MATH training split. This is within the research-use grant of every licence listed, and none of them restricts publishing such measurements. We redistribute no model weights, no derived checkpoints, including our own, and no dataset contents. The artefacts we release, under CC-BY-4.0, are the provenance map (every endpoint with its lineage, recipe group and marker indices), the result keys behind every number in the paper, the three consensus-labelled audits with their annotation guidelines, and the analysis code. Models whose distributors state no licence are never used for a headline number.

\FloatBarrier
\section{Experimental Setup}
How each index is defined, how every series is sampled and graded, and how the baselines are run.

\subsection{From Parameters to Measurements}
\label{app:operationalization}
This section maps each grammaticalization parameter to the index that measures it.

\subsection{Method Details}
\label{app:methoddetails}
This section details the models, data, indices and alarm test of \Cref{sec:method}.

\paragraph{Models and checkpoints.}
We need RL trajectories whose intermediate checkpoints are public, from bases that differ in origin.
The primary series is OLMo-3-7B RL-Zero-Math, trained with RL directly from the fully open OLMo-3 base \citep{olmo2025olmo}; a fully open base, with its pretraining checkpoints, lets us test rather than assume where the marker comes from (S1-pre), a question raised for Qwen bases, whose RLVR gains depend on behaviours already present before RL \citep{shao2025spurious}.
The second series, OLMo-3-7B-Think, applies RL to a model that already reasons after supervised fine-tuning and preference tuning, so the markers exist at $t_0$ and we test whether the cline continues.
For provenance we compare these trajectories with released endpoints grouped by their documented recipe (\Cref{tab:provenance}): zero-RL runs from bases that do not use the marker, our own GRPO runs among them, trained with verl \citep{sheng2025hybridflow} on Qwen2.5-Math \citep{yang2024qwen25math} and Qwen3-4B bases; descendants of one base trained on reasoning traces, QwQ-32B \citep{qwen2025qwq} and R1-Distill-Qwen-32B \citep{deepseekai2025r1} from Qwen2.5-32B; and the pretraining and mid-training branches of the OLMo-3 base (S1-pre). OLMo-3.1-32B-Think replicates the Think trajectory at 32B, and gpt-oss-120b at three reasoning efforts is an endpoint without a released base.
A secondary GRPO run from Qwen2.5-1.5B-Instruct (no licence metadata; robustness only) and the supervised fine-tuning trajectory of the OLMo-3 thinker (same base, no RL) serve as controls.
We use \res{design.nckpt} checkpoints in total. A checkpoint is any evaluated point of a training series; an endpoint is a released or final model, of which the provenance map lists \res{h1.prov.n_rows}; the zero-RL count uses base--final pairs (\res{h1.zr.n_runs} runs from \res{h1.zr.n_bases} bases); identifiers, revisions and licences are in \Cref{app:models}.

\paragraph{Data and sampling.}
Traces come from MATH-500 \citep{hendrycks2021measuring,lightman2024lets} (the 32B endpoints add AIME 2024), sampled with vLLM \citep{kwon2023efficient} under each series' own training template and temperature (\Cref{app:details}).
Markers are matched on detokenised text, never on token identifiers, because tokenisers split \tok{Wait,}, \tok{ Wait} and \tok{wait} differently.

\paragraph{Revision events.}
Indices defined on revision events are reported in \Cref{app:prereg}, because a human audit did not validate their trends. A \emph{revision event} is a sentence that changes a previously stated value, negates an earlier statement or restarts a derivation (rule-based detector plus an open-weight LLM judge whose consistency with adjudicated LLM reference labels, and whose agreement with a human audit, we report; \Cref{app:details}).
Over the paradigm $F$ of forms that open revisions (\form{wait}, \form{hold on}, \form{hmm}, \form{actually}, \form{but}, \form{let me check}, \dots),
\begin{equation}
  H_{\mathrm{para}}(t) = -\sum_{f \in F} p_t(f)\log_2 p_t(f),
  \label{eq:hpara}
\end{equation}
where $p_t(f)$ is the share of $f$ among the forms that open revision events; $N_{\mathrm{eff}} = 2^{H_{\mathrm{para}}}$ is the effective paradigm size.
\emph{Obligatoriness} is the share of revision events opened by some form of the paradigm,
\begin{equation}
  \mathrm{OBL}(t) = \Prob_t\big(f \in F \mid \text{revision}\big),
  \label{eq:obl}
\end{equation}
and its converse, the marker's \emph{reliability} $\mathrm{REL}(t)$, is the probability that a revision event follows within two sentences of the marker.

\paragraph{A seeded-marker control.}
To test the provenance claim causally, we registered a transplant experiment before running it. A base that does not use the marker (Qwen2.5-Math-1.5B) is first fine-tuned on a small number $N \in \{64, 512\}$ of correct reasoning traces in which sentences open with \tok{Wait,}, and then trained with the same GRPO recipe as our zero-RL runs. Two matched arms use the same traces with the opener replaced by the neutral filler \tok{Next,} or deleted. Comparing the marker's rate after fine-tuning with its rate after RL separates what the traces install from what RL amplifies, and the \tok{Next,} arm separates the form of an opener from its function. The seed traces are short, finished and use \form{wait} only as a sentence-initial opener. Results are in \Cref{sec:exp:reshape} and \Cref{app:prereg}.

\paragraph{Interventions.}
The operator test inserts a form after a sentence boundary at which the prefix is correct and the checkpoint's own next sentence continued rather than revised, so the insertion is counterfactual.
Arms are \tok{Wait,} (the operator form), \tok{Hold on,} and \tok{Hmm,} (members of its paradigm), the neutral filler \tok{Next,} (the reference), \tok{Okay,} and no insertion. \tok{Hmm,} is itself a reflection marker \citep{qian2025demystifying,yu2026revisiting} and \tok{Okay,} is the opener cue of concurrent work \citep{reasoningcues2026rollouts}, so neither can serve as a null; we keep \tok{Okay,} for comparability with that work. A secondary set adds a random function word of matched length, which tests whether any insertion perturbs the model.
The base arm is crossed: prefixes derived from base traces and from late traces are each continued by the base and by the late policy, so a null in the base model cannot be an artefact of an off-distribution prefix.
The outcome is whether the first three sentences revise the step, judged by an open-weight model that never sees the inserted form (it is stripped before judging, for every arm), and whether the revision makes a correct value wrong.
We report the risk difference $\mathrm{RD}_{\mathrm{op}}(t) = \Prob_t(\text{revise} \mid \tok{Wait,}) - \Prob_t(\text{revise} \mid \tok{Next,})$ beside the odds ratio
\begin{equation}
  \mathrm{OR}_{\mathrm{op}}(t) = \frac{\mathrm{odds}_t(\text{revise} \mid \tok{Wait,})}{\mathrm{odds}_t(\text{revise} \mid \tok{Next,})}.
  \label{eq:or}
\end{equation}
A deletion arm removes a marker the checkpoint did emit; together with insertion it separates operators (both effects) from rituals (neither).

\paragraph{Statistics.}
All intervals are 95\% cluster-bootstrap intervals over problems, paired across the checkpoints of a series; series are never pooled, and a hypothesis holds if it holds on two of three series.
Trend claims use Spearman correlations with step; insertion effects are estimated by cluster bootstrap. We registered a fixed test sequence (operator, bleaching, cline indices, monitors) with Holm correction inside each step, but did not apply it; \Cref{app:stats} lists the trend tests actually run, and every analysis is marked as registered, amended (dated) or exploratory in \Cref{app:prereg}.
The open series each come from one training run, so their intervals cover evaluation sampling only; our own runs report the standard deviation over seeds.

\subsection{Experimental Details}
\label{app:details}
This section gives the prompts, decoding settings and grading used for every series.
\paragraph{Prompts and decoding.} Each series is sampled with its own training template. For OLMo-3 RL-Zero this is the repository chat template, which renders:
\begin{tcolorbox}[enhanced,colback=cBoxBg,colframe=cBoxFrame,boxrule=0.5pt,arc=1.5pt,fontupper=\small\ttfamily]
Solve the following math problem step by step. The last line of your response should be the answer to the problem in form Answer: \$Answer (without quotes) where \$Answer is the answer to the problem.\textbackslash n\textbackslash n\{problem\}\textbackslash n\textbackslash nRemember to put your answer on its own line after "Answer:"
\end{tcolorbox}
OLMo-3 Think uses its chat template with the default system prompt and a \texttt{<think>} prefill; markers are counted inside the think block. Qwen series use the \verb|\boxed{}| instruction of their training recipes. Decoding follows each release's recommendation: $T = 0.6$, top-$p$ 0.95 for the thinking models released with that setting (OLMo-3 Think and its SFT trajectory, OLMo-3.1 32B Think, QwQ-32B, R1-Distill-Qwen-32B), and $T = 1.0$ otherwise (RL-Zero, the pretraining branches, base models, the zero-RL Qwen runs and gpt-oss-120b; DAPO with its top-$p$ 0.7). Problems and samples per problem: all 500 MATH-500 problems with 2 samples for OLMo-3 RL-Zero, the pretraining branches and OLMo-3.1 32B Think, and with 4 for the Qwen2.5-1.5B GRPO run; a fixed subset of 300 problems with 2 samples for OLMo-3 Think and its SFT trajectory; 200 problems with 4 samples for the Qwen2.5-32B panel and with 8 for our GRPO seeds (traces per series: RL-Zero \res{n.traces.s1}, Think \res{n.traces.s2}, SFT \res{n.traces.s2ctl}, pretraining branches \res{n.traces.s1pre}, Qwen2.5-1.5B GRPO \res{n.traces.s3a}, our GRPO seeds \res{n.traces.s7}, OLMo-3.1 32B Think \res{n.traces.s9}, Qwen2.5-32B panel \res{n.traces.s8}, gpt-oss-120b \res{n.traces.s10}); at most 8{,}192 new tokens (12{,}288 for the Think models; 7{,}000 for the early pretraining branches, whose context is 8k; 4{,}096 and 3{,}072 for the 1.5B runs). Insertion continuations use $T = 1.0$ and at most 1{,}536 tokens, signal-detection windows at most 300 tokens, and judges greedy decoding. A generic step-by-step template on two checkpoints per series and a greedy subset are robustness rows.
\paragraph{Marker matching and revision events.} Markers are matched with regular expressions on detokenised text (case-insensitive, all orthographic variants), never on token ids. The paradigm $F$ is \form{wait}, \form{hold on}, \form{hmm}, \form{actually}, \form{but}, \form{however}, \form{alternatively}, \form{oh}, \form{no}, \form{let me (re-)check / double-check / verify}, \form{I made a mistake}, \form{that's not right}, \form{let me reconsider}. Revision events combine a rule (a value stated earlier is replaced by a different one, or an explicit negation) with an open-weight judge (\texttt{Qwen2.5-7B-Instruct}, temperature 0, one-letter answer) that sees the sentence with its sentence-initial discourse word mechanically removed; the insertion judge never sees the inserted form. Prompts are at \url{https://anonymous.4open.science/r/wait-provenance/}.

\subsection{Baseline Details}
\label{app:baselines}
All baselines are re-implemented or re-run on our checkpoints; none of their numbers is copied. \textbf{Prior measures.} DeepSeek-R1: rate of the ten reflective words \form{wait}, \form{mistake}, \form{however}, \form{but}, \form{retry}, \form{error}, \form{verify}, \form{wrong}, \form{evaluate}, \form{check} per 1k tokens (the original counts are unnormalised). Open-Reasoner-Zero: its five case-sensitive reflection patterns from the released code. SimpleRL-Zoo: its seven keywords and its behaviour prompt with an open-weight judge. \textbf{Interventions.} Budget forcing appends \form{Wait}, \form{Hmm} or \form{Alternatively} at the end-of-thinking point (the \texttt{Answer:} line for RL-Zero, \texttt{</think>} for Think models, the \verb|\boxed{}| sentence for Qwen), 1--6 times, greedy decoding. NoWait bans its 17 listed words and \citet{yu2026revisiting} their 35 markers by setting their logits to $-\infty$. \textbf{Monitors.} SELFDOUBT uses the published dictionaries and hedge-to-verify ratio $h/(v+1)$, plus a variant with our paradigm $F$ added, since the published lexicon does not contain \form{wait}; lexical hints use the published hedge lexicon; the inflection monitor of \citet{boppana2026reasoning} uses its verbatim prompt with an open-weight judge; the hidden-state probe follows \citet{zhang2025reasoning} but reads a fixed position (the last token of the marker sentence) instead of marker-delimited chunks. Every monitor is Platt-scaled on the same calibration split.

\FloatBarrier
\section{Additional Results}
The full provenance results, per-series indices and further controls.

\subsection{Additional Results}
\label{app:results}
This section collects the full provenance results and the per-series index tables.
\begin{figure*}[p]
  \centering
  \includegraphics[width=\textwidth,height=0.86\textheight,keepaspectratio]{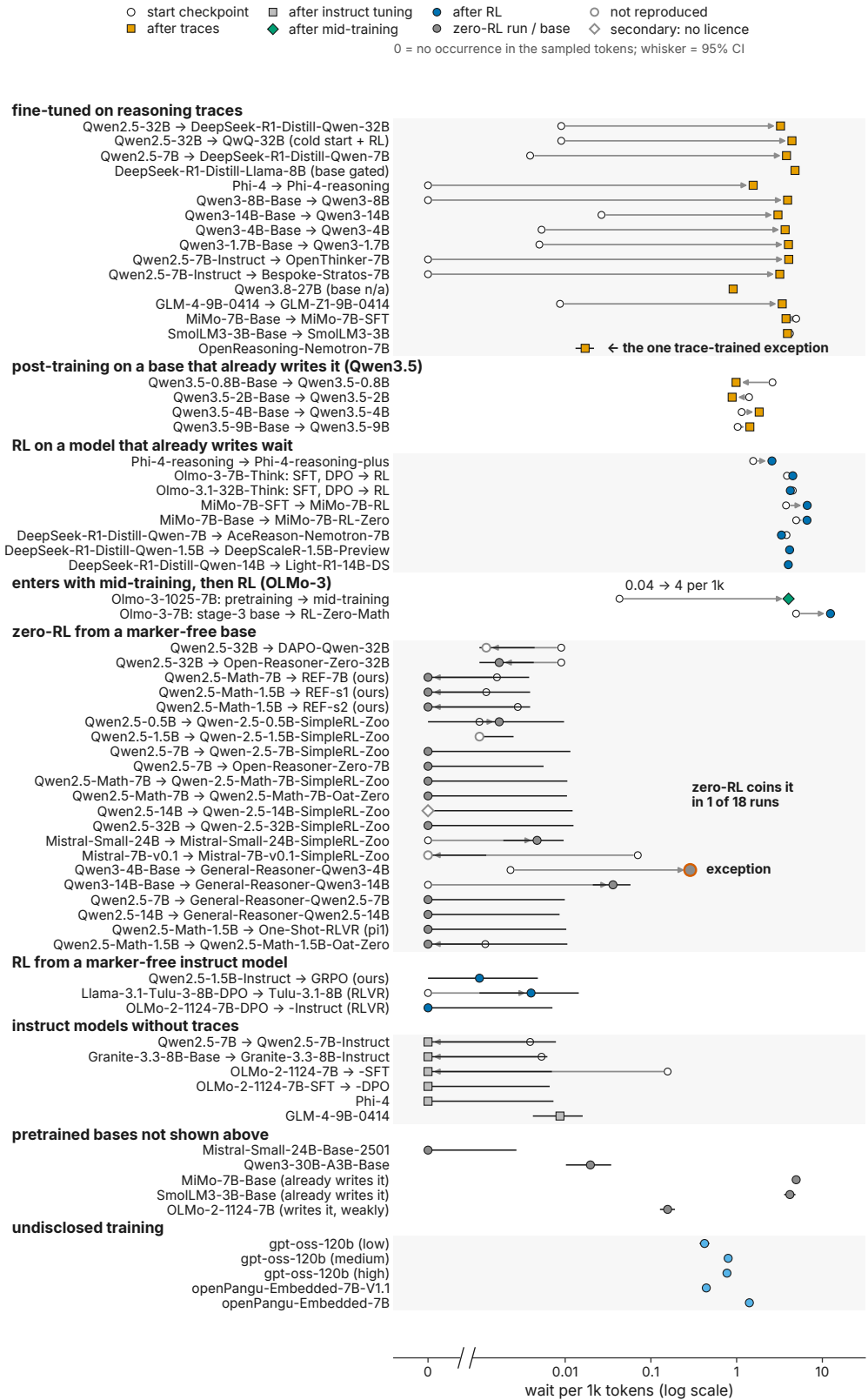}
  \caption{\textbf{All \res{h1.prov.n_rows} endpoints of the provenance map.} One dot per model after training, adding instruct models without trace training (light squares), pretrained bases not shown elsewhere (MiMo-7B-Base and SmolLM3-3B-Base already write \form{wait}) and gpt-oss-120b at three reasoning efforts (training undisclosed).}
  \label{fig:hero:full}
\end{figure*}
\begin{figure}[t]
  \centering
  \includegraphics[width=\linewidth]{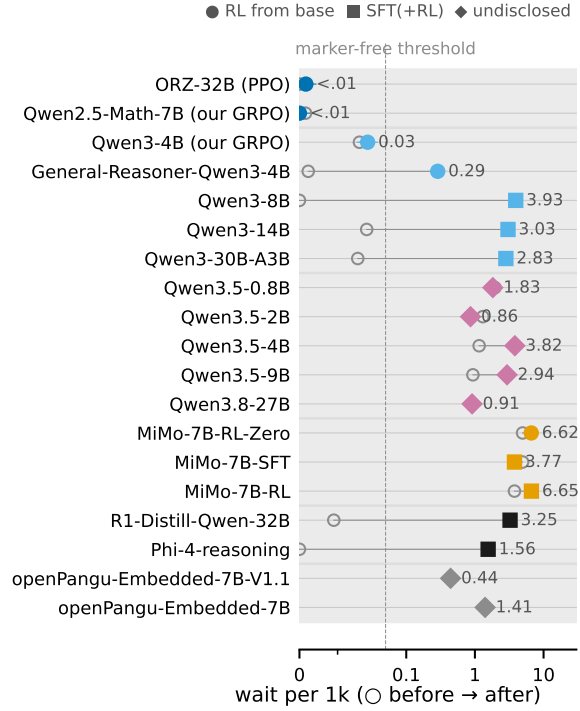}
  \caption{\textbf{Across major model families, \form{wait} follows the training route.} Rate before ($\circ$) and after training, 95\% CI; shape marks the route, colour the family (Qwen2.5, Qwen3, Qwen3.5/3.8, MiMo, DeepSeek/Phi, openPangu).}
  \label{fig:major}
\end{figure}

\begin{figure*}[t]
  \centering
  \includegraphics[width=\textwidth]{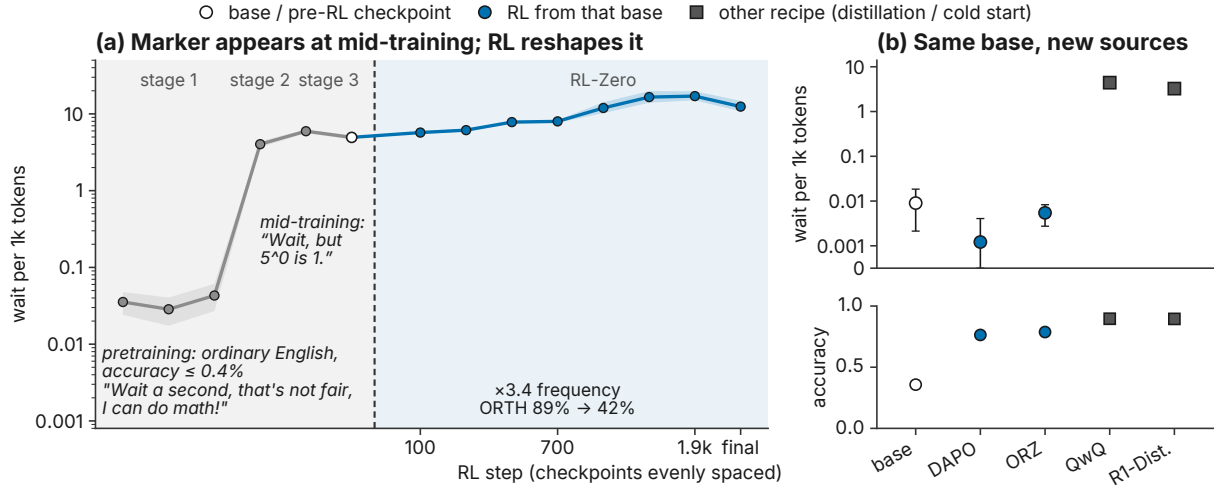}
  \caption{\textbf{In OLMo-3, \form{wait} enters with mid-training; Qwen2.5-32B's descendants use it only after trace training.} (a)~Rate with 95\% bands and comma share across OLMo-3 training stages and RL-Zero steps. (b)~Qwen2.5-32B and its descendants: rate with 95\% CIs, and accuracy.}
  \label{fig:origin}
\end{figure*}

\paragraph{Reading \Cref{fig:origin,tab:clines,tab:clines:app,tab:insertion}.}
In \Cref{fig:origin}a the pretraining branches are stage~1 checkpoints, at which the model solves almost no problem; the dashed line marks the change from training stages to RL steps, and checkpoints are evenly spaced. Quotes are verbatim, chosen by a fixed rule: the shortest marker-initial sentence of 4--64 characters with at least four words. In \Cref{fig:origin}b the rate axis is symmetric-log, so that zero is shown. In \Cref{tab:clines}, intervals cover evaluation sampling only ($B = 2{,}000$; one training run per series). The prior measures are DeepSeek-R1's ten reflective keywords \citep{deepseekai2025r1}, the rate of \form{wait} and SimpleRL-Zoo's keyword count \citep{zeng2025simplerlzoo}, all per 1k tokens, and revision events per 100 sentences; $H_{\mathrm{para}}$ is in bits, rarefied to equal counts. Revision events, $H_{\mathrm{para}}$, OBL and REL depend on revision labels (primary judge Qwen2.5-7B) that the human audit did not validate for these trends (\Cref{app:prereg}); they are shown for completeness, not as findings. In \Cref{tab:insertion}, each row continues the same 300 prefixes (steps judged correct) with the base, a middle or the last checkpoint, four continuations per arm, judged blind to the arm; RD is in points with the half-width of the 95\% cluster-bootstrap CI over prefixes ($B = 10{,}000$); RD$_{\mathrm{syn}}$ compares \tok{Wait,} with the mean of \tok{Hold on,} and \tok{Hmm,}; False is the share of revisions after \tok{Wait,} that replace the correct value with a wrong one. Its bottom block uses 150 prefixes from base-model traces, which rules out an off-distribution explanation of the base model's null.
\begin{table*}[t]
  \centering
  \tabstyle\footnotesize 
  \setlength{\tabcolsep}{1.5pt}
  \fitwidth{\begin{tabular}{@{}ll cccc c ccccc@{}}
    \toprule
    & & \multicolumn{4}{c}{\textbf{Prior measures}} & & \multicolumn{5}{c}{\textbf{Grammaticalization indices}} \\
    \cmidrule(lr){3-6} \cmidrule(lr){8-12}
    \textbf{Series} & \textbf{Stage} & R1 kw.\hup & FREQ\hup & SimpleRL kw.\hup & Rev.\ rate & & SI (\%)\hup & ORTH (\%)\hup & $H_{\mathrm{para}}$\hdown & OBL (\%)\hup & REL (\%)\hdown \\
    \midrule
    \multirow{4}{*}{\makecell[l]{OLMo-3 7B\\RL-Zero}}
      & base & \cell{bm.s1.base.r1kw} & \cell{h1.s1.base.freq} & \cell{bm.s1.base.simplerl} & \cell{h1.s1.base.revrate} & & \cell{h1.s1.base.si} & \cell{h1.s1.base.orth} & \cell{h1.s1.base.hpara} & \cell{h1.s1.base.obl} & \cell{h1.s1.base.rel} \\
      & mid & \cell{bm.s1.mid.r1kw} & \cell{h1.s1.mid.freq} & \cell{bm.s1.mid.simplerl} & \cell{h1.s1.mid.revrate} & & \cell{h1.s1.mid.si} & \cell{h1.s1.mid.orth} & \cell{h1.s1.mid.hpara} & \cell{h1.s1.mid.obl} & \cell{h1.s1.mid.rel} \\
      & late & \cell{bm.s1.late.r1kw} & \cell{h1.s1.late.freq} & \cell{bm.s1.late.simplerl} & \cell{h1.s1.late.revrate} & & \cell{h1.s1.late.si} & \cell{h1.s1.late.orth} & \cell{h1.s1.late.hpara} & \cell{h1.s1.late.obl} & \cell{h1.s1.late.rel} \\
      & $\rho$ (step) & \rt{bm.s1.trend.rho_r1kw} & \rt{h1.s1.trend.rho_freq} & \rt{bm.s1.trend.rho_simplerl} & \na & & \rt{h1.s1.trend.rho_si} & \rt{h1.s1.trend.rho_orth} & \rt{h1.s1.trend.rho_hpara} & \rt{h1.s1.trend.rho_obl} & \rt{h1.s1.trend.rho_rel} \\
    \midrule
    \multirow{4}{*}{\makecell[l]{OLMo-3 7B\\Think}}
      & base & \cell{bm.s2.base.r1kw} & \cell{h1.s2.base.freq} & \cell{bm.s2.base.simplerl} & \cell{h1.s2.base.revrate} & & \cell{h1.s2.base.si} & \cell{h1.s2.base.orth} & \cell{h1.s2.base.hpara} & \cell{h1.s2.base.obl} & \cell{h1.s2.base.rel} \\
      & mid & \cell{bm.s2.mid.r1kw} & \cell{h1.s2.mid.freq} & \cell{bm.s2.mid.simplerl} & \cell{h1.s2.mid.revrate} & & \cell{h1.s2.mid.si} & \cell{h1.s2.mid.orth} & \cell{h1.s2.mid.hpara} & \cell{h1.s2.mid.obl} & \cell{h1.s2.mid.rel} \\
      & late & \cell{bm.s2.late.r1kw} & \cell{h1.s2.late.freq} & \cell{bm.s2.late.simplerl} & \cell{h1.s2.late.revrate} & & \cell{h1.s2.late.si} & \cell{h1.s2.late.orth} & \cell{h1.s2.late.hpara} & \cell{h1.s2.late.obl} & \cell{h1.s2.late.rel} \\
      & $\rho$ (step) & \rt{bm.s2.trend.rho_r1kw} & \rt{h1.s2.trend.rho_freq} & \rt{bm.s2.trend.rho_simplerl} & \na & & \rt{h1.s2.trend.rho_si} & \rt{h1.s2.trend.rho_orth} & \rt{h1.s2.trend.rho_hpara} & \rt{h1.s2.trend.rho_obl} & \rt{h1.s2.trend.rho_rel} \\
    \midrule
    \multirow{4}{*}{\makecell[l]{OLMo-3.1\\32B Think}}
      & base & \na & \cell{h1.s9.base.freq} & \na & \cell{h1.s9.base.revrate} & & \cell{h1.s9.base.si} & \cell{h1.s9.base.orth} & \cell{h1.s9.base.hpara} & \cell{h1.s9.base.obl} & \cell{h1.s9.base.rel} \\
      & mid & \na & \cell{h1.s9.mid.freq} & \na & \cell{h1.s9.mid.revrate} & & \cell{h1.s9.mid.si} & \cell{h1.s9.mid.orth} & \cell{h1.s9.mid.hpara} & \cell{h1.s9.mid.obl} & \cell{h1.s9.mid.rel} \\
      & late & \na & \cell{h1.s9.late.freq} & \na & \cell{h1.s9.late.revrate} & & \cell{h1.s9.late.si} & \cell{h1.s9.late.orth} & \cell{h1.s9.late.hpara} & \cell{h1.s9.late.obl} & \cell{h1.s9.late.rel} \\
      & $\rho$ (step) & \na & \rt{h1.s9.trend.rho_freq} & \na & \na & & \rt{h1.s9.trend.rho_si} & \rt{h1.s9.trend.rho_orth} & \rt{h1.s9.trend.rho_hpara} & \rt{h1.s9.trend.rho_obl} & \rt{h1.s9.trend.rho_rel} \\
    \midrule
    \makecell[l]{OLMo-3 SFT\\(no RL)} & $\rho$ (step) & \rt{bm.s2ctl.trend.rho_r1kw} & \rt{h1.s2ctl.trend.rho_freq} & \rt{bm.s2ctl.trend.rho_simplerl} & \na & & \rt{h1.s2ctl.trend.rho_si} & \rt{h1.s2ctl.trend.rho_orth} & \rt{h1.s2ctl.trend.rho_hpara} & \rt{h1.s2ctl.trend.rho_obl} & \rt{h1.s2ctl.trend.rho_rel} \\
    \bottomrule
  \end{tabular}}
  \caption{\textbf{Keyword counts register that the frequency of \form{wait} changes; the form indices register how.} Base, middle and last checkpoint of each trajectory: mean $\pm$ 95\% CI half-width (cluster bootstrap over problems); $\rho$: Spearman correlation with RL step; arrows: direction grammaticalization predicts; \na: not run.}
  \label{tab:clines}
\end{table*}

\begin{table*}[t]
  \centering
  \tabstyle\footnotesize 
  \setlength{\tabcolsep}{1.5pt}
  \fitwidth{\begin{tabular}{@{}ll c c cccc@{}}
    \toprule
    & & \textbf{Prior} & & \multicolumn{4}{c}{\textbf{Grammaticalization indices}} \\
    \cmidrule(lr){3-3} \cmidrule(lr){5-8}
    \textbf{Series} & \textbf{Stage} & FREQ\hup & & SI (\%)\hup & ORTH (\%)\hup & OBL (\%)\hup & REL (\%)\hdown \\
    \midrule
    \multirow{4}{*}{\makecell[l]{Qwen2.5-1.5B\\GRPO}}
      & base & \cell{h1.s3a.base.freq} & & \cell{h1.s3a.base.si} & \cell{h1.s3a.base.orth} & \cell{h1.s3a.base.obl} & \cell{h1.s3a.base.rel} \\
      & mid & \cell{h1.s3a.mid.freq} & & \cell{h1.s3a.mid.si} & \cell{h1.s3a.mid.orth} & \cell{h1.s3a.mid.obl} & \cell{h1.s3a.mid.rel} \\
      & late & \cell{h1.s3a.late.freq} & & \cell{h1.s3a.late.si} & \cell{h1.s3a.late.orth} & \cell{h1.s3a.late.obl} & \cell{h1.s3a.late.rel} \\
      & $\rho$ (step) & \rt{h1.s3a.trend.rho_freq} & & \rt{h1.s3a.trend.rho_si} & \rt{h1.s3a.trend.rho_orth} & \rt{h1.s3a.trend.rho_obl} & \rt{h1.s3a.trend.rho_rel} \\
    \midrule
    \multicolumn{8}{@{}l}{\textit{32B: one base, four post-trained endpoints}} \\
    \multicolumn{2}{@{}l}{\quad ORZ-32B (RL)} & \cell{h1.s8orz.final.freq} & & \cell{h1.s8orz.final.si} & \cell{h1.s8orz.final.orth} & \cell{h1.s8orz.final.obl} & \cell{h1.s8orz.final.rel} \\
    \multicolumn{2}{@{}l}{\quad DAPO-32B (RL)} & \cell{h1.s8dapo.final.freq} & & \cell{h1.s8dapo.final.si} & \cell{h1.s8dapo.final.orth} & \cell{h1.s8dapo.final.obl} & \cell{h1.s8dapo.final.rel} \\
    \multicolumn{2}{@{}l}{\quad QwQ-32B (RL)} & \cell{h1.s8qwq.final.freq} & & \cell{h1.s8qwq.final.si} & \cell{h1.s8qwq.final.orth} & \cell{h1.s8qwq.final.obl} & \cell{h1.s8qwq.final.rel} \\
    \multicolumn{2}{@{}l}{\quad R1-Distill-32B (SFT)} & \cell{h1.s8r1d.final.freq} & & \cell{h1.s8r1d.final.si} & \cell{h1.s8r1d.final.orth} & \cell{h1.s8r1d.final.obl} & \cell{h1.s8r1d.final.rel} \\
    \midrule
    \multicolumn{8}{@{}l}{\textit{Frontier-scale endpoint without a base checkpoint}} \\
    \multicolumn{2}{@{}l}{gpt-oss-120b, low effort} & \cell{h1.s10low.final.freq} & & \cell{h1.s10low.final.si} & \cell{h1.s10low.final.orth} & \cell{h1.s10low.final.obl} & \cell{h1.s10low.final.rel} \\
    \multicolumn{2}{@{}l}{\quad medium} & \cell{h1.s10med.final.freq} & & \cell{h1.s10med.final.si} & \cell{h1.s10med.final.orth} & \cell{h1.s10med.final.obl} & \cell{h1.s10med.final.rel} \\
    \multicolumn{2}{@{}l}{\quad high} & \cell{h1.s10high.final.freq} & & \cell{h1.s10high.final.si} & \cell{h1.s10high.final.orth} & \cell{h1.s10high.final.obl} & \cell{h1.s10high.final.rel} \\
    \bottomrule
  \end{tabular}}
  \caption{\textbf{Cline indices for Qwen2.5-1.5B GRPO, the 32B endpoints and gpt-oss-120b.} Same protocol and conventions as \Cref{tab:clines}.}
  \label{tab:clines:app}
\end{table*}

\begin{table*}[t]
  \centering
  \tabstyle
  \setlength{\tabcolsep}{1.8pt}
  \fitwidth{\begin{tabular}{@{}ll cccccc c ccccc@{}}
    \toprule
    & & \multicolumn{6}{c}{\textbf{P(revise a correct step) after the inserted form (\%)}} & & \multicolumn{4}{c}{\textbf{Operator force of \tok{Wait,}}} & \\
    \cmidrule(lr){3-8} \cmidrule(lr){10-13}
    \textbf{Series} & \textbf{Policy} & $\varnothing$ & \tok{Next,} & \tok{Okay,} & \tok{Hold on,} & \tok{Hmm,} & \tok{Wait,} & & RD$_{\mathrm{op}}$\hup & RR$_{\mathrm{op}}$\hup & OR$_{\mathrm{op}}$\hup & RD$_{\mathrm{syn}}$\hup & False (\%) \\
    \midrule
    \multirow{3}{*}{\makecell[l]{OLMo-3 7B\\RL-Zero}}
      & base & \rt{h4.s1.base.prev_none_late} & \rt{h4.s1.base.prev_next_late} & \rt{h4.s1.base.prev_okay_late} & \rt{h4.s1.base.prev_holdon_late} & \rt{h4.s1.base.prev_hmm_late} & \rt{h4.s1.base.prev_wait_late} & & \cell{h4.s1.base.rd_op_late} & \rt{h4.s1.base.rr_op_late} & \rt{h4.s1.base.or_op_late} & \rt{h4.s1.base.rd_syn_late} & \rt{h4.s1.base.falserev_wait_late} \\
      & mid & \rt{h4.s1.mid.prev_none_late} & \rt{h4.s1.mid.prev_next_late} & \rt{h4.s1.mid.prev_okay_late} & \rt{h4.s1.mid.prev_holdon_late} & \rt{h4.s1.mid.prev_hmm_late} & \rt{h4.s1.mid.prev_wait_late} & & \cell{h4.s1.mid.rd_op_late} & \rt{h4.s1.mid.rr_op_late} & \rt{h4.s1.mid.or_op_late} & \rt{h4.s1.mid.rd_syn_late} & \rt{h4.s1.mid.falserev_wait_late} \\
      & last & \rt{h4.s1.late.prev_none_late} & \rt{h4.s1.late.prev_next_late} & \rt{h4.s1.late.prev_okay_late} & \rt{h4.s1.late.prev_holdon_late} & \rt{h4.s1.late.prev_hmm_late} & \rt{h4.s1.late.prev_wait_late} & & \cell{h4.s1.late.rd_op_late} & \rt{h4.s1.late.rr_op_late} & \rt{h4.s1.late.or_op_late} & \rt{h4.s1.late.rd_syn_late} & \rt{h4.s1.late.falserev_wait_late} \\
    \midrule
    \multirow{3}{*}{\makecell[l]{OLMo-3 7B\\Think}}
      & base & \rt{h4.s2.base.prev_none_late} & \rt{h4.s2.base.prev_next_late} & \rt{h4.s2.base.prev_okay_late} & \rt{h4.s2.base.prev_holdon_late} & \rt{h4.s2.base.prev_hmm_late} & \rt{h4.s2.base.prev_wait_late} & & \cell{h4.s2.base.rd_op_late} & \rt{h4.s2.base.rr_op_late} & \rt{h4.s2.base.or_op_late} & \rt{h4.s2.base.rd_syn_late} & \rt{h4.s2.base.falserev_wait_late} \\
      & mid & \rt{h4.s2.mid.prev_none_late} & \rt{h4.s2.mid.prev_next_late} & \rt{h4.s2.mid.prev_okay_late} & \rt{h4.s2.mid.prev_holdon_late} & \rt{h4.s2.mid.prev_hmm_late} & \rt{h4.s2.mid.prev_wait_late} & & \cell{h4.s2.mid.rd_op_late} & \rt{h4.s2.mid.rr_op_late} & \rt{h4.s2.mid.or_op_late} & \rt{h4.s2.mid.rd_syn_late} & \rt{h4.s2.mid.falserev_wait_late} \\
      & last & \rt{h4.s2.late.prev_none_late} & \rt{h4.s2.late.prev_next_late} & \rt{h4.s2.late.prev_okay_late} & \rt{h4.s2.late.prev_holdon_late} & \rt{h4.s2.late.prev_hmm_late} & \rt{h4.s2.late.prev_wait_late} & & \cell{h4.s2.late.rd_op_late} & \rt{h4.s2.late.rr_op_late} & \rt{h4.s2.late.or_op_late} & \rt{h4.s2.late.rd_syn_late} & \rt{h4.s2.late.falserev_wait_late} \\
    \midrule
    \bottomrule
  \end{tabular}}
  \caption{\textbf{An inserted \tok{Wait,} pushes every checkpoint, the base included, toward revising a correct step; the size of the push is not identified.} The registered control on prefixes from base-model traces was also run; its results are in the supplementary material. Share of continuations that revise a correct step after each inserted form; RD, RR, OR: risk difference, ratio and odds ratio of \tok{Wait,} against \tok{Next,}.}
  \label{tab:insertion}
\end{table*}

\paragraph{Other reflection openers under zero-RL.}
A zero-RL run could replace \form{wait} with another opener of the same paradigm. In our GRPO runs from Qwen2.5-Math, the paradigm's sentence-initial openers at the last checkpoint are almost all the contrastive connective \form{however} (\res{h1.s7r1.final.reflopen_share_however}\%, \res{h1.s7r2.final.reflopen_share_however}\% and \res{h1.s7r7b.final.reflopen_share_however}\% in the three runs), and \form{hmm}, \form{hold on} and \form{let me check} make up \res{h1.s7r1.final.reflopen_share_hmm}\%, \res{h1.s7r1.final.reflopen_share_hold_on}\% and \res{h1.s7r1.final.reflopen_share_let_me_check}\% in the first run; no reflection opener takes the place of \form{wait}.

\paragraph{Benchmarks beyond MATH-500.}
Each model in \Cref{tab:datasets} is sampled with its MATH-500 settings: AIME24 and AIME25 with 16 samples per problem, AMC23 with 8, GSM8K (first 500 problems) and MMLU-Pro with 2. MMLU-Pro is a fixed sample of 500 questions, 125 each from physics, chemistry, biology and engineering, scored with its official five-shot prompt and answer extraction; unlike the official protocol we sample rather than decode greedily, allow 8,192 tokens and give reasoning models the raw five-shot prompt instead of their chat template. The thinking models often reach the 16,384-token limit on AIME, so the table counts finished responses only. The AMC 2023 set has no licence or source stated on the Hugging Face dataset (competition problems \textcopyright{} MAA); like unlicensed models, it is used for this table only.

The driver sweep of \Cref{hyp:drivers} (KL coefficient and length penalty from a marker-rich base) was not run for compute (\Cref{app:prereg}).

\subsection{Further Analyses}
This section reports length and lineage controls and what the results mean for readers of markers.

\subsubsection{Length and Lineage Controls}
\label{sec:analysis}
Longer traces contain more of everything, so every rate is also computed per response and per 1k words; the ordering of models and checkpoints is the same under all three units (\Cref{sec:exp:source}).
\textbf{Supervised fine-tuning as a control.} If learning from demonstrations alone produced the zero-RL profile, the supervised fine-tuning trajectory of the same OLMo-3 base would show it too. It does not: rate, comma form and chains stay close to their starting values (\Cref{sec:exp:reshape}), and the prior keyword counts move little ($\rho = \res{bm.s2ctl.trend.rho_r1wait}$ for the count of \form{wait}, \res{bm.s2ctl.trend.rho_simplerl} for SimpleRL-Zoo's keywords).

\subsubsection{Consequence for Readers of Markers}
\label{sec:analysis:monitors}
\label{sec:analysis:drivers}

\Cref{hyp:monitor} predicted that monitors which read \form{wait} literally lose precision as RL makes the marker a more liberal alarm, while a monitor calibrated to the checkpoint does not. The signal-detection result does not by itself fix the direction, because sensitivity rises as false alarms rise: at the last checkpoint of OLMo-3 RL-Zero a revision marker follows \res{h2.s1.final.fa}\% of error-free prefixes (\form{wait} alone: \res{h2.s1.final.fa_wait}\%). The monitor comparison itself was not run, so the prediction is untested.

\subsubsection{A Running Example}
\label{app:case}
Problem \res{fig1.case.pid} is the first MATH-500 problem that the base, step 700 and the last checkpoint of OLMo-3 RL-Zero all solve, a selection rule fixed before reading the traces. In traces of similar length (\res{fig1.case.base.ntok}, \res{fig1.case.mid.ntok} and \res{fig1.case.last.ntok} tokens), \form{wait} occurs \res{fig1.case.base.nwait}, \res{fig1.case.mid.nwait} and \res{fig1.case.last.nwait} times, and reduplicated chains appear \res{fig1.case.base.nchain}, \res{fig1.case.mid.nchain} and \res{fig1.case.last.nchain} times. The base writes \emph{``Wait, is that correct?''} and \emph{``Wait, that can't be right.''}. The last checkpoint writes \emph{``Wait, but hold on.''} and \emph{``Wait, 2 is 2/1.''} (verbatim; the shortest marker-initial sentences of 4--70 characters and at least four words). At the base, the marker opens questions about a result. At the last checkpoint, it also opens restatements of values that were already correct.

\subsubsection{Sensitivity Analysis}
\label{sec:exp:sens}

The three main findings are robust to the counting choices (\Cref{tab:sens}). The zero-RL count retains a single counterexample at a threshold of 0.1 per 1k tokens, without the interval condition, on released runs only, without the reproduction gate and with the unlicensed run. The stricter 0.02 threshold adds one more, and every run with the earlier decode confirms the finding. The rise along OLMo-3 RL-Zero holds per token and per sentence, on finished responses and under greedy decoding. Its base rate also varies little across sampling seeds. Finally, our GRPO recipe consistently erases the seeded marker at both seed sizes and under the longer length cap.

\begin{table}[t]
  \centering\tabstyle\footnotesize
  \fitwidth{\begin{tabular}{@{}llr@{}}
    \toprule
    Finding & Analysis choice & Value \\
    \midrule
    \multirow{8}{*}{\shortstack[l]{Zero-RL keeps\\the marker absent\\(confirming / runs)}}
      & \cellcolor{gray!10}registered rule ($\ge$0.05 per 1k and above the base's CI) & \cellcolor{gray!10}\textbf{\res{sens.zr.reg.n_confirm}/\res{sens.zr.reg.n_runs}} \\
      & threshold 0.1 per 1k & \res{sens.zr.thr10.n_confirm}/\res{sens.zr.thr10.n_runs} \\
      & threshold 0.02 per 1k & \res{sens.zr.thr02.n_confirm}/\res{sens.zr.thr02.n_runs} \\
      & $\ge$0.05 per 1k, no CI comparison & \res{sens.zr.point.n_confirm}/\res{sens.zr.point.n_runs} \\
      & released runs only & \res{sens.zr.released.n_confirm}/\res{sens.zr.released.n_runs} \\
      & earlier decode, $T = 1.0$ & \res{sens.zr.t10.n_confirm}/\res{sens.zr.t10.n_runs} \\
      & without the reproduction gate & \res{sens.zr.nogate.n_confirm}/\res{sens.zr.nogate.n_runs} \\
      & with the unlicensed run & \res{sens.zr.withsec.n_confirm}/\res{sens.zr.withsec.n_runs} \\
    \addlinespace[0.5mm]\midrule
    \multirow{5}{*}{\shortstack[l]{OLMo-3 RL-Zero\\multiplies it\\(step 400 / base)}}
      & all responses, per 1k tokens & \res{h1.s1.step0400.a_boot} \\
      & finished responses only & \res{h1.s1.step0400.a_boot_fin} \\
      & per 100 sentences & \res{sens.s1.sent.ratio} \\
      & greedy decoding (per 1k; per 100 sent.) & \res{sens.s1.greedy.ratio}; \res{sens.s1.greedy_sent.ratio} \\
      & base rate over \res{sens.s1.seedbase.n} sampling seeds (per 1k) & \res{sens.s1.seedbase.min}--\res{sens.s1.seedbase.max} \\
    \addlinespace[0.5mm]\midrule
    \multirow{3}{*}{\shortstack[l]{GRPO erases a\\seeded marker\\(step 400 / SFT)}}
      & 512 seed traces, 3,072-token cap & \res{h1.seed.W512.trend.a_wait} \\
      & 64 seed traces, 3,072-token cap & \res{h1.seed.W64.trend.a_wait} \\
      & 512 seed traces, 8,192 + YaRN, finished only & \res{h1.seed.W512LY.trend.a_wait_fin} \\
    \bottomrule
  \end{tabular}}
  \vspace{-2mm}
  \caption{\textbf{The main findings hold across analysis choices.} Unregistered rows are descriptive.}
  \label{tab:sens}
  \vspace{-2mm}
\end{table}

\FloatBarrier
\section{Pre-registration and Analysis Details}
The theory behind the registered predictions, every prediction with its outcome, and the statistical procedures.

\subsection{Grammaticalization and What RL Could Do to a Marker}
\label{sec:theory}
This section summarises the theory and the predictions we derived from it.

\paragraph{Grammaticalization.}
A lexical item grammaticalizes when, in particular contexts, it comes to serve a grammatical function \citep{hopper2003grammaticalization}; the textbook case is the English motion phrase \form{going to} becoming a future marker.
\citet{lehmann2015thoughts} makes the process measurable with six parameters on two axes.
Along the \emph{paradigmatic} axis, a grammaticalizing sign loses \emph{integrity} (its meaning bleaches), gains \emph{paradigmaticity} (it joins a small, closed set of alternatives), and loses \emph{paradigmatic variability} (its use becomes obligatory in its slot).
Along the \emph{syntagmatic} axis, its \emph{scope} shrinks, its \emph{bondedness} to neighbouring material grows, and its \emph{syntagmatic variability} falls (its position becomes fixed).
\citet{bybee2003mechanisms} identifies frequency of use as the driver: repetition automates a construction, habituation bleaches its meaning, and the frequent form becomes autonomous from its source.
\citet{hopper1991principles} adds principles that hold during the change, of which two matter here: \emph{layering}, the coexistence of old and new uses of a form, and \emph{persistence}, the survival of traces of the old meaning in where the new form may occur.
Discourse markers such as \form{well} or \form{I mean} develop through closely related paths \citep{brinton2005lexicalization}.

\paragraph{What outcome-based RL could do to a marker.}
The theory supplies the vocabulary for this process. Its first lesson, that a grammatical marker needs a \emph{source} construction already in use, is the one that organises our findings (\Cref{sec:exp:source}).
Under outcome-based RL such as GRPO \citep{shao2024deepseekmath}, all tokens of a sampled trace share its advantage.
A marker that precedes a useful move, such as a re-check that sometimes catches an error, is therefore reinforced wherever it occurs in successful traces, not only where an error was present.
What RL reinforces is the contingency \gloss{marker $\rightarrow$ move}, not \gloss{error $\rightarrow$ marker}.
Repeated over many updates, this raises the marker's frequency, which is Bybee's driver.
It concentrates the marker in the slots where the move pays off (fixation), makes it the default entry to the move (obligatoriness, a smaller paradigm), and loosens its dependence on the presence of an error (bleaching).
We define two quantities for a marker $m$ at checkpoint $\ckpt{t}$.
Its \emph{report meaning} is how much $m$ tells about an error, $\Prob_t(\text{error} \mid m) - \Prob_t(\text{error})$.
We call $m$ an \emph{operator} when its effect on the model's next move is procedural and independent of its lexical meaning; its \emph{operator force} is the causal effect of emitting $m$ on that move, $\Prob_t(\text{revise} \mid \mathrm{do}(m)) - \Prob_t(\text{revise} \mid \mathrm{do}(m_0))$ for a neutral filler $m_0$ (in our experiments \tok{Next,}).
\Cref{sec:exp:alarm} measures the first with injected errors.

\begin{claimbox}
\thesistext
\end{claimbox}

\paragraph{Rival accounts.}
Two alternatives predict the same frequency curves and different everything else.
\circled{1}~\emph{Honest report}: RL improves error detection and the marker reports it. Then the marker's sensitivity $d'$ rises with its frequency, and inserting it after a correct step changes little, because the policy checks the state rather than obeying the word.
\circled{2}~\emph{Ritual}: noisy credit assignment stamps in constructions that do nothing, like the adventitious reinforcement of \citet{skinner1948superstition}. Then the marker is frequent but causally inert under insertion and deletion, and rituals should be rarer when larger groups make advantage estimates less noisy.
A third contrast is internal to linguistics: frequency drives bleaching in grammaticalization \citep{bybee2003mechanisms}, whereas the law of conformity states that frequent words change \emph{less} \citep{hamilton2016diachronic}. The two make opposite predictions about which markers change fastest.

\paragraph{Measurement.}
\Cref{sec:method} defines the indices formally, and \Cref{tab:operationalization} in \Cref{app:operationalization} maps every parameter to its measurement. Before the main run we fixed six predictions with their falsifiers; \Cref{app:prereg} states them and reports every outcome.
\begin{table}[h!]
  \centering
  \small
  \setlength{\tabcolsep}{3pt}
  \renewcommand{\arraystretch}{1.1}
  \begin{tabularx}{\linewidth}{@{}>{\raggedright\arraybackslash}p{2.1cm} >{\raggedright\arraybackslash}X c@{}}
    \toprule
    \textbf{Parameter (change)} & \textbf{Measurement on checkpoints} & \textbf{Hyp.} \\
    \midrule
    Frequency of use (context) & rate of \form{wait} and its paradigm per 1k tokens and per 100 sentences; known to rise \citep{deepseekai2025r1} & -- \\
    Syntagmatic variability ($\downarrow$, fixation) & sentence-initial share $\mathrm{SI}$, canonical-form share $\mathrm{ORTH}$, position entropy $H_{\mathrm{pos}}$ of \form{wait} & \Cref{hyp:fix} \\
    Bondedness ($\uparrow$), scope ($\downarrow$) & next-word entropy $H_{\mathrm{next}}$; tokens from marker to next action & \Cref{hyp:fix} \\
    Integrity ($\downarrow$, bleaching) & hit and false-alarm rates, $d'$ and $c$ on prefixes with vs.\ without an injected error; usage-embedding change (APD, PRT, JSD); substitutability of synonyms & \Cref{hyp:bleach} \\
    Paradigmatic variability ($\downarrow$), paradigmaticity ($\uparrow$) & entropy $H_{\mathrm{para}}$ and effective size $N_{\mathrm{eff}}$ of the forms opening revisions; policy-side entropy $H_{\mathrm{sub}}$; obligatoriness $\mathrm{OBL}$ & \Cref{hyp:oblig} \\
    Procedural function (operator) & risk difference and odds ratio of revising a correct step after inserted \tok{Wait,} vs.\ \tok{Next,} and synonyms, late vs.\ base policy & \Cref{hyp:operator} \\
    Frequency of use (driver) & index slopes under KL coefficient, group size and length penalty & \Cref{hyp:drivers} \\
    Layering (consequence) & precision and calibration of literal-reading monitors vs.\ step & \Cref{hyp:monitor} \\
    \bottomrule
  \end{tabularx}
  \caption{\textbf{Each grammaticalization parameter becomes a quantity measured on RL checkpoints.} Parameters follow \citet{lehmann2015thoughts}; arrows give the direction predicted under grammaticalization. Definitions are in \Cref{sec:method}.}
  \label{tab:operationalization}
\end{table}

\paragraph{Scope of the theory import.}
Grammaticalization is used as a source of predictions about a training process, not as a claim about how people process language or about the history of English. Its vocabulary (source, frequency, attrition) organises our measurements; most of its specific predictions about RL fail on our data, and we do not claim that RL is grammaticalization.

\paragraph{Discussion: Language change under optimisation is recipe-specific.}
Language change under optimisation operates as recipe-specific pressures on training data rather than a single cline. The grammaticalization vocabulary describes two phenomena we find: a marker requires a source, and its form can wear down. However, several registered predictions fail (\Cref{app:prereg}): the marker does not bleach, its form erodes instead of fixing, and the changes do not construct one path across recipes. Instead, RL multiplied a marker that mid-training supplied and erased one that light fine-tuning seeded.

\subsection{Pre-registered Predictions and Outcomes}
\label{app:prereg}
This section lists every prediction we registered before the main run, with its outcome.
We also pre-registered a decoding intervention on the alarm finding, Calibrated Wait (D50), which masks the marker after low-entropy steps; it was run on four models and did not pass its criteria, and its registration, every arm and all results are at \url{https://anonymous.4open.science/r/wait-provenance/} (\texttt{DECISIONS.md}, D50--D51).

\label{sec:exp:not}
\label{sec:exp:h2}
\label{sec:exp:h3}

The predictions below, their falsifiers and the order of the primary tests were fixed in the experiment plan before the main run; the gate criteria were amended once on 2026-09-30, after the pilot and before any number from the second or third series, and the label-free insertion outcome was declared before it was computed. The source test of \Cref{sec:exp:source} was formulated after the first marker-free runs and is reported as such. \Cref{tab:prereg} lists every outcome; the paragraphs after it give the numbers.
\begin{table*}[t]
  \centering
  \tabstyle
  \setlength{\tabcolsep}{3pt}
  \begin{tabularx}{\textwidth}{@{}>{\raggedright\arraybackslash}p{2.9cm} >{\raggedright\arraybackslash}X >{\raggedright\arraybackslash}X >{\raggedright\arraybackslash}p{2.3cm}@{}}
    \toprule
    \textbf{Prediction} & \textbf{Pre-registered test} & \textbf{Result} & \textbf{Verdict} \\
    \midrule
    \hyp{1} Fixation & sentence-initial and \tok{Wait,} shares rise with step ($\rho \geq 0.5$ on 2 of 3 series) & S1: SI \rt{h1.s1.base.si}$\rightarrow$\rt{h1.s1.final.si}\% ($\rho = \rt{h1.s1.trend.rho_si}$); \tok{Wait,} \rt{h1.s1.base.orth}$\rightarrow$\rt{h1.s1.final.orth}\% ($\rho = \rt{h1.s1.trend.rho_orth}$); S2: \tok{Wait,} $\rho = \rt{h1.s2.trend.rho_orth}$ & falsified: position inherited, form erodes (S1) or is restored (S2) \\
    \hyp{2} Bleaching & false alarms rise, $d'$ does not (falsified if $\Delta d'$ lower bound $> 0.25$) & S1: $d'$ \rt{h2.s1.base.dprime}$\rightarrow$\rt{h2.s1.final.dprime} [\rt{h2.s1.final.dprime.lo}, \rt{h2.s1.final.dprime.hi}]; $c$ \rt{h2.s1.base.crit}$\rightarrow$\rt{h2.s1.final.crit}; FA \rt{h2.s1.base.fa}$\rightarrow$\rt{h2.s1.final.fa}\% & \textbf{falsified}: a more sensitive, more liberal alarm \\
    \hyp{3} Paradigm loss, obligatoriness & paradigm entropy $\rho \leq -0.5$; $\geq 60\%$ of revisions carry a marker & $\rho(H_{\mathrm{para}})$: S1 \rt{h1.s1.trend.rho_hpara}, S2 \rt{h1.s2.trend.rho_hpara}, 32B \rt{h1.s9.trend.rho_hpara}; obligatoriness at the end of S1 \rt{h1.s1.final.obl}\% & paradigm narrows on S1 only; obligatoriness falsified \\
    \hyp{4} Operator & \tok{Wait,} $-$ \tok{Next,} $\geq 10$ pts, OR $\geq 2$, $\geq 1.5\times$ the base model; same trend under both prefix sources & judge: \rt{h4.s1.final.rd_op_late} pts, $R_{\mathrm{amp}} = \rt{h4.s1.trend.r_amp_late}$; second judge \rt{h4.s1.final.rd_op_late_oss} pts; label-free value change \rt{h4.s1.final.rdval_op_late} [\rt{h4.s1.final.rdval_op_late.lo}, \rt{h4.s1.final.rdval_op_late.hi}], but its detector fails its check (precision \rt{gold.s1.vc_precision_qwen3}\%) & inconclusive: judge result exploratory (human $\kappa = \rt{gold.v2.kappabin_human_judge}$), label-free outcome invalid \\
    \hyp{5} Drivers & KL and length pressure set the speed (controlled GRPO) & controlled runs from a marker-rich base not run & untested \\
    \hyp{6} Monitors & literal readers lose precision with RL; a calibrated monitor does not & not run & untested \\
    Cline across series (gate) & $\geq 2$ of \{paradigm $\downarrow$, judged reliability $\downarrow$, attrition $\uparrow$\} with $|\rho| \geq 0.7$ on $\geq 2$ of 3 series, flat across pretraining & S1 3 of 3; S2 0 of 3; Qwen2.5-1.5B no marker; pretraining branches not flat; along SFT the indices barely move (comma drop \rt{h1.s2ctl.step5000.commadrop}$\rightarrow$\rt{h1.s2ctl.step43000.commadrop}\%) & \textbf{not met} \\
    Keyword counts saturate (\S5b) & late half: keyword $|\rho| < 0.5$ while $\geq 2$ indices keep $|\rho| \geq 0.7$ & keywords $\rho = \rt{bm.s1.late_half.rho_r1kw}$, \rt{bm.s1.late_half.rho_simplerl}; ours \rt{h1.s1.late_half.rho_orth}, \rt{h1.s1.late_half.rho_hpara} & falsified: indices add constructs, not a better count \\
    Liberal criterion (D41, before the 32B data) & $c$ falls, $d'$ does not, on 32B Think & $c$ \rt{h2.s9.step0100.crit}$\rightarrow$\rt{h2.s9.late.crit}; $d'$ \rt{h2.s9.step0100.dprime}$\rightarrow$\rt{h2.s9.late.dprime} & falsified: the liberal shift is specific to zero-RL \\
    Seeded marker grows (D43, before the runs) & a marker seeded by fine-tuning on 512 traces is amplified by RL ($A > 1$) & \tok{Wait,} \rt{h1.seed.W512.sft.freq4}$\rightarrow$\rt{h1.seed.W512.step0400.freq4} per 1k ($A = \rt{h1.seed.W512.trend.a_wait}$); \tok{Next,} arm keeps \rt{h1.seed.X512.trend.a_nextopen} & \textbf{falsified}: RL erases the seeded marker \\
    Marker-free arms stay free (D43, P-seed1) & with \form{wait} deleted from the seed traces, \form{wait} stays $\leq 0.05$ per 1k after RL & D512: \rt{h1.seed.D512.sft.r2}$\rightarrow$\rt{h1.seed.D512.step0400.r2}; D64: \rt{h1.seed.D64.sft.r2}$\rightarrow$\rt{h1.seed.D64.step0400.r2} per 1k & holds \\
    Amplification in a controlled fork (D48, before the matched evaluation) & our GRPO recipe from the OLMo-3 base amplifies \form{wait} ($A(400)$ lower bound $> 1.2$) or not (CI includes 1, or upper bound $< 1.2$); disagreeing measures $\Rightarrow$ undetermined & $\beta{=}0.001$: $A = \rt{h1.fork_ref1m.step0400.a_boot}$ (all), \rt{h1.fork_ref1m.step0400.a_boot_fin} (fin.); $\beta{=}0$: \rt{h1.fork_b0m.step0400.a_boot}, \rt{h1.fork_b0m.step0400.a_boot_fin} & undetermined; $\beta{=}0$: none, recipes differ \\
    Length control for the transplant (D46, D46b) & finished-only $A(400) \ge 0.5$: length artefact; $< 0.2$: marker penalised & $A = \rt{h1.seed.W512LY.trend.a_wait_fin}$ (fin.), \rt{h1.seed.W512LY.trend.a_wait} (all) & marker penalised \\
    Dose (D43, P-seed4) & $A_{\mathrm{wait}}$ rises with the number of seed traces (W64 vs W512) & $A = \rt{h1.seed.W64.trend.a_wait}$ (W64) vs \rt{h1.seed.W512.trend.a_wait} (W512) & not shown: RL removes the marker at both doses \\
    \midrule
    Source (added after the first marker-free runs) & no zero-RL run from a base below 0.05 per 1k ends above 0.5 per 1k & \rt{h1.zr.n_runs} runs, highest last checkpoint \rt{h1.zr.ce.gr3_4.rate} per 1k (General-Reasoner-Qwen3-4B; the stricter counting rule D47 flags it as installing the marker); registered 7B replication: \rt{h1.s7r7b.final.nwait} occurrences in \rt{h1.s7r7b.final.ntok} tokens at step 400 & holds \\
    \bottomrule
  \end{tabularx}
  \caption{\textbf{Of six pre-registered hypotheses, three fail, one is unsupported and two were not run; further registered tests fail too.} Tests and thresholds as fixed before the main run (\Cref{app:stats}). S1: OLMo-3 RL-Zero; S2: OLMo-3 Think; 32B: OLMo-3.1 32B Think; SI and \tok{Wait,} shares in \%; FA: false-alarm rate.}
  \label{tab:prereg}
\end{table*}

\paragraph{Counting zero-RL runs (D47, D47a).}
Before evaluating the released zero-RL models, we fixed when a run counts against the source claim: its base writes \form{wait} less than 0.05 times per 1k tokens, its final model at least 0.05 times, and the final model's exact 95\% lower bound lies above the base's upper bound (D47). After the first non-Qwen runs reached far lower accuracy than their authors report, and before any re-evaluation, we added a reproduction rule (D47a). A released run counts only if our evaluation reaches at least half of the MATH-500 accuracy its authors publish and fewer than 20\% of its responses are truncated; our own runs need only the truncation condition. Runs below that are marked as not reproduced, and runs whose authors publish no MATH-500 accuracy are marked n.r.; both are listed in \Cref{tab:provenance} and counted neither way. Under these rules \res{h1.zr.n_runs} runs count, with \res{h1.zr.n_counter} counterexamples. The only counted run whose final model writes \form{wait} at all in a genuine use is SimpleRL-Zoo on Mistral-Small-24B (\res{h1.bb_sz_m24.final.nwait} matches in \res{h1.bb_sz_m24.final.ntok} tokens; base: none). Read verbatim, its few genuine uses are all \emph{``But wait!''} written after the boxed final answer, never a pivot inside the reasoning; its remaining matches lie inside degenerate token sequences. DAPO-Qwen-32B is marked n.r.: a third-party evaluation reports 71.8\% MATH-500 accuracy for this checkpoint \citep{hu2025openreasonerzero}, and ours gives \res{ctl.s8dapo.final.acc}\%; we keep the rule as registered.
Our single-seed GRPO run from Qwen3-4B-Base (D49) counts as a confirming run: its rate goes from \res{h1.bb_q3r1_base.base.r2} per 1k tokens at the base to \res{h1.bb_q3r1.step0400.r2} at step 400, after a transient \res{h1.bb_q3r1.step0100.r2} at steps 100 and 200; with one seed we do not separate the base from the recipe.
\begin{table*}[t]
  \centering
  \tabstyle\footnotesize\renewcommand{\arraystretch}{0.97}
  \setlength{\tabcolsep}{2pt}
  \fitwidth{\begin{tabular}{@{}ll cc cc@{}}
    \toprule
    & & \multicolumn{2}{c}{\textbf{MATH-500 (\%)}} & \multicolumn{2}{c}{\textbf{\form{wait} per 1k tokens}} \\
    \cmidrule(lr){3-4} \cmidrule(lr){5-6}
    \textbf{Model} & \textbf{Training step} & before & after & before & after \\
    \midrule
    \multicolumn{6}{@{}l}{\textsc{rl from models that do not use the marker (zero-rl from a base, except the last row)}} \\
    DAPO-Qwen-32B\textsuperscript{n.r.} & DAPO, Qwen2.5-32B & 35.8 & 76.4 & $<$.01\ci{$<$.01}{0.02} & \nsig{$<$.01\ci{$<$.01}{$<$.01}} \\
    Open-Reasoner-Zero-32B$^\ast$ & PPO, Qwen2.5-32B & 10.8 & 91.4 & 0\ci{0}{$<$.01} & \nsig{$<$.01\ci{$<$.01}{$<$.01}} \\
    REF-s1 (ours) & GRPO, Qwen2.5-Math-1.5B & 19.1 & 74.4 & $<$.01\ci{$<$.01}{$<$.01} & \nsig{0\ci{0}{$<$.01}} \\
    REF-s2 (ours) & GRPO, Qwen2.5-Math-1.5B & 17.8 & 74.2 & $<$.01\ci{$<$.01}{$<$.01} & \nsig{0\ci{0}{$<$.01}} \\
    REF-7B (ours) & GRPO, Qwen2.5-Math-7B & 16.9 & 76.0 & $<$.01\ci{$<$.01}{$<$.01} & \nsig{0\ci{0}{$<$.01}} \\
    Qwen-2.5-0.5B-SimpleRL-Zoo & GRPO, Qwen2.5-0.5B & 8.6 & 30.8 & $<$.01\ci{$<$.01}{$<$.01} & \nsig{$<$.01\ci{$<$.01}{$<$.01}} \\
    Qwen-2.5-1.5B-SimpleRL-Zoo$^\circ$ & GRPO, Qwen2.5-1.5B & 2.6 & 12.6 & $<$.01\ci{$<$.01}{$<$.01} & \nsig{$<$.01\ci{$<$.01}{$<$.01}} \\
    Qwen-2.5-7B-SimpleRL-Zoo & GRPO, Qwen2.5-7B & 48.2 & 77.2 & 0\ci{0}{0.01} & \nsig{0\ci{0}{0.01}} \\
    Open-Reasoner-Zero-7B & PPO, Qwen2.5-7B & 13.2 & 81.8 & $<$.01\ci{$<$.01}{0.02} & \nsig{0\ci{0}{$<$.01}} \\
    Qwen-2.5-Math-7B-SimpleRL-Zoo & GRPO, Qwen2.5-Math-7B & 43.0 & 75.6 & 0\ci{0}{$<$.01} & \nsig{0\ci{0}{0.01}} \\
    Qwen2.5-Math-7B-Oat-Zero & Dr.\ GRPO, Qwen2.5-Math-7B & 35.8 & 79.8 & 0\ci{0}{$<$.01} & \nsig{0\ci{0}{0.01}} \\
    Qwen-2.5-14B-SimpleRL-Zoo$^\dagger$ & GRPO, Qwen2.5-14B & 53.2 & 82.6 & 0\ci{0}{0.01} & \nsig{0\ci{0}{0.01}} \\
    Qwen-2.5-32B-SimpleRL-Zoo & GRPO, Qwen2.5-32B & 50.4 & 83.8 & 0\ci{0}{0.01} & \nsig{0\ci{0}{0.01}} \\
    Mistral-7B-v0.1-SimpleRL-Zoo$^\circ$ & GRPO, Mistral-7B-v0.1 & 0.4 & 5.0 & 0.07\ci{0.02}{0.16} & 0\ci{0}{$<$.01}\,\textcolor{cOurs}{$\downarrow$} \\
    Mistral-Small-24B-SimpleRL-Zoo & GRPO, Mistral-Small-24B & 21.4 & 75.8 & 0\ci{0}{0.01} & \nsig{$<$.01\ci{$<$.01}{$<$.01}} \\
    DeepSeek-Math-7B-SimpleRL-Zoo$^\circ$ & GRPO, DeepSeek-Math-7B & 1.4 & 19.2 & 0\ci{0}{0.02} & \nsig{0\ci{0}{0.01}} \\
    General-Reasoner-Qwen2.5-7B & GRPO, Qwen2.5-7B & 61.6 & 75.6 & 0\ci{0}{0.01} & \nsig{0\ci{0}{$<$.01}} \\
    General-Reasoner-Qwen2.5-14B & GRPO, Qwen2.5-14B & 67.0 & 77.0 & 0\ci{0}{$<$.01} & \nsig{0\ci{0}{$<$.01}} \\
    General-Reasoner-Qwen3-4B & GRPO, Qwen3-4B & 71.2 & 79.0 & $<$.01\ci{$<$.01}{0.01} & \cellcolor{cCmpA!10}0.29\ci{0.24}{0.34}\,\textcolor{cInterv}{$\uparrow$} \\
    General-Reasoner-Qwen3-14B & GRPO, Qwen3-14B & 79.2 & 85.2 & 0\ci{0}{0.01} & \cellcolor{cCmpA!2}0.04\ci{0.02}{0.06}\,\textcolor{cInterv}{$\uparrow$} \\
    One-Shot-RLVR-Qwen2.5-Math-1.5B-pi1 & GRPO (1 example), Qwen2.5-Math-1.5B & 32.8 & 72.6 & 0\ci{0}{$<$.01} & \nsig{0\ci{0}{0.01}} \\
    Qwen2.5-Math-1.5B-Oat-Zero & Dr.\ GRPO, Qwen2.5-Math-1.5B & 24.6 & 72.6 & $<$.01\ci{$<$.01}{$<$.01} & \nsig{0\ci{0}{0.01}} \\
    REF-Qwen3-4B (ours) & GRPO, Qwen3-4B & 41.2 & 84.1 & 0.02\ci{0.01}{0.03} & \cellcolor{cCmpA!2}\nsig{0.03\ci{0.02}{0.04}} \\
    Qwen2.5-1.5B-Open-R1-GRPO-math-v1$^\dagger$ & GRPO, Qwen2.5-1.5B-Inst. & 42.0 & 45.9 & $<$.01\ci{$<$.01}{$<$.01} & \nsig{$<$.01\ci{$<$.01}{$<$.01}} \\
    \midrule
    \multicolumn{6}{@{}l}{\textsc{fine-tuned on reasoning traces (distillation or cold start)}} \\
    QwQ-32B & cold start+RL, Qwen2.5-32B & 35.8 & 89.6 & $<$.01\ci{$<$.01}{0.02} & \cellcolor{cCmpA!27}4.42\ci{4.34}{4.49}\,\textcolor{cInterv}{$\uparrow$} \\
    DeepSeek-R1-Distill-Qwen-32B & SFT, Qwen2.5-32B & 35.8 & 89.5 & $<$.01\ci{$<$.01}{0.02} & \cellcolor{cCmpA!25}3.25\ci{3.18}{3.32}\,\textcolor{cInterv}{$\uparrow$} \\
    Phi-4-reasoning & SFT, Phi-4 & 82.0 & 93.0 & 0\ci{0}{$<$.01} & \cellcolor{cCmpA!20}1.56\ci{1.50}{1.62}\,\textcolor{cInterv}{$\uparrow$} \\
    Phi-4-reasoning-plus & SFT+RL, Phi-4 & 82.0 & 91.1 & 0\ci{0}{$<$.01} & \cellcolor{cCmpA!24}2.58\ci{2.51}{2.65}\,\textcolor{cInterv}{$\uparrow$} \\
    Qwen3-8B & SFT+RL, Qwen3-8B & 66.5 & 84.9 & 0\ci{0}{$<$.01} & \cellcolor{cCmpA!27}3.93\ci{3.87}{4.00}\,\textcolor{cInterv}{$\uparrow$} \\
    Qwen3-14B & SFT+RL, Qwen3-14B & 66.5 & 87.9 & 0.03\ci{0.01}{0.05} & \cellcolor{cCmpA!25}3.03\ci{2.98}{3.10}\,\textcolor{cInterv}{$\uparrow$} \\
    Qwen3-30B-A3B & SFT+RL, Qwen3-30B-A3B & 52.2 & 86.8 & 0.02\ci{0.01}{0.03} & \cellcolor{cCmpA!24}2.83\ci{2.77}{2.88}\,\textcolor{cInterv}{$\uparrow$} \\
    OpenThinker-7B & SFT, Qwen2.5-7B-Inst. & 74.8 & 79.9 & 0\ci{0}{$<$.01} & \cellcolor{cCmpA!27}4.05\ci{3.98}{4.13}\,\textcolor{cInterv}{$\uparrow$} \\
    Bespoke-Stratos-7B & SFT, Qwen2.5-7B-Inst. & 74.8 & 79.2 & 0\ci{0}{$<$.01} & \cellcolor{cCmpA!25}3.19\ci{3.11}{3.26}\,\textcolor{cInterv}{$\uparrow$} \\
    \midrule
    \multicolumn{6}{@{}l}{\textsc{training stages of olmo-3-1025-7b (before = previous stage)}} \\
    Pretraining & stage 1, step 707k & 0.0 & 0.4 & 0.04\ci{0.03}{0.05} & \cellcolor{cCmpA!2}\nsig{0.03\ci{0.02}{0.04}} \\
    Pretraining & stage 1, step 1.41M & 0.4 & 0.1 & 0.03\ci{0.02}{0.04} & \cellcolor{cCmpA!3}\nsig{0.04\ci{0.03}{0.06}} \\
    Mid-training & stage 2 & 0.1 & 35.7 & 0.04\ci{0.03}{0.06} & \cellcolor{cCmpA!27}4.03\ci{3.92}{4.14}\,\textcolor{cInterv}{$\uparrow$} \\
    Base & stage 3 & 35.7 & 51.2 & 4.03\ci{3.92}{4.14} & \cellcolor{cCmpA!28}4.94\ci{4.85}{5.03}\,\textcolor{cInterv}{$\uparrow$} \\
    \midrule
    \multicolumn{6}{@{}l}{\textsc{rl from a model that uses the marker (after = last checkpoint)}} \\
    Olmo-3-7B-RL-Zero-Math & RL-Zero & 52.3\std{1.2} & 84.9\std{0.9} & 5.07\std{0.17} & \cellcolor{cCmpA!35}\best{12.68}\std{1.55}\,\textcolor{cInterv}{$\uparrow$} \\
    Olmo-3-7B-Think & SFT, DPO, RL & 90.0 & 91.0 & 3.86\ci{3.79}{3.94} & \cellcolor{cCmpA!27}4.52\ci{4.44}{4.59}\,\textcolor{cInterv}{$\uparrow$} \\
    Olmo-3.1-32B-Think & SFT, DPO, RL & 91.4 & 93.0 & 4.52\ci{4.46}{4.58} & \cellcolor{cCmpA!27}4.23\ci{4.17}{4.29}\,\textcolor{cOurs}{$\downarrow$} \\
    MiMo-7B-RL-Zero & RL-Zero, MiMo-7B & 65.8 & 76.4 & 4.95\ci{4.86}{5.04} & \cellcolor{cCmpA!30}\second{6.62}\ci{6.53}{6.71}\,\textcolor{cInterv}{$\uparrow$} \\
    DeepScaleR-1.5B-Preview & RL, R1-Distill-1.5B & 80.1 & 89.1 & 5.04\ci{4.95}{5.12} & \cellcolor{cCmpA!27}4.15\ci{4.07}{4.23}\,\textcolor{cOurs}{$\downarrow$} \\
    Light-R1-14B-DS & RL, R1-Distill-14B & 89.0 & 88.4 & 3.77\ci{3.70}{3.85} & \cellcolor{cCmpA!27}4.01\ci{3.93}{4.08}\,\textcolor{cInterv}{$\uparrow$} \\
    \midrule
    \multicolumn{6}{@{}l}{\textsc{newest generation: qwen3.5 (before = its base)}} \\
    Qwen3.5-0.8B & post-training & 43.8 & 48.1 & 1.96\ci{1.92}{1.99} & \cellcolor{cCmpA!21}1.83\ci{1.80}{1.86}$^\P$\,\textcolor{cOurs}{$\downarrow$} \\
    Qwen3.5-2B & post-training & 74.5 & 76.8 & 1.31\ci{1.27}{1.34} & \cellcolor{cCmpA!16}0.86\ci{0.84}{0.89}\,\textcolor{cOurs}{$\downarrow$} \\
    Qwen3.5-4B & post-training & 89.6 & 88.2 & 1.15\ci{1.11}{1.20} & \cellcolor{cCmpA!26}3.82\ci{3.78}{3.85}$^\P$\,\textcolor{cInterv}{$\uparrow$} \\
    Qwen3.5-9B & post-training & 88.1 & 92.1 & 0.93\ci{0.89}{0.98} & \cellcolor{cCmpA!24}2.94\ci{2.91}{2.97}\,\textcolor{cInterv}{$\uparrow$} \\
    \midrule
    \multicolumn{6}{@{}l}{\textsc{olmo-2 7b lineage (before = the base)}} \\
    OLMo-2-1124-7B-SFT & SFT, OLMo-2-7B & 4.1 & 25.9 & 0.16\ci{0.13}{0.19} & 0\ci{0}{$<$.01}\,\textcolor{cOurs}{$\downarrow$} \\
    OLMo-2-1124-7B-DPO & DPO after SFT & 4.1 & 35.0 & 0.16\ci{0.13}{0.19} & 0\ci{0}{$<$.01}\,\textcolor{cOurs}{$\downarrow$} \\
    OLMo-2-1124-7B-Instruct & RLVR after DPO & 4.1 & 37.5 & 0.16\ci{0.13}{0.19} & 0\ci{0}{$<$.01}\,\textcolor{cOurs}{$\downarrow$} \\
    \bottomrule
  \end{tabular}}
  \caption{\textbf{Full provenance table: all but one reproduced zero-RL run stay without \form{wait}; documented routes to it otherwise pass through training data.} Format as in \Cref{tab:main}. Models without a measured starting point are in \Cref{tab:nobase}. Shading: \form{wait} per 1k after training (log scale, darker = more); \textcolor{cInterv}{$\uparrow$}/\textcolor{cOurs}{$\downarrow$}: rise/fall with non-overlapping intervals; grey rows: our runs.}
  \label{tab:provenance}
  \par\vspace{2pt}\begin{minipage}{\textwidth}\footnotesize $^\dagger$Secondary, not counted: the release has no licence text (the open run also starts from an instruct model). $^\diamond$Run with the vendor's required Chinese system prompt (its evaluation note says the system prompt is empty; without it the model degenerates); \form{wait} per 1k thinking tokens. $^\ast$Base and final use the ORZ prompt template, under which the base scores low; the accuracy gain partly reflects the template. $^\circ$Not reproduced: accuracy below half the published value or at least 20\% truncated; not counted (D47a, \Cref{app:prereg}). \textsuperscript{n.r.}No MATH-500 reference published by its authors; counted neither way (D47a, \Cref{app:prereg}). $^\ddagger$Base gated (manual approval); not used. $^\S$Base gated, not measured. $^\P$At least 20\% of responses reach the length limit; not marked, direction from finished responses in \Cref{sec:exp:source}. QwQ-32B's cold-start data are not public \citep{qwen2025qwq}.\end{minipage}
\end{table*}

\paragraph{Grammaticalization or frequency?}
Frequency alone predicts only that a marker is used more often.
Grammaticalization predicts, beyond frequency, that the marker becomes fixed in position, obligatory in its slot and one of fewer alternatives, that its link to errors weakens while its causal force grows, and that these changes do not reverse (unidirectionality, a strong tendency in language change; \citealp{haspelmath1999why}).
Each of these can be measured on checkpoints and each can fail.

\noindent Six predictions were derived from the thesis. Each is evaluated per series and holds if it holds on at least two of three series; the falsifiers and the order in which the primary tests are run were fixed before any result (\Cref{app:stats}).

\begin{hypothesis}[Fixation]\label{hyp:fix}
Along RL steps, reasoning markers become positionally and orthographically fixed: the sentence-initial share of \form{wait} and the share of its capitalised-comma form \tok{Wait,} rise, and the entropy of its position falls. \emph{Falsified if} the Spearman correlation with step is below 0.5 on at least two of three series, or the supervised fine-tuning trajectory of the same base shows the same trend at the same rate.
\end{hypothesis}

\begin{hypothesis}[Bleaching]\label{hyp:bleach}
Markers decouple from error presence: on prefixes with and without an injected error, the false-alarm rate $\Prob(\text{marker} \mid \text{no error})$ rises while the sensitivity $d'$ does not. \emph{Falsified if} $d'$ rises in step with the false-alarm rate (the marker is an honest detector; operationally, the lower 95\% bound of $\Delta d'$ exceeds $+0.25$), or the false-alarm rate is flat.
\end{hypothesis}

\begin{hypothesis}[Paradigm loss and obligatoriness]\label{hyp:oblig}
The paradigm of forms that open revisions shrinks (its entropy falls) and some marker becomes near-obligatory before revision events. \emph{Falsified if} the paradigm entropy is flat or rising ($\rho > -0.5$ with step), or fewer than 60\% of revisions carry a marker at the last checkpoint on all series.
\end{hypothesis}

\begin{hypothesis}[Operator]\label{hyp:operator}
Inserting \tok{Wait,} after a \emph{correct} step causes revision of that step far more often in late RL checkpoints than in the base model on the same prefixes, and more often than the synonyms \tok{Hold on,} and \tok{Hmm,}; the gap widens with RL. The matched control is the neutral filler \tok{Next,} at the same position. \emph{Falsified if} at the last checkpoint the risk difference to \tok{Next,} is below 10 points or the odds ratio is below 2 (interval including 1), or the effect is less than 1.5 times the base model's, or the synonyms equal the exact form at every checkpoint.
\end{hypothesis}

\begin{hypothesis}[Drivers]\label{hyp:drivers}
In controlled GRPO, KL and length pressure set the speed of the cline: a lower KL coefficient speeds fixation and bleaching, a length penalty speeds erosion, and larger groups leave fewer inert rituals. \emph{Falsified if} the condition slopes overlap within seed variation for all knobs.
\end{hypothesis}

\begin{hypothesis}[Monitors]\label{hyp:monitor}
Literal-reading judges and hedge-counting uncertainty heuristics lose precision against true error presence as RL proceeds, while a bleaching-aware monitor does not. \emph{Falsified if} the literal monitors' precision is flat across checkpoints, or the bleaching-aware monitor drifts as much as they do.
\end{hypothesis}

\noindent The theory also makes predictions that neither rival makes: usage clusters of \form{wait} split into an older lexical and a newer operator sense whose share grows (layering); no index reverses once it has moved (unidirectionality); and the operator keeps a trace of its source by preceding revision moves more than a neutral connective does (persistence). We report these as secondary tests (\Cref{sec:analysis}).

\paragraph{Fixation (\hyp{1}) and obligatoriness (\hyp{3}).}
The position of \form{wait} was already fixed before RL, and its canonical form erodes along RL-Zero instead of spreading (\Cref{sec:exp:reshape}); at the last checkpoint \res{h1.s1.late.obl}\% of revisions carry a marker, below the 60\% we required.

\paragraph{Bleaching (\hyp{2}): falsified.}
\Cref{hyp:bleach} predicted that the marker would decouple from errors: false alarms would rise while sensitivity did not. On prefixes with and without an injected error in OLMo-3 RL-Zero, a revision marker (the registered outcome, any form of the paradigm) appears within two sentences after \res{h2.s1.final.fa}\% of error-free prefixes at the final checkpoint (base \res{h2.s1.base.fa}\%) and the criterion falls from \res{h2.s1.base.crit} to \res{h2.s1.final.crit}, but $d'$ \emph{rises}, from \res{h2.s1.base.dprime} to \res{h2.s1.final.dprime} (95\% CI [\res{h2.s1.final.dprime.lo}, \res{h2.s1.final.dprime.hi}]); for \form{wait} alone, false alarms rise from \res{h2.s1.base.fa_wait}\% to \res{h2.s1.final.fa_wait}\% and $d'$ from \res{h2.s1.base.dprime_wait} to \res{h2.s1.final.dprime_wait}. The marker becomes a more liberal alarm that is also more sensitive to real errors; it does not lose its link to them. What weakens is its link to revision: fewer markers are followed by a revision event (\Cref{sec:exp:reshape}). A reader that takes \form{wait} as evidence of an error is calibrated to a criterion that RL moves. The liberal shift is specific to zero-RL. Along OLMo-3.1 32B Think, whose RL follows supervised fine-tuning, the criterion moves the other way, from \res{h2.s9.step0100.crit} to \res{h2.s9.late.crit}, and $d'$ stays within its interval (\res{h2.s9.step0100.dprime} to \res{h2.s9.late.dprime}, 95\% CI [\res{h2.s9.late.dprime.lo}, \res{h2.s9.late.dprime.hi}]), which falsifies the liberal-criterion prediction we registered for that series. Along OLMo-3 7B Think the registered direction holds only weakly: the criterion moves from \res{h2.s2.base.crit} to \res{h2.s2.late.crit} (95\% CI [\res{h2.s2.late.crit.lo}, \res{h2.s2.late.crit.hi}]) and $d'$ from \res{h2.s2.base.dprime} to \res{h2.s2.late.dprime}.

\begin{figure}[t]
  \centering
  \includegraphics[width=\linewidth]{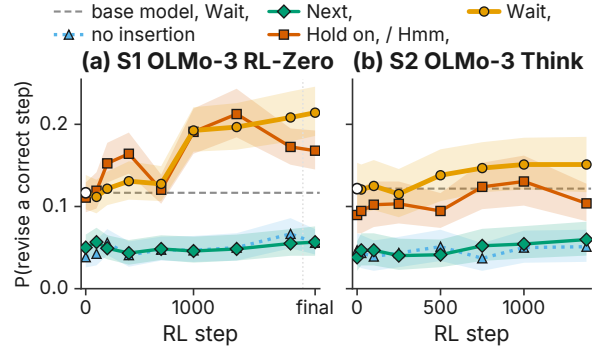}
  \caption{\textbf{By the judges' labels, an inserted \tok{Wait,} pushes every checkpoint toward revision, the base included.} Judged probability of revising a correct step per inserted form along (a) OLMo-3 RL-Zero and (b) OLMo-3 7B Think, with 95\% cluster-bootstrap bands. Exploratory: the judge's sensitivity against the human audit is low (\Cref{app:stats}).}
  \label{fig:sdt}
\end{figure}

\paragraph{Operator (\hyp{4}): an inserted \tok{Wait,} changes no more values than a neutral filler.}
\Cref{hyp:operator} predicted that an inserted \tok{Wait,} after a correct step would become an operator: that late checkpoints would revise the step far more often than after the neutral filler \tok{Next,}, and far more than the base model does. We pre-registered a label-free outcome for this test before computing it: whether the continuation restates the quantity of the last step with a different value.
At the last checkpoint of OLMo-3 RL-Zero, \tok{Wait,} changes no more values than \tok{Next,} (risk difference \res{h4.s1.late.rdval_op_late} points, 95\% CI [\res{h4.s1.late.rdval_op_late.lo}, \res{h4.s1.late.rdval_op_late.hi}]; prefixes from base-model traces: \res{h4.s1.late.rdval_op_base} [\res{h4.s1.late.rdval_op_base.lo}, \res{h4.s1.late.rdval_op_base.hi}]), and the difference does not grow with RL ($\rho = \res{h4.s1.trend.rho_rdval_late}$). The test uses the prefixes whose last step contains an extractable equation (\res{h4.s1.late.nexcl_val_late} of 300 excluded). The Qwen2.5-32B models give the same null: in the base, \tok{Wait,} and \tok{Next,} are followed by a changed value in \res{h4.s8base.final.prevval_wait}\% and \res{h4.s8base.final.prevval_next}\% of windows.
By the judges' labels, in contrast, \tok{Wait,} is followed by a revision more often than \tok{Next,} at every checkpoint: \res{h4.s1.late.rd_op_late} points at the last checkpoint against \res{h4.s1.base.rd_op_late} for the base model on the same prefixes ($R_{\mathrm{amp}} = \res{h4.s1.trend.r_amp_late}$; second judge \res{h4.s1.late.rd_op_late_oss} points), and \res{h4.s8base.final.rd_op} points in the Qwen2.5-32B base, which never writes the marker itself (\Cref{tab:insertion,fig:sdt}). A replicate with freshly sampled base-model prefixes reproduces the rise: \res{h4.s1seed1.base.rd_op_base} points at the base, \res{h4.s1seed1.late.rd_op_base} at step 1900 and \res{h4.s1seed1.final.rd_op_base} at the final checkpoint (95\% CI [\res{h4.s1seed1.final.rd_op_base.lo}, \res{h4.s1seed1.final.rd_op_base.hi}]); second judge \res{h4.s1seed1.base.rd_op_base_oss}, \res{h4.s1seed1.late.rd_op_base_oss} and \res{h4.s1seed1.final.rd_op_base_oss}. These judged effects are exploratory: on 40 blind insertion windows with expert consensus labels, the primary judge agrees at $\kappa = \res{gold.v2.kappabin_human_judge}$ (sensitivity \res{gold.v2.sens_judge}\%, specificity \res{gold.v2.spec_judge}\%), below the threshold we fixed before the audit.
The two outcomes rarely coincide. On the same windows at the last checkpoint, only \res{h4.s1.late.p_vc_given_rev_late}\% of judged revisions are flagged as value changes (\res{h4.s1.late.n_both_late} of \res{h4.s1.late.n_rev_late}), and only \res{h4.s1.late.p_rev_given_vc_late}\% of flagged value changes are judged revisions; in the Qwen2.5-32B base the shares are \res{h4.s8base.final.p_vc_given_rev}\% and \res{h4.s8base.final.p_rev_given_vc}\%. The registered check of the value-change detector explains part of this: an open-weight LLM rater (Qwen3-32B) confirms a value change in only \res{gold.s1.vc_n_yes} of \res{gold.s1.vc_n_checked} windows the detector flags (precision \res{gold.s1.vc_precision_qwen3}\%, 95\% CI [\res{gold.s1.vc_precision_qwen3.lo}, \res{gold.s1.vc_precision_qwen3.hi}]), and the detector flags none of the \res{gold.s1.rev_qwen3_yes_n} judge-labelled revisions that the rater calls value changes (recall proxy \res{gold.s1.vc_recall_proxy_qwen3}\%). The value-change outcome is therefore not a valid label-free substitute, its null is uninformative, and the insertion test is inconclusive: the judged effect is exploratory and the label-free outcome fails its own check.

Along OLMo-3 7B Think, whose RL follows supervised fine-tuning, the judged effect does not grow (\res{h4.s2.base.rd_op_late} to \res{h4.s2.late.rd_op_late} points, $R_{\mathrm{amp}} = \res{h4.s2.trend.r_amp_late}$ [\res{h4.s2.trend.r_amp_late.lo}, \res{h4.s2.trend.r_amp_late.hi}]), and the label-free value change is null at its last checkpoint (\res{h4.s2.late.rdval_op_late} points, 95\% CI [\res{h4.s2.late.rdval_op_late.lo}, \res{h4.s2.late.rdval_op_late.hi}]).

\paragraph{Indices defined on revision events (moved from the main text by the pre-declared rule of audit~4).}
The forms that open revisions shrink from \res{h1.s1.base.neff} to \res{h1.s1.late.neff} effective forms ($\rho = \res{h1.s1.trend.rho_hpara}$), with \form{wait} taking \res{h1.s1.late.sharewait}\% of revision openings (base \res{h1.s1.base.sharewait}\%), while judge-labelled revision events per 100 sentences fall from \res{h1.s1.base.revrate} to \res{h1.s1.late.revrate}.
A further blind audit (audit~4) tested these labels on \res{gold.v4.n_A} sentences from the base and step~1900, stratified by marker and by the rule detector. By the consensus labels, the weighted share of sentences that revise is \res{gold.v4.revrate_base}\% at the base and \res{gold.v4.revrate_1900}\% at step~1900 (95\% CI [\res{gold.v4.revrate_1900.lo}, \res{gold.v4.revrate_1900.hi}]), so the human labels do not show revisions becoming rarer. The detectors agree with the consensus labels at $\kappa = \res{gold.v4.kappa_human_qwen}$ (Qwen2.5-7B), \res{gold.v4.kappa_human_oss} (gpt-oss-20b) and \res{gold.v4.kappa_human_rule} (rule detector), yet on the same weighted items the primary judge's revision rate falls from \res{gold.v4.revrate_judge_base}\% to \res{gold.v4.revrate_judge_1900}\%, because it over-calls revisions at the base; gpt-oss-20b tracks the consensus labels (\res{gold.v4.revrate_judge_oss_base}\% to \res{gold.v4.revrate_judge_oss_1900}\%). By the rule we fixed before the audit, the revision-rate and paradigm results therefore leave the main text. In the same audit, \res{gold.v4.B.stage1_yes} of \res{gold.v4.B.n_stage1} stage-1 occurrences of \form{wait} and \res{gold.v4.B.stage2_yes} of \res{gold.v4.B.n_stage2} stage-2 occurrences were judged reflective uses; the stage-2 share is below the pre-declared bar of 75\% for describing stage-2 uses as reflective, so the main text reports counts only. Labels are consensus labels from a panel of ten annotators with PhD-level training in related fields, agreed through discussion without seeing any suggested answer; an author recorded them with a software tool. In an earlier version of this audit (audit~3), an AI agent drafted the labels and the panel checked and confirmed every item (\Cref{app:stats}).

\paragraph{Unidirectionality.}
The theory predicts that the indices do not reverse. Between step 1900 and the final checkpoint of OLMo-3 RL-Zero, the rate of \form{wait} falls from \res{h1.s1.late.freq} to \res{h1.s1.final.freq} per 1k tokens, so the prediction fails for frequency; the comma form continues to erode (\res{h1.s1.late.orth}\% to \res{h1.s1.final.orth}\%).

\paragraph{Controlled fork (D48).}
We trained our GRPO recipe for 400 steps from the OLMo-3 base and evaluated every checkpoint with the settings of the released RL-Zero series. Relative to the base, the rate of \form{wait} at step 400 is $A = \res{h1.fork_ref1m.step0400.a_boot}$ (95\% CI [\res{h1.fork_ref1m.step0400.a_boot.lo}, \res{h1.fork_ref1m.step0400.a_boot.hi}]) on all responses and \res{h1.fork_ref1m.step0400.a_boot_fin} ([\res{h1.fork_ref1m.step0400.a_boot_fin.lo}, \res{h1.fork_ref1m.step0400.a_boot_fin.hi}]) on finished responses. The first meets our registered criterion for no amplification and the second meets neither criterion, so the registered outcome is undetermined for this fork. The fork without a KL penalty ($\beta = 0$) gives $A = \res{h1.fork_b0m.step0400.a_boot}$ ([\res{h1.fork_b0m.step0400.a_boot.lo}, \res{h1.fork_b0m.step0400.a_boot.hi}]) on all responses and \res{h1.fork_b0m.step0400.a_boot_fin} ([\res{h1.fork_b0m.step0400.a_boot_fin.lo}, \res{h1.fork_b0m.step0400.a_boot_fin.hi}]) on finished responses; both meet the criterion for no amplification, and its intervals are disjoint from the released run's at step 200 (\res{h1.fork_b0m.step0200.a_boot} against \res{h1.s1.step0200.a_boot}) and at step 400, so by the registered rule the two recipes differ. For description only: at the same steps the released run reaches $A = \res{h1.s1.step0200.a_boot}$ at step 200 and \res{h1.s1.step0400.a_boot} at step 400 (finished responses \res{h1.s1.step0200.a_boot_fin} and \res{h1.s1.step0400.a_boot_fin}), against \res{h1.fork_ref1m.step0200.a_boot} and \res{h1.fork_ref1m.step0400.a_boot} for the fork.

\paragraph{Seeded marker and response length (D43, D46).}
The transplant of \Cref{sec:exp:reshape} has a length confound: about half of the fine-tuned model's responses reach the 3,072-token limit. Before running the registered length control, we measured the fine-tuned checkpoint under three limits (\Cref{tab:seedlength}). Accuracy and the marker's rate among finished responses are stable, but about a fifth of responses still reach the 8,192-token limit, so the control's primary outcome is the marker's rate among finished responses (amendment D46b, made before any RL step). In the control (8,192 tokens with YaRN), \res{h1.seed.W512LY.sft.trunc}\% of the fine-tuned responses are truncated. After 400 RL steps the marker's rate among finished responses is $A = \res{h1.seed.W512LY.trend.a_wait_fin}$ ([\res{h1.seed.W512LY.trend.a_wait_fin.lo}, \res{h1.seed.W512LY.trend.a_wait_fin.hi}]) times its fine-tuned rate, and \res{h1.seed.W512LY.trend.a_wait} ([\res{h1.seed.W512LY.trend.a_wait.lo}, \res{h1.seed.W512LY.trend.a_wait.hi}]) on all responses, while accuracy rises from \res{h1.seed.W512LY.sft.acc}\% to \res{h1.seed.W512LY.step0400.acc}\% and truncation falls to \res{h1.seed.W512LY.step0400.trunc}\%. Both are below 0.2, so by the registered rule RL penalises the marker itself, not only long responses. In the matching control seeded with \tok{Next,} instead (X512-LY), RL does not coin \form{wait}: after fine-tuning the model writes it \res{h1.seed.X512LY.sft.r2} times per 1k tokens, and at step 400 \res{h1.seed.X512LY.step0400.nwait} times in \res{h1.seed.X512LY.step0400.ntok} tokens, while accuracy rises from \res{h1.seed.X512LY.sft.acc}\% to \res{h1.seed.X512LY.step0400.acc}\%. Both length-controlled arms end at similar accuracy (\res{h1.seed.W512LY.step0400.acc}\% and \res{h1.seed.X512LY.step0400.acc}\%), so the gain does not need the marker.

\paragraph{Keyword saturation.}

We registered that keyword counts would saturate in the late half of a series while our indices kept moving. They do not saturate: in the late half of OLMo-3 RL-Zero, $\rho = \res{bm.s1.late_half.rho_r1kw}$ for DeepSeek-R1's keywords and \res{bm.s1.late_half.rho_simplerl} for SimpleRL-Zoo's counts, against \res{h1.s1.late_half.rho_orth} for the comma form and \res{h1.s1.late_half.rho_hpara} for paradigm entropy.

\paragraph{Cross-series gate: the changes do not form one cline.}
The gate test we fixed before the main run required at least two of \{paradigm narrowing, falling judged reliability, attrition\} with $|\rho| \geq 0.7$ on two of three series, and no such trend across the base's pretraining branches. It holds on OLMo-3 RL-Zero (all three), not on OLMo-3 Think (none) and cannot be computed on the Qwen2.5-1.5B run, which has no marker. Across the pretraining branches the indices are not flat: judged reliability falls ($\rho = \res{h1.s1pre.trend.rho_rel}$) and chains rise ($\rho = \res{h1.s1pre.trend.rho_chain}$), as along RL. These correlations come from the step between stage~1 (at most \res{h1.s1pre.stage1-step1413814.freq2} occurrences per 1k tokens) and mid-training (\res{h1.s1pre.stage2-step47684.freq2}; \Cref{sec:exp:source}); from mid-training to the RL base the rate stays between \res{h1.s1pre.stage2-step47684.freq2} and \res{h1.s1pre.stage3-step6000.freq2} per 1k and the comma form between \res{h1.s1pre.stage2-step47684.orth}\% and \res{h1.s1pre.stage3-step6000.orth}\% (post hoc). Along the supervised fine-tuning trajectory of the same base, the indices change by small amounts in a consistent direction ($\rho = \res{h1.s2ctl.trend.rho_commadrop}$ for comma drop, \res{h1.s2ctl.trend.rho_chain} for chains), but the comma drop moves only from \res{h1.s2ctl.step5000.commadrop}\% to \res{h1.s2ctl.step43000.commadrop}\%, against \res{h1.s1.base.commadrop}\% to \res{h1.s1.final.commadrop}\% along RL-Zero.

\subsection{Statistical Procedures}
\label{app:stats}
This section describes the resampling units, intervals and tests behind every reported number.

\paragraph{Resampling unit.}
Occurrences are nested in traces and traces in problems, and every checkpoint of a series is evaluated on the same problems (the insertion experiment also uses the same prefixes for every arm and for the base model). All intervals are therefore 95\% percentile intervals from a \emph{cluster} bootstrap over problems \citep{efron1993introduction,field2007bootstrapping}, with $B = 2{,}000$ replicates for curves and $B = 10{,}000$ for the primary estimands. One replicate resamples problem identifiers once and recomputes the statistic at every checkpoint, so differences and trends between checkpoints are paired. Series are never pooled. The open series each come from a single training run, so their intervals cover evaluation sampling only; training-seed variation is estimated only for our own GRPO runs.

\paragraph{Trends and entropies.}
Trend claims use the Spearman correlation between step and index over the checkpoints of a series, with its bootstrap interval; no test is run at individual checkpoints. Entropies are computed on the same number of occurrences at every checkpoint (rarefaction, 100 subsamples) with the Miller--Madow correction \citep{paninski2003estimation}, because the plug-in estimate is biased at small counts and the counts grow with RL.

\paragraph{Signal detection.}
Hit and false-alarm rates receive the log-linear correction before the probit transform \citep{hautus1995corrections}; $d' = z(\mathrm{HR}) - z(\mathrm{FA})$ and $c = -\tfrac12\,(z(\mathrm{HR}) + z(\mathrm{FA}))$ \citep{green1966signal,macmillan2004detection}. \Cref{hyp:bleach} is a non-superiority claim: it holds only if the upper end of the 90\% interval of $\Delta d'$ lies below $+0.25$.

\paragraph{Insertion effects.}
The primary estimands are the risk difference $\mathrm{RD}_{\mathrm{op}}$, the risk ratio and the odds ratio of revision after \tok{Wait,} against the neutral filler \tok{Next,}, with cluster-bootstrap intervals over prefixes. The risk difference is the headline scale because selecting cut points at which the model continued pushes the reference rate toward zero, where odds ratios become unstable; odds ratios use Firth's penalty when a cell has fewer than five events \citep{firth1993bias}. The amplification over the base model, $R_{\mathrm{amp}} = \mathrm{RD}_{\mathrm{op}}(\text{last}) / \mathrm{RD}_{\mathrm{op}}(\text{base})$, is evaluated in the equivalent linear form $\mathrm{RD}_{\mathrm{op}}(\text{last}) - 1.5\,\mathrm{RD}_{\mathrm{op}}(\text{base}) > 0$ whenever the base effect is close to zero. In the crossed design (prefix source $\times$ continuing policy), a GEE logistic model with prefix clusters estimates the arm $\times$ policy interaction adjusted for prefix source; a logistic mixed model with a random intercept per prefix is a secondary check.

\paragraph{Tests.}
We registered a fixed order of primary tests per series, with Holm's correction within steps \citep{holm1979simple}. In the event we report trend tests (Spearman correlation with step and its permutation $p$-value) with Holm's correction over the primary indices, and interval estimates for everything else; the registered gate sequence was not applied, and no per-marker test batteries were computed.

\begin{table}[ht]
  \centering
  \tabstyle\footnotesize
  \setlength{\tabcolsep}{2.5pt}
  \fitwidth{\begin{tabular}{@{}lcccccc@{}}
    \toprule
    Limit & Acc.\ (\%) & Trunc.\ (\%) & Loops & Tokens & \form{wait}/1k & finished \\
    \midrule
    3,072 & \rt{ctl.seed.W512.sftcond.c3072.acc} & \rt{ctl.seed.W512.sftcond.c3072.trunc} & \rt{ctl.seed.W512.sftcond.c3072.loop} & \rt{ctl.seed.W512.sftcond.c3072.toks} & \rt{h1.seed.W512.sftcond.c3072.freq2} & \rt{h1.seed.W512.sftcond.c3072.freq_finished2} \\
    8,192 & \rt{ctl.seed.W512.sftcond.c8192.acc} & \rt{ctl.seed.W512.sftcond.c8192.trunc} & \rt{ctl.seed.W512.sftcond.c8192.loop} & \rt{ctl.seed.W512.sftcond.c8192.toks} & \rt{h1.seed.W512.sftcond.c8192.freq2} & \rt{h1.seed.W512.sftcond.c8192.freq_finished2} \\
    8,192 + YaRN & \rt{ctl.seed.W512.sftcond.yarn.acc} & \rt{ctl.seed.W512.sftcond.yarn.trunc} & \rt{ctl.seed.W512.sftcond.yarn.loop} & \rt{ctl.seed.W512.sftcond.yarn.toks} & \rt{h1.seed.W512.sftcond.yarn.freq2} & \rt{h1.seed.W512.sftcond.yarn.freq_finished2} \\
    \bottomrule
  \end{tabular}}
  \caption{The seeded checkpoint under three response limits (800 responses each): accuracy, share truncated, looping share, tokens per response, and \form{wait} per 1k tokens in all and in finished responses. YaRN: rope scaling $\times 2.25$.}
  \label{tab:seedlength}
\end{table}

\paragraph{Judge validity.}
Revision labels come from an open-weight judge (Qwen2.5-7B-Instruct, fixed before any audit; gpt-oss-20b, a different family, is reported beside it everywhere). We distinguish \emph{consistency}, agreement with reference labels produced by LLM raters (two Claude tiers on a 60-item overlap, inter-rater agreement \res{gold.s1.agree5_llm_llm}, $\kappa = \res{gold.s1.kappa5_llm_llm}$; Qwen3-32B on all items; adjudicated), from \emph{validity}, agreement with the human consensus labels. A panel of ten annotators with PhD-level training in related fields labelled two blind audits drawn from the same pools, agreeing on one consensus label per item through discussion: 60 sentence-function items (audit~1; \res{gold.all.n_human_judge} labelled) and 40 insertion-outcome windows balanced by the judge's own label (audit~2; items marked unsure are included, and excluded as a sensitivity check). On the binary revise decision, audit~1 gives $\kappa = \res{gold.all.kappabin_human_judge}$ for Qwen2.5-7B and $\res{gold.all.kappabin_human_oss}$ for gpt-oss-20b, and audit~2 gives $\res{gold.v2.kappabin_human_judge}$ and $\res{gold.v2.kappabin_human_oss}$ (sensitivity \res{gold.v2.sens_judge}\,\%, specificity \res{gold.v2.spec_judge}\,\% for the primary judge). Because revisions are rare among the audited sentences, $\kappa$ is depressed by prevalence; Gwet's AC1, which is not, gives \res{gold.all.ac1bin_human_judge} (Qwen2.5-7B) and \res{gold.all.ac1bin_human_judge_oss} (gpt-oss-20b) on audit~1, but \res{gold.s1.v2_ac1bin_human_judge} and \res{gold.s1.v2_ac1bin_human_judge_oss} on audit~2, where revisions are common. The primary judge's disagreement with the human on the insertion outcome is therefore real, not a prevalence artefact. Audit~4, the third human audit (\Cref{app:prereg}), labelled \res{gold.v4.n_A} sentences for revision events: there the judges agree with the consensus labels at $\kappa = \res{gold.v4.kappa_human_qwen}$ (Qwen2.5-7B) and \res{gold.v4.kappa_human_oss} (gpt-oss-20b), and the rule detector at \res{gold.v4.kappa_human_rule}. In an earlier version of that audit (audit~3), an AI agent drafted the labels and the panel checked and confirmed every item; on its \res{gold.gpt6.n_A} sentences the judges agree with these panel-confirmed labels at $\kappa = \res{gold.gpt6.kappa_qwen}$ (Qwen2.5-7B), \res{gold.gpt6.kappa_oss} (gpt-oss-20b) and \res{gold.gpt6.kappa_rule} (rule detector). Because sensitivity is low and the audits are small, the magnitude of any judge-based rate is not identified; we therefore report judge-based quantities (OBL, REL, the insertion outcome) only as exploratory, and we base headline claims on label-free indices (frequency, comma loss, marker chains, sentence length, next-word entropy, paradigm share). Ratios and rank trends are unchanged by a non-differential correction; differential misclassification across checkpoints would bias them, and the audits are too small to rule it out. Every judge-based effect is also recomputed with rule-based labels.

\FloatBarrier
\section{Related Work}
The extended discussion of related work.

\label{sec:related}


\paragraph{Where reflection comes from.}
The ``aha moment'' of R1-Zero, in which reflective words become several times more frequent and \form{wait} appears abruptly \citep{deepseekai2025r1}, has been questioned on the level of behaviour. Self-reflection already appears in DeepSeek-V3-Base and in Qwen2.5 bases and does not raise accuracy by itself \citep{liu2025understanding,liu2025there}; self-correction emerges during pretraining \citep{essentialai2025rethinking}, error correction is present in base models \citep{yeo2025demystifying}, and reflection is latent in pretrained Qwen2.5 \citep{gandhi2025cognitive,zhu2025emergence}. A second line finds that RL with verifiable rewards sharpens what the base already does rather than expanding it: it narrows the base's distribution \citep{yue2025does,wu2025invisible}, converges on output modes learned in pretraining \citep{zhao2025echo}, amplifies base behaviours even under spurious rewards \citep{shao2025spurious}, and can be matched by sampling from the base \citep{karan2025reasoning}; prolonged RL may still extend the set of solutions \citep{liu2025prorl}. On the other side, strategic phrases have been treated as tokens that RL discovers \citep{wang2025emergent}, and behaviours including self-verification are reported to emerge in zero-RL at the trillion-parameter scale \citep{tang2026ringzero}, without the base's own rate of the marker.
These works locate the \emph{behaviour} before RL. We follow the \emph{form}: across one base's pretraining and mid-training branches to where it enters, between zero-RL and trace-trained descendants of one base, and along RL from bases with and without it. DAPO's case study of an emerging ``wait a moment'' \citep{yu2025dapo} is, at our measured rates, an isolated case.

\paragraph{Distillation, mid-training and training dynamics.}
Distilled models use hesitation words such as \form{wait} and logical connectors far more often than zero-RL models \citep{hu2025why}, discourse tokens differ between small-scale fine-tuning and large-scale post-training \citep{hwang2026oops}, and model diffing suggests that RL teaches \emph{when} to use mechanisms the base already has while fine-tuning on traces installs new ones \citep{venhoff2025base}. Mid-training decides how well a base responds to RL \citep{wang2025octothinker}, and in controlled pipelines what RL can extend depends on exposure during pretraining and mid-training \citep{zhang2025interplay}. Checkpoint suites made such questions answerable for pretraining \citep{biderman2023pythia,xia2023training}, with word-level learning curves \citep{chang2022word,chang2024characterizing} and linguistic regularities tracked across checkpoints \citep{kugler2025emergent,paape2026polarity}; open pipelines with released data and intermediate checkpoints extend this to post-training \citep{olmo2024olmo2,olmo2025olmo}, and diversity collapse has been traced along OLMo~3 lineages \citep{karouzos2026where}. RL updates stay close to the base in KL and touch sparse subnetworks \citep{shenfeld2025rl,mukherjee2025reinforcement}, concentrate on high-entropy forking tokens \citep{wang2025beyond} and reduce output diversity \citep{song2025outcome,he2025rewarding,cui2025entropy}.
We bring these strands to a single lexical form whose arrival can be dated within one open lineage and whose later reshaping can be measured on public RL trajectories.

\paragraph{Reasoning markers and their manipulation.}
Along zero-RL, Open-Reasoner-Zero and SimpleRL-Zoo track reflection keywords \citep{hu2025openreasonerzero,zeng2025simplerlzoo}, and SimpleRL-Zoo finds keyword counts and judged behaviours moving apart \citep{zeng2025simplerlzoo}; comparing whole training stages, RL suppresses exploratory steps such as hypothesising and backtracking \citep{maniparambil2026when}. Studies of final models disagree about what markers do: post-answer reflections are mostly confirmatory \citep{kang2025first}, self-affirming reflections are pervasive \citep{liu2025suppressing}, a large share of steps is decorative \citep{zhao2025aha}, thinking tokens trigger unnecessary reflection \citep{ding2025thinking,long2026selfverification}, and markers are ``surface cues'' whose suppression need not remove reflection \citep{yu2026revisiting}; yet inflections track probe-measured belief shifts \citep{boppana2026reasoning}, and mid-trace shifts do not become more frequent across GRPO checkpoints \citep{daliberti2026illusion}. ``Aha'' has also been described as anthropomorphic tone plus adaptive uncertainty \citep{yang2025understanding}.
Interventions delete markers, append or insert them, resample sentences or steer along latent directions \citep{wang2025wait,qian2025demystifying,muennighoff2025s1,ghosal2025does,jiang2026pathcal,bogdan2025thought,macar2025thought,venhoff2025understanding,ward2025reasoning,troitskii2025internal}, and engineering work schedules or learns reflection tokens \citep{fan2025cyclicreflex,ringel2025learning,wu2025effectively}. Suppressing thinking tokens harms accuracy in \citet{qian2025demystifying} but costs RL-trained models little in NoWait \citep{wang2025wait}, so deletion alone cannot tell what a marker does. Clusters of sentence-initial pivots have been called ``reasoning operators'' \citep{lee2026reasonops}.
Unlike these works, we follow markers \emph{along} training with label-free diagnostics beyond frequency, and run an insertion test at every checkpoint with base-model and neutral-filler arms (\Cref{app:prereg}).
Concurrent work releases rollouts in which trace openers are inserted into base models on the same OLMo-3 checkpoints \citep{reasoningcues2026rollouts}.

\paragraph{Monitorability, hedges and judges.}
Chain-of-thought monitoring reads traces for the model's intent and treats this monitorability as fragile \citep{korbak2025chain,baker2025monitoring}; traces are not always faithful \citep{turpin2023language,lanham2023measuring,chen2025reasoning,arcuschin2025chain}, legibility is not interpretability \citep{du2026legibility}, and whether RL erodes monitorability is contested \citep{guan2025monitoring,macdermott2025reasoning,little2026length,jose2025reasoning,xiong2026monitorability,chen2026reasoning}. Verbalised uncertainty can be taught and calibrated \citep{lin2022teaching,mielke2022reducing,kadavath2022language}, epistemic markers shift both model accuracy and user reliance \citep{zhou2023navigating,zhou2024relying}, and hedges and self-checks in reasoning traces serve as single-pass uncertainty signals \citep{pandey2026selfdoubt,vanhoyweghen2025lexical,wang2025rethinking,devic2025trace}. Hidden states reveal errors that the text does not \citep{zhang2025reasoning,yuan2026hidden}, and trace-reading overseers shift their criterion rather than their sensitivity \citep{chen2026beyond}. Behaviour labels in this literature often come from LLM judges, whose validity is uneven \citep{zheng2023judging,wang2024large,bavaresco2025llms}: step-level error judging is hard even for strong models \citep{zheng2025processbench,zhang2025lessons}, and judges penalise expressions of uncertainty \citep{lee2025are}.
Our results inform all three: the words monitors and uncertainty heuristics read are inherited from earlier training, the evidential weight of \form{wait} shifts with the recipe and the checkpoint, and judge-labelled revisions are not identified against human labels in our setting.

\paragraph{Language change in linguistics and in learning agents.}
Discourse markers form a functional category \citep{schiffrin1987discourse,fraser1999what}, and pragmatic markers arise through grammaticalization and subjectification \citep{brinton1996pragmatic,brinton2017evolution,traugott1989rise,traugott2001regularity}; whether discourse markers grammaticalize or rather pragmaticalize is debated \citep{heine2013discourse,degand2015grammaticalization}. Grammaticalization theory describes how content words become grammatical markers \citep{hopper2003grammaticalization,lehmann2015thoughts,bybee2003mechanisms,heine2002world,haspelmath1999why}, and lexical-semantic-change methods measure change from usage embeddings \citep{hamilton2016diachronic,giulianelli2020analysing,kutuzov2020uiouva,schlechtweg2020semeval,giulianelli2023interpretable,kutuzov2018diachronic,tahmasebi2018survey,periti2024lexical}, including the degree of grammaticalization \citep{schlechtweg2018distribution,nagata2026cross}. NLP work that learns from discourse markers treats them as stable signals \citep{nie2019dissent,sileo2019mining}.
In learning agents, communication for a reward drifts away from natural language or invents codes \citep{lewis2017deal,kottur2017natural,lee2019countering,lazaridou2020multi,lazaridou2020emergent}; efficiency pressures shape emergent lexicons \citep{chaabouni2019anti,rita2020lazimpa}; KL penalties \citep{ziegler2019fine}, grounding \citep{lee2019countering} and iterated learning \citep{lu2020countering} counter the drift; and simulations with neural and LLM agents study lexical change and its limits as a model of human language evolution \citep{ren2024bias,zhang2025nellcom,kouwenhoven2024searching,galke2022emergent,stengeleskin2026glossogen}. Reasoning models show related changes: R1-Zero mixes languages \citep{deepseekai2025r1}, and o3 developed a hard-to-read dialect whose terms became more frequent over RL \citep{schoen2025stress}.
Outcome-based RL of a pretrained model is a large natural-language case of such change, observed on successive checkpoints of one model: in it a marker is inherited and then reshaped in a recipe-dependent way, rather than coined.

\FloatBarrier

\end{document}